%% file: main.tex
\documentclass{article} %
\usepackage{iclr2027_conference,times}

\input{math_commands.tex}

\usepackage{hyperref}
\usepackage{url}

\input{preamble}

\title{Just Weather Scoring: Efficient End-to-End Nowcasting with Distributional Diffusion}

\author{
Jannik Wiese\thanks{equal contribution}
\And
Johannes Schusterbauer\footnotemark[1]
\And
Tommaso Martorella
\And
Bj\"orn Ommer
\AND
\normalfont
\centerline{CompVis @ LMU Munich,
\quad Munich Center for Machine Learning (MCML)}
\\
}

\iclrfinalcopy %
\begin{document}

\maketitle

\input{sec/0_abstract}

\FloatBarrier

\begin{figure}[h]
    \centering
    \begin{subfigure}{0.74\textwidth}
    \includegraphics[width=\linewidth]{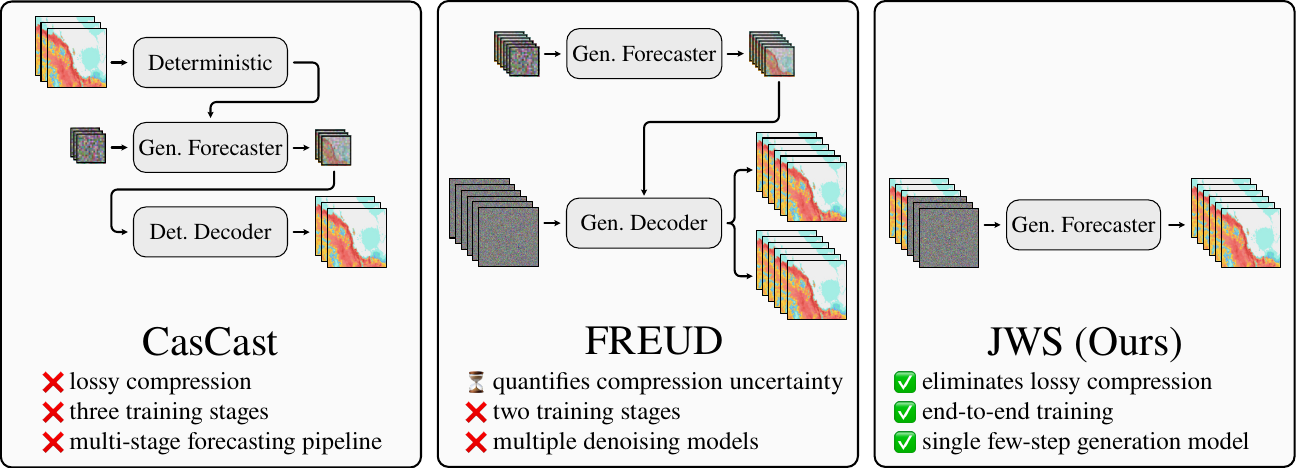}
    \end{subfigure}
    \begin{subfigure}{0.231\textwidth}
    \includegraphics[width=\linewidth]{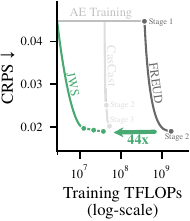}
    \end{subfigure}
    \caption{\textbf{Nowcasting made easy.} We introduce \ours, an end-to-end, data-space nowcasting model, greatly simplifying training and inference while being $44 \times$ more efficient to train than the state-of-the-art FREUD model at comparable CRPS.}
    \label{fig:front-page-fig}
\end{figure}

\FloatBarrier

\section{Introduction}
\label{sec:intro}

\input{sec/1_intro}

\section{Related Work}
\label{sec:related}

\input{sec/2_related}

\section{Method}
\label{sec:method}

\input{sec/3B_method}

\section{Experiments}
\label{sec:experiments}

\input{sec/4_experimentsB}

\section{Conclusion}
\label{sec:conclusion}
\input{sec/5_conclusion}

\newpage

\subsection*{Acknowledgments}
We thank Ming Gui and Ulrich Prestel for providing helpful feedback on an early draft of the paper and Owen Vincent for continuous technical support.
This project has been supported by the bidt project KLIMA-MEMES, the Horizon Europe project ELLIOT (GA No.\ 101214398), the project ``GeniusRobot'' (01IS24083) funded by the Federal Ministry of Research, Technology and Space (BMFTR), the BMWE ZIM-project (No.\ KK5785001LO4) ``conIDitional LoRA'', and the German Federal Ministry for Economic Affairs and Energy within the project ``NXT GEN AI METHODS - Generative Methoden für Perzeption, Prädiktion und Planung''. The authors gratefully acknowledge the Gauss Center for Supercomputing for providing compute through the NIC on JUWELS/JUPITER at JSC and the HPC resources supplied by the NHR@FAU Erlangen.

\subsection*{AI use statement}
In this work, we used generative AI tools to design and provide feedback on the methodology and experiments, as well as to assist with implementing methods.
We have not used generative AI tools to help develop theoretical or conceptual frameworks, formulate mathematical claims, provide critical ingredients for proving mathematical claims, propose or refine hypotheses, assist with translation, cleaning, and reformatting datasets, support qualitative and thematic data analysis, or interpret results.
Synthetic data generation and assistance in writing proofs are not applicable to this work.
Additionally, we used generative AI tools to create or modify scientific figures, develop and edit software code, summarize existing literature, brainstorm, source and search for information, edit the research paper for readability, and identify relevant literature.
We have reviewed all AI-assisted work. Methodology, experiments, and brainstorming results were discussed among at least two researchers. LLM-generated code was verified and tested. The LLM-assisted literature search was followed up by a manual literature review and verification of the summaries by at least two researchers. Finally, AI-generated figures were manually verified and improved. Figures were discussed with at least two researchers to ensure correctness and clarity.
We take responsibility for the final content of this work, including text, claims, or artifacts produced with generative AI.

\subsection*{Reproducibility statement}

\ours{} is trained using publicly available datasets~\cite{veillette2020SEVIR, larvor2020meteonet}. We detail training in Appendix~\ref{sec:sup:implementation_details} and data processing in Appendix~\ref{sec:sup:metrics:datasets}.
\ours{} is a simple method that works with standard transformers (\cref{tab:exp:architecture_comparison}) requiring only minimal modifications (\cref{sec:method}). Additionally, we detail hyperparameter settings in the Appendix in~\cref{tab:sup:hyperparams}.
We build upon existing evaluation pipelines from~\citet{gong2024cascast} and \citet{schusterbauer2026weatherrf-freud} and detail evaluation protocols and metrics in Appendix~\cref{sec:sup:metrics}.
Therefore, \ours{} is easy to reproduce from scratch.
However, we additionally release code, model checkpoints, and evaluation pipelines.

\bibliography{iclr2027_conference}
\bibliographystyle{iclr2027_conference}

\appendix

\input{sec/X_suppl}

\end{document}

%% file: math_commands.tex
\usepackage{amsmath,amsfonts,bm}

\def\eqref#1{equation~\ref{#1}}

\def\1{\bm{1}}

\DeclareMathAlphabet{\mathsfit}{\encodingdefault}{\sfdefault}{m}{sl}
\SetMathAlphabet{\mathsfit}{bold}{\encodingdefault}{\sfdefault}{bx}{n}

%% file: preamble.tex
\usepackage{array}
\usepackage[table]{xcolor}
\usepackage{graphicx}
\usepackage{amsmath}
\usepackage{amssymb}
\usepackage{booktabs}
\usepackage{float}
\usepackage{comment}
\usepackage[accsupp]{axessibility}
\usepackage{tikz}

\usepackage{algorithm}
\usepackage{algpseudocode}

\usepackage{tabularx}
\usepackage{multirow}
\usepackage{adjustbox}
\usepackage{makecell}
\usepackage{environ}
\usepackage{subcaption} %
\DeclareCaptionFormat{corruptiononeline}{\makebox[\linewidth][c]{#1#2#3}}
\usepackage{placeins}
\usepackage{minted}
\usepackage{listings}
\usepackage[capitalise]{cleveref}
\usepackage{pifont}
\usepackage{wrapfig}

\crefname{figure}{Fig.}{Figs.}
\crefname{table}{Tab.}{Tabs.}
\crefname{section}{Sec.}{Secs.}

\Crefname{figure}{Figure}{Figures}
\Crefname{table}{Table}{Tables}
\Crefname{section}{Section}{Sections}

\let\cite\citep

\newcolumntype{Y}{>{\centering\arraybackslash}X}

\newif\ifisdraft
\isdraftfalse
\isdrafttrue
\NewEnviron{drafty}{%
    \ifisdraft
        {\color{lightgray} \BODY}
    \fi
}

\NewEnviron{TODO}{%
    \ifisdraft
        {\color{red} \BODY}
    \fi
}

\newcommand{\ven}[1]{\tiny\color{gray}(#1)}

\usepackage{xcolor}
\usepackage{listings}
\usepackage{algorithm}

\definecolor{codecomment}{RGB}{55,130,135}
\definecolor{codefunction}{RGB}{235,35,125}
\definecolor{codeoperator}{RGB}{60,35,240}

\lstdefinestyle{paperpython}{
    language=Python,
    basicstyle=\ttfamily\footnotesize,
    commentstyle=\color{codecomment},
    keywordstyle=\color{codeoperator},
    morekeywords={True,False,None},
    emph={sample_t,randn_like,net,max,abs,mean,sum,
          upper_triangle_indices,masked_mean,update},
    emphstyle=\color{codefunction},
    literate=*{+}{{{\color{codeoperator}+}}}1
              {-}{{{\color{codeoperator}-}}}1
              {*}{{{\color{codeoperator}*}}}1
              {/}{{{\color{codeoperator}/}}}1,
    columns=fullflexible,
    keepspaces=true,
    showstringspaces=false,
    frame=none,
    numbers=none,
    aboveskip=0.5em,
    belowskip=0.5em,
    escapeinside={(*@}{@*)}
}

\newcommand{\codecref}[1]{%
  {\hypersetup{linkcolor=codecomment}%
   \color{codecomment}%
   \crefformat{equation}{(Eq.~##2##1##3)}%
   \cref{#1}}%
}

\definecolor{codeblue}{RGB}{70,136,140}     %
\definecolor{codefunc}{RGB}{220,20,120}     %
\definecolor{codesign}{RGB}{40,40,255}      %
\definecolor{algoyellow}{RGB}{255,215,0}

\newcommand{\Ours}{\textit{Just Weather Scoring}}
\newcommand{\ours}{JWS}

\newcommand{\Paradigm}{\textit{Masked Asynchronous Diffusion}}
\newcommand{\paradigm}{MAD}

\newcommand{\Objective}{\textit{simple Distributional Diffusion Models}}
\newcommand{\objective}{sDDM}

%% file: sec/0_abstract.tex
\begin{abstract}
Generative diffusion models are well-suited for probabilistic precipitation nowcasting, but existing approaches often rely on separately trained compression or deterministic forecasting components and remain costly at inference due to iterative denoising.
We introduce \Ours~(\ours), a single-stage, end-to-end diffusion model which addresses both issues by forecasting directly in radar space and enabling few-step generation.
Radar-space modeling greatly simplifies training and inference and eliminates uncertainty arising from lossy compression.
\ours{} combines \Paradigm, a timestep-sampling scheme that preserves clean context while adapting diffusion training to high-dimensional spatio-temporal data, with a simple scoring-rule objective that aligns training with probabilistic forecasting and unlocks few-step generation.
On the SEVIR and MeteoNet benchmarks, \ours{} achieves state-of-the-art probabilistic forecasting performance at reduced training and inference cost.
Even our smallest model remains competitive using substantially fewer parameters and more than $17 \times$ faster inference.\\
\textbf{Code:} {\small\url{https://github.com/CompVis/jws}} \\
\textbf{Project page:} {\small\url{https://compvis.github.io/jws}}
\end{abstract}

%% file: sec/1_intro.tex
Short-term, high-resolution precipitation nowcasting is essential for applications ranging from renewable energy planning~\cite{sperati_application_2016, wu2018Probabilistic} to emergency response~\cite{salio2025nowcasting}.
Useful nowcasts must resolve spatial details, update rapidly, and quantify uncertainty arising from chaotic dynamics and limited observability~\cite{ravuri21:GAN}.
Compared to numerical weather prediction, machine learning offers faster inference and learns directly from data.
Yet, regression averages plausible forecasts and thus produces increasingly blurry forecasts as uncertainty grows.
Generative models instead sample plausible precipitation from the conditional distribution, enabling sharp individual forecasts and ensemble-based uncertainty quantification.
Reliable probabilistic forecasting requires representing the full range of possibilities rather than concentrating on a few outcomes. Therefore, diffusion models are a natural fit due to their distributional coverage.

However, many diffusion-based nowcasting approaches rely on multi-stage training pipelines.
Inspired by image generation~\cite{rombach2022ldm}, latent space methods first train a compression model and learn denoising of forecasts only on compressed representations to reduce cost~\cite{gao2023prediff, leinonen2023ldcast}.
Other approaches use a deterministic forecast to condition generation~\cite{gong2024cascast} or define a residual prediction task~\cite{yu2024DiffCast}, resulting in up to three components.
Beyond training complexity, these components introduce uncertainty and biases.
While visually minor imperfections are acceptable in image synthesis, small errors in precipitation intensity and location can have a consequential impact.
FREUD~\cite{schusterbauer2026weatherrf-freud} partially addresses this concern by modeling uncertainty from lossy compression with a generative decoder. 
Nevertheless, it retains a separate compression stage, and quantification of compression uncertainty requires ensembling many reconstructions per forecast, leading to multiplicative cost.

Therefore, prior work leaves the central challenge unresolved: how to build a simple, uncertainty-aware nowcasting model that makes efficient use of training and inference compute.
We introduce \Ours~(\ours), an end-to-end generative model that avoids additional stages and operates directly on uncompressed data.
We enable forecasting with a single transformer that is parametrized to predict clean radar maps~\cite{li2025back} and downsamples only by patchification. Our single-stage approach is more efficient and capable than pipelines with multiple trained components.
Further, we introduce \Paradigm~(\paradigm), preserving clean conditioning while sampling correlated, framewise noise levels for predicted timesteps, aligning training with forecasting.
Thus, by training with \paradigm{}, \ours{} supports flexible conditioning.
Spatiotemporal resolution-aware timestep sampling in \paradigm~adapts corruption levels to uncompressed radar data.

As iterative denoising remains costly, we aim to reduce sampling steps.
Flow Map methods~\cite{boffi2025flow} reduce denoising steps by directly learning the integration over diffusion time. However, canonical approaches~\cite{geng2025mean, frans2025one, boffi2025how} incur substantial overhead to construct training targets.
Functional Generative Networks (FGNs)~\cite{alet2025fgn} use scoring-rule supervision to directly optimize probabilistic forecasts but introduce stochasticity only by noise injection into deterministic predictors, potentially limiting distributional coverage.

We seek to combine scoring-rule supervision with diffusion model coverage.
Distributional Diffusion Models (DDM)~\cite{bortoli2025distributional} replace vector field regression with scoring-rule supervision, training denoising models to represent the conditional posterior through ensembles of vector fields.
With a conditional posterior, reverse transitions recover the data distribution even on a coarse time grid, enabling few-step inference~\cite{bortoli2025distributional}.
However, balancing few- and many-step generation previously required careful tuning of interdependent variables.
We propose \Objective~(\objective), which are trained with a hyperparameter-free objective, linearly interpolating distributional CRPS and squared error.

With \paradigm~and \objective, \ours~generates radar sequences directly in pixel space, eliminating additional pipeline stages and allocating computation entirely to generative forecasting.
Our key insight is that standard transformers can support efficient end-to-end nowcasting by combining task-specific timestep sampling with distributional supervision.
This yields simple and efficient training and inference with superior probabilistic forecasting, without lossy compression or deterministic priors.

Our contributions are as follows:
\begin{itemize}
    \item \textbf{End-to-end training:} \ours{} is trained directly on sensor data without separately trained compression or deterministic forecasting stages, simplifying training and inference.
    \item \textbf{Nowcasting-centric recipe:} \Paradigm{} aligns conditioning observed during training and inference while preserving test-time flexibility.
    \item \textbf{Simple distributional objective:} \Objective{} are trained with a hyperparameter-free objective, improving forecasting and enabling few-step sampling. 
    \item \textbf{Efficient nowcasting:} Taken together, these components enable even our smallest model to achieve competitive nowcasting with approximately $17\times$ fewer parameters than FlowCast and substantially lower inference cost than prior methods.
\end{itemize}

%% file: sec/2_related.tex
An extended Related Work section is found in Appendix \cref{sec:sup:extended_rw}.

\paragraph{Precipitation nowcasting.}
Nowcasting refers to short-term, high-resolution forecasting.
As rapid updates are typically required, numerical weather prediction systems (NWP) are unsuitable for precipitation nowcasting~\cite{shi15:ConvLSTM, leinonen2023ldcast, ravuri21:GAN}.
Machine learning offers faster inference~\cite{shi15:ConvLSTM, agrawal19:UNet, wang2017predrnn, gao2022earthformer}.
Yet, training with regression objectives yields blurry forecasts under uncertainty, thus the field shifted towards generative nowcasting methods~\cite{ravuri21:GAN, zhang2023skilful, leinonen2023ldcast, gao2023prediff}.
Initially, generative adversarial networks were used~\cite{goodfellow2014gan, ravuri21:GAN, zhang2023skilful, ji22:CLGAN}, yet adversarial training can induce mode collapse~\cite{kossale2022Mode}. Therefore, diffusion-based nowcasting has become popular due to strong mode coverage and a principled mathematical foundation~\cite{ho2020ddpm, dhariwal2021beatGANs, lipman2023flowmatching}.
To reduce denoising complexity, two-stage latent diffusion methods are commonly used~\cite{leinonen2023ldcast, gao2023prediff, gong2024cascast, ribeiro2026flowcast, schusterbauer2026weatherrf-freud} or generation is conditioned on deterministic priors~\cite{yu2024DiffCast, gong2024cascast, zhu2026pixelflowcast}.
Yet, FREUD~\cite{schusterbauer2026weatherrf-freud} finds these priors reduce distributional coverage, and inherently lossy compression induces additional uncertainty.
They ensemble generative reconstructions, introducing computational overhead and maintaining separately trained stages.
\citet{alet2025fgn} propose to directly supervise ensembles and predict multiple forecasts by perturbing parameters through noise injection. Yet, all stochasticity must be captured by the noise vector, thereby potentially limiting coverage.
We instead propose to train end-to-end radar-space nowcasting, eliminating compression stages completely while retaining diffusion model coverage.

\paragraph{Pixel-space modeling.}
Latent diffusion~\cite{rombach2022ldm} enables high-resolution image~\cite{esser2024scaling} and video~\cite{ho2022video} generation.
Perceptual~\cite{zhang2018lpips} and adversarial~\cite{goodfellow2014gan} losses typically encourage losing mostly imperceptible details.
However, the importance of small-scale variations in nowcasting limits applicability.
Previously, architecture adaptations for pixel-space generation were explored~\cite{hoogeboom2023simple, yu2025pixeldit, crowson2024hdit}.
Recently, \citet{li2025back} showed that standard transformers are indeed suitable.
However, pixel-space nowcasting is especially challenging due to high data dimensionality.
While RainDiff~\cite{nguyen2025raindiff} and PixelFlowCast~\cite{zhu2026pixelflowcast} perform pixel-space nowcasting, they rely on specialized architectures or only model residuals of deterministic forecasts.
In comparison, we use standard transformers for end-to-end precipitation nowcasting.

\paragraph{Few-step sampling.}
Iterative denoising remains a bottleneck of diffusion.
Distillation~\cite{salimans2022progressivea, sauer2023adversarial, song2023consistencya, yin2024onestep} drastically reduces latency, but requires two-stage training.
Flow Maps~\cite{boffi2025flow} directly integrate larger steps.
However, prominent MeanFlows~\cite{geng2025mean} induce substantial training overhead, while Shortcut models~\cite{frans2025one} invest compute to produce self-bootstrapped targets.
Alternatively, Distributional Diffusion Models (DDM)~\cite{bortoli2025distributional} train denoising models with ensembles and scoring rule supervision, enabling few-step inference.
Moreover, distributional supervision aligns with probabilistic forecasting~\cite{alet2025fgn}.
To scale DDM training, improved Distributional Diffusion Models (iDDM)~\cite{martorella2026iddm} expand the ensemble only for final layers. Yet, their loss relies on tuning interdependent parameters.
We replace iDDM's tuned parameter schedules with a simple interpolation between CRPS and squared error, achieving better nowcasting results without additional loss hyperparameters.

%% file: sec/3B_method.tex
\subsection{Preliminaries}
\label{sec:preliminaries}
\label{sec:spatiotemporal}

\paragraph{Precipitation nowcasting.}
Given $C$ radar observations $\mathbf{x}^{1:C}$, with $\mathbf{x}^t\in\mathbb{R}^{H\times W}$ at time $t$, the goal of probabilistic precipitation nowcasting is to model the distribution $p(\mathbf{x}^{C+1:C+L}\mid\mathbf{x}^{1:C})$ of the next $L$ precipitation frames. At inference, we sample multiple sequences to form a forecast ensemble, keeping the observed past fixed.

\paragraph{Spatio-temporal diffusion.}
Diffusion models learn to transform samples from a simple prior distribution into samples from the data distribution through denoising~\cite{ho2020ddpm,albergo2023stochastic,lipman2023flowmatching}. In our setting, each sample is a complete radar sequence $\mathbf{x}_1\in\mathbb{R}^{T\times H\times W}$ of length $T=C+L$. Given a clean sequence $\mathbf{x}_1\sim p_{\mathrm{data}}$ and independent Gaussian noise $\mathbf{x}_0\sim\mathcal{N}(0,\mathbf{I})$, we use a linear interpolant to construct noisy sequences at diffusion time $\tau\in[0,1]$:
\begin{equation}
    \mathbf{x}_\tau=\tau\mathbf{x}_1+(1-\tau)\mathbf{x}_0,
    \qquad \text{with}\quad \tau\sim p(\tau).
    \label{eq:fm_interpolant}
\end{equation}
Here $\tau\in[0,1]$ controls corruption from noise ($\tau=0$) to clean data ($\tau=1$) and differs from the frame index $t$. The linear path has velocity $\mathbf{v}=\mathbf{x}_1-\mathbf{x}_0$. A Flow Matching~\cite{lipman2023flowmatching} model $\mathbf{v}_\theta$ is trained to predict this velocity by minimizing the squared error
\begin{equation}
    \mathcal{L}_{\mathrm{FM}}
    =\mathbb{E}_{\mathbf{x}_0,\mathbf{x}_1,\tau}
    \bigl[\|\mathbf{v}-\mathbf{v}_\theta(\mathbf{x}_\tau,\tau)\|_2^2\bigr].
    \label{eq:fm_loss}
\end{equation}
At inference, we initialize a Gaussian noise sequence and numerically integrate the learned ODE $d\mathbf{x}_\tau/d\tau=\mathbf{v}_\theta(\mathbf{x}_\tau,\tau)$ from $\tau=0$ to $\tau=1$. The resulting sample follows the model distribution $p_{\theta}$, approximating $p_{\text{data}}$~\citep{lipman2023flowmatching}.
Sampling independent conditional forecasts enables ensemble-based uncertainty estimation. As ensemble size increases, sample variance converges to the learned predictive variance, which approximates aleatoric uncertainty~\cite{chan2024HyperDiffusion}.

Standard sequence diffusion uses a single diffusion time shared across frames. Hence, it corrupts all frames to the same degree. More generally, framewise corruption can be described by $\boldsymbol{\tau}=(\tau^1,\ldots,\tau^T)$, with $\mathbf{x}_{\boldsymbol{\tau}}^t=\tau^t\mathbf{x}_1^t+(1-\tau^t)\mathbf{x}_0^t$. This accommodates clean context ($\tau^t=1$) alongside noisy frames ($\tau^t<1$), whose diffusion times may be shared or vary across the sequence.

\subsection{Pixel-Space Nowcasting}
\label{sec:pixelspace}

\begin{figure}[!ht]
    \centering
    \includegraphics[width=0.8\linewidth]{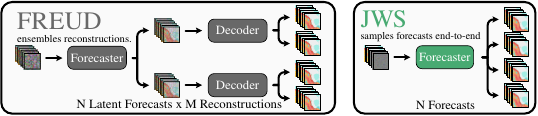}
    \caption{\textbf{Forecast ensembles in pixel space.}
    \textbf{Left:} FREUD~\cite{schusterbauer2026weatherrf-freud} samples multiple reconstructions per forecast to represent reconstruction uncertainty.
    \textbf{Right:} JWS directly samples radar forecasts with a single model, removing the decoding stage.}
    \label{fig:method}
\end{figure}

We aim to perform spatiotemporal diffusion directly on sensor data, generating radar fields in a single stage (\cref{fig:front-page-fig}). 
Eliminating latent compression avoids the associated uncertainty from lossy compression, devoting the inference budget to ensembling forecasts (\cref{fig:method}).
To support high-dimensional diffusion, we adapt conditioning, prediction targets, and training-timestep sampling.

\paragraph{Conditioning on observations.}
Forecasting inherently depends on available past observations. While training only for predicting $L$ frames given $C$ context frames is the trivial solution, more flexible conditioning is needed in practice~\cite{schusterbauer2026weatherrf-freud}.
Therefore, we first sample $k \sim \mathcal{U}(1, k_{\text{max}}), k_{\text{max}} \leq T$ and then sample a set $\mathcal{C}$ of $k = \lvert \mathcal{C} \rvert$ frame indices randomly without replacement. The corresponding frames are used to form clean context. Therefore, $\tau^t = 1, \forall t \in \mathcal{C}$.
With probability $p_U$ we train for unconditional generation and set $\mathcal{C} = \emptyset$.
For all frames that are not in the conditioning set, we sample individual diffusion times from a distribution $p(\tau)$. Thus $\tau^t \sim p(\tau), \forall t \notin \mathcal{C}$.
At inference, this permits forecasting by conditioning on past observations ($\mathcal{C} = \{1, \dots, C\}$), infilling by excluding missing frames from $\mathcal{C}$, and conditioning on corrupted data by setting $\tau^t \neq 1$ to express confidence in individual frames.

\begin{figure}[!tb]
    \centering
    \def\corruptionPanelHeight{2.2cm}
    \captionsetup[subfigure]{justification=centering}
    \subcaptionbox{$\boldsymbol{\tau}$ sampling.\label{fig:met:tbar}}{%
        \includegraphics[height=\corruptionPanelHeight]{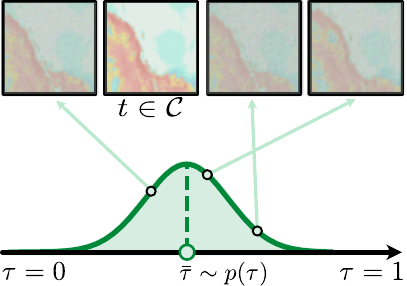}%
    }\hfill
    \begingroup
        \captionsetup[subfigure]{format=corruptiononeline,singlelinecheck=false}
        \subcaptionbox{Timestep shift.\label{fig:dim-shifted-time}}{%
            \includegraphics[height=\corruptionPanelHeight]{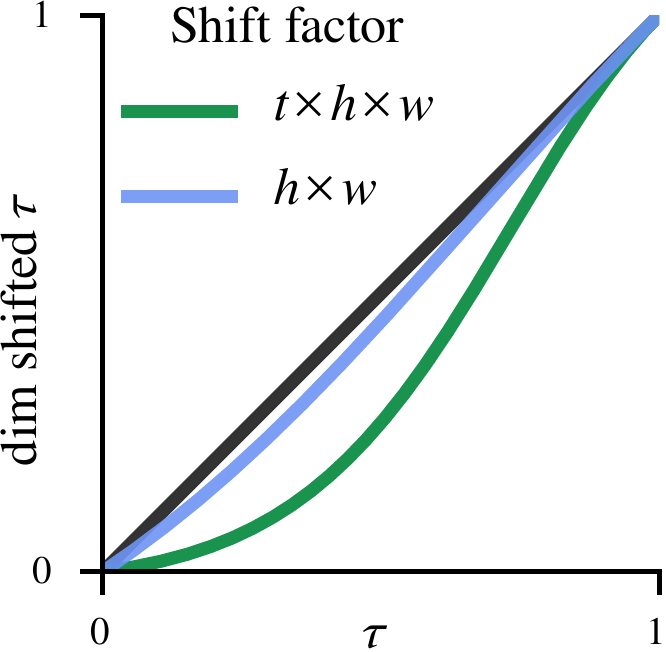}%
        }%
    \endgroup\hfill
    \subcaptionbox{Radar corruption at $\tau=0.4$ without timestep shifting.\label{fig:radar-noise-res}}{%
        \raisebox{\depth}{\resizebox*{!}{\corruptionPanelHeight}{%
            \setlength{\tabcolsep}{1pt}
            \scriptsize
            \begin{tabular}{cccc}
                \normalsize $48^2$ & \normalsize $96^2$ & \normalsize $192^2$ & \normalsize $384^2$ \\
                \includegraphics[height=\corruptionPanelHeight]{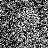} &
                \includegraphics[height=\corruptionPanelHeight]{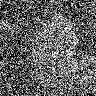} &
                \includegraphics[height=\corruptionPanelHeight]{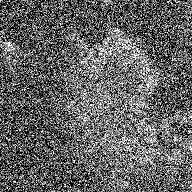} &
                \includegraphics[height=\corruptionPanelHeight]{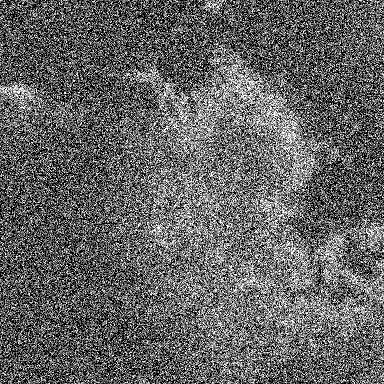}
            \end{tabular}
        }}%
    }
    \caption{\textbf{Training corruption.} (a) Framewise $\tau$ is sampled around a shared sequence-level center $\bar\tau$, retaining variation across frames while keeping context frames clean. (b) Dimensionality-aware timestep shifting increases corruption for larger spatiotemporal inputs. (c) Higher-resolution radar fields retain more structure at fixed corruption ($\tau=0.4$), motivating stronger noise.}
    \label{fig:met:corruption}
\end{figure}

\paragraph{Timestep sampling.}
Sampling the remaining timesteps $\tau^t, \forall t \notin \mathcal{C}$ independently would distribute them widely across different noise levels.
Similar to \citet{wewer2025spatial} and \citet{schusterbauer2026patchforcing} we find superior performance by concentrating them around a shared sequence-level center $\bar \tau$ while retaining variation across frames (\cref{fig:met:tbar}).
We therefore sample centers uniformly $\bar \tau \sim \mathcal{U}(0,1)$, and draw framewise timesteps as:
\begin{equation}
    \tilde{\tau}^t\sim\mathcal{N}(\bar{\tau},s_{\bar{\tau}}^2),
    \qquad
    s_{\bar{\tau}}=\min\!\left\{\sigma,\frac{\bar{\tau}}{2},
    \frac{1-\bar{\tau}}{2}\right\},
    \qquad t\notin\mathcal{C}.
    \label{eq:tbar}
\end{equation}
Draws outside $[0,1]$ are replaced by uniform draws from $[0,\bar{\tau}]$ or $[\bar{\tau},1]$ respectively.
Note that setting $\sigma=0$ recovers shared-timestep Random Masking Video Diffusion (RaMViD)~\cite{hoppe2022ramvid} used by FREUD~\cite{schusterbauer2026weatherrf-freud}.
Additionally, setting $p_U=1$ and replacing \cref{eq:tbar} with an independent uniform distribution recovers Diffusion Forcing~\cite{chen2024diffusionforcing}.
Setting $\sigma>0$ and $p_U<1$ retains asynchronous denoising as in Diffusion Forcing and preserves clean context.
We call our generalized scheme \Paradigm{} (\paradigm).

\paragraph{Dimensionality-aware corruption.}
As observed by \citet{chen2023noiseschedule} and \citet{hoogeboom2023simple}, higher-resolution inputs retain more structure at the same diffusion time $\tau$ (\cref{fig:radar-noise-res}). Stronger corruption is therefore needed at higher resolutions to maintain comparable denoising difficulty.

Prior work on pixel-space generative models~\cite{li2025back} samples diffusion timesteps from a logit-normal distribution ($\tau=\operatorname{sigmoid}(u)$, with $u\sim\mathcal{N}(\mu,\sigma_{\mathrm{LN}}^2)$) and scales the noise amplitude proportionally to the image side length relative to a $256^2$\,px reference.
We instead seek a timestep sampling scheme that explicitly accounts for both spatial and temporal resolution.

We therefore extend resolution-dependent timestep shifting~\cite{esser2024scaling} to the full sequence dimensionality $D=THW$ (\cref{fig:dim-shifted-time}), relative to a reference base resolution $D_0=T_0H_0W_0$:
\begin{equation}
    r(\tilde{\tau}^t)=\left(\frac{D_0}{D}\right)^{(1-\tilde{\tau}^t)/2},
    \qquad
    \tau^t=\frac{r(\tilde{\tau}^t)\tilde{\tau}^t}
    {1-\tilde{\tau}^t+r(\tilde{\tau}^t)\tilde{\tau}^t}.
    \label{eq:spatiotemporal_timestep_shift}
\end{equation}
For $D>D_0$, this shifts intermediate towards noise, with vanishing correction towards clean data.

In summary, we first sample a sequence-level center $\bar \tau \sim \mathcal{U}(0,1)$, and then draw framewise timesteps $\tilde \tau^t$ according to \cref{eq:tbar}. We then apply dimensionality-aware shifting to obtain the final timesteps $\tau^t$. This preserves both endpoints ($\tau=0$ and $\tau=1$), leaving clean context unchanged.
We find that this approach outperforms rescaled logit-normal sampling for spatiotemporal forecasting.

\paragraph{Prediction target.}
As discussed by \citet{li2025back}, under the manifold assumption, clean data has low-dimensional structure, unlike noise and velocity targets. This makes it easier to predict with limited network capacity in high-dimensional pixel space. Since prediction and loss parameterizations can be chosen independently, following \citet{li2025back}, we predict the clean sequence $\hat{\mathbf{x}}_1=f_\theta(\mathbf{x}_{\boldsymbol{\tau}},\boldsymbol{\tau})$ and compute the velocity-space loss (\cref{eq:fm_loss}) through the conversion:
\begin{equation}
    \mathbf{v}_\theta^t=
    \frac{\hat{\mathbf{x}}_1^t-\mathbf{x}_{\boldsymbol{\tau}}^t}{1-\tau^t},
    \qquad t\notin\mathcal{C},\quad \tau^t<1.
    \label{eq:x_to_v}
\end{equation}

\begin{wrapfigure}{r}{0.28\textwidth}
    \vspace{-0.4cm}
    \centering
    \includegraphics[width=0.9\linewidth]{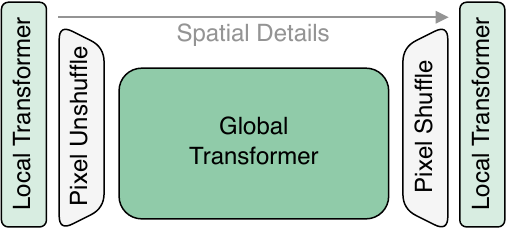}
    \caption{\textbf{HDiT backbone.}}
    \label{fig:met:HDiT}
    \vspace{-0.3cm}
\end{wrapfigure}
While our method works with standard transformers, patch size imposes a trade-off between spatial detail and computational cost. Hierarchical models address this trade-off by preserving high-resolution features while moving expensive processing to lower resolutions.
We therefore use an HDiT-style~\cite{crowson2024hdit} backbone (\cref{fig:met:HDiT}), with spatial attention in high-resolution layers and factorized spatiotemporal attention only in the lower-resolution bottleneck after an additional downsampling step.

\subsection{Simple Distributional Diffusion}
\label{sec:distributional}

\begin{figure}[h]
    \centering
    \includegraphics[width=0.85\linewidth]{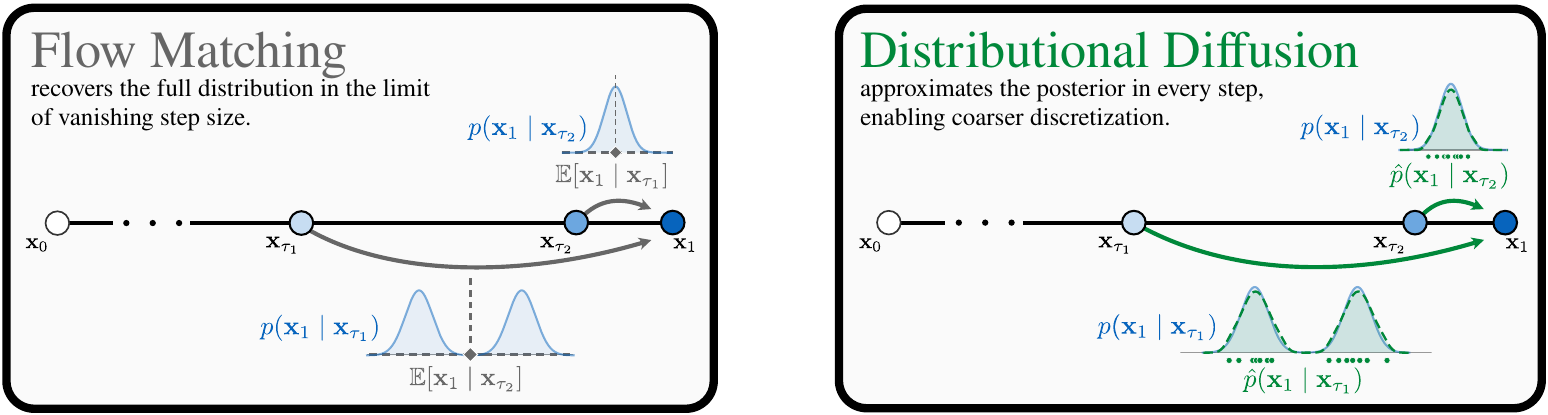}
    \caption{\textbf{Distributional Diffusion Models.} Distributional Diffusion Models~\cite{bortoli2025distributional} learn the posterior distribution, enabling sampling with coarser diffusion time discretization.}
    \label{fig:met:DDM}
\end{figure}

Flow matching models trained with \cref{eq:fm_loss} approximate the conditional mean velocity $\mathbf{v}^\star=\mathbb{E}[\mathbf{x}_1-\mathbf{x}_0\mid\mathbf{x}_{\boldsymbol{\tau}},\boldsymbol{\tau}]$. In the limit of vanishing step size, integrating this field recovers the data distribution. However, this no longer holds under coarse discretization with few denoising steps~\cite{bortoli2025distributional} (\cref{fig:met:DDM}).
Instead, Distributional Diffusion Models (DDMs)~\cite{bortoli2025distributional} learn the
conditional posterior velocity $p(\mathbf{v}|\mathbf{x}_{\boldsymbol{\tau}},\boldsymbol{\tau})$.
For distributional supervision, a scoring rule over an ensemble replaces squared error regression.
To produce an ensemble prediction during training, the model receives auxiliary noise $\xi\sim\mathcal{N}(0,\mathbf{I})$.

For each fixed $(\mathbf{x}_{\boldsymbol{\tau}},\boldsymbol{\tau})$, drawing $M\geq2$ independent auxiliary noise samples produces a population of predictions (particles), $V=\{\mathbf{v}_i\}_{i=1}^{M}$, where $\mathbf{v}_i=\mathbf{v}_\theta(\mathbf{x}_{\boldsymbol{\tau}},\boldsymbol{\tau},\xi_i)$. In our model, these velocities are obtained from stochastic clean-data predictions through \cref{eq:x_to_v}.

\paragraph{From energy scores to a simple objective.}
For the particle population $V$ and target velocity $\mathbf{v}$, the generalized energy score~\citep{bortoli2025distributional} loss is
\begin{equation}
    \mathcal{L}_{\lambda,\beta}
    =\underbrace{\frac{1}{M}\sum_i\|\mathbf{v}-\mathbf{v}_i\|_2^\beta}_{\text{confinement}}
    -\frac{\lambda}{2}\,
    \underbrace{\frac{1}{M(M-1)}\sum_{i\ne j}
    \|\mathbf{v}_i-\mathbf{v}_j\|_2^\beta}_{\text{interaction}}.
    \label{eq:ddm_loss}
\end{equation}
The confinement term encourages agreement with the target, while the interaction term promotes diversity. The parameters $\lambda\in[0,1]$ and $\beta\in(0,2]$ control interaction strength and the distance exponent, respectively. For $\lambda=1$ and $0<\beta<2$, the score is strictly proper: its expectation is uniquely minimized by the target posterior. Setting $(\lambda,\beta)=(0,2)$ recovers squared-error loss.
iDDM~\cite{martorella2026iddm} extends DDM by varying $\lambda$ and $\beta$ with diffusion time to favor distributional supervision at high noise levels and regression near clean data.

We aim to build a simplified distributional objective that does not require $\lambda$ and $\beta$ schedule tuning. Further, radar-space prediction enables the direct use of CRPS, the proper scoring rule for scalar predictive distributions commonly used in evaluation of probabilistic forecasting methods.
We therefore linearly interpolate two fixed losses:
\begin{equation}
    \boxed{\mathcal{L}_{\mathrm{simple}}(\tau^t)
    =(1-\tau^t)\,\mathrm{CRPS}+\tau^t\,\mathrm{MSE}.}
    \label{eq:simple_loss}
\end{equation}
CRPS supervises scalar conditional marginals at each pixel, while particlewise MSE favors conditional means.
The timestep-dependent interpolation transitions from marginal distributional supervision towards regression as the model approaches clean data. This is consistent with the observation that the conditional distribution becomes increasingly concentrated as $\tau\to1$~\cite{biroli2024dynamical, martorella2026iddm}.
Thus, our objective for \Objective~(\objective) is an easy-to-use drop-in replacement for squared error supervision which achieves superior performance to Flow Matching and iDDM in few- and many-step settings.
We show the full training in \cref{alg:training}.

\paragraph{Efficient training and sampling.}
Standard flow matching can be viewed as single-particle training: each noisy input produces one prediction. Computing the interaction term requires $M\geq2$ predictions for the same input. Following iDDM~\cite{martorella2026iddm}, we avoid repeating the full network computation by sharing early transformer layer computations, then making $M$ copies. To construct the particles, we concatenate independent auxiliary noise $\xi_i$ to each copy along the token feature dimension, after modulating the noise with a learned $\tau$-dependent gate. The remaining transformer layers process each particle separately.
At inference, we use standard Euler sampling and keep context frames fixed throughout the denoising process. %

%% file: sec/4_experimentsB.tex
We use \emph{SEVIR}~\cite{veillette2020SEVIR} and \emph{MeteoNet}~\cite{larvor2020meteonet} to evaluate our method. 
After ablating model components on the \emph{SEVIR}~\cite{veillette2020SEVIR} benchmark, we show generalization to the \emph{MeteoNet}~\cite{larvor2020meteonet} data.
Following standard evaluation~\cite{gong2024cascast, schusterbauer2026weatherrf-freud}, we report Continuous Ranked Probability Score (CRPS), Structural Similarity Index (SSIM), Heidke Skill Score (HSS), and Critical Success Index (CSI) over 6 thresholds. Additionally, we compute rank histograms and reliability index (RI)~\cite{dellemonache2006Probabilistic, wilks2019Indices} for calibration. We use 50 sampling steps and evaluate 512 samples for ablations. Metric and evaluation details are found in Appendix \cref{sec:sup:metrics}.
Implementation details for our method and training procedure are found in Appendix \cref{sec:sup:implementation_details}.

\subsection{Building JWS}\label{sec:exp:building}

The following experiments progressively build towards \ours~starting from a naive pixel-space image diffusion baseline~\cite{li2025back}. We use small models (24M params) unless noted otherwise.
Additional ablations of hyperparameters (e.g., patch size, ensemble expansion, and training steps) are found in Appendix \cref{sec:sup:extended_ablations}.

\input{tabs/4_exp/ablation_tsample_paradigm}

\input{tabs/4_exp/sevir_objective_comparison}

\paragraph{Diffusion timestep sampling.}
\cref{tab:exp:t-sampling} compares different $p(\tau)$ for training. We follow \citet{schusterbauer2026weatherrf-freud} and instantiate \paradigm{} with $\sigma=0$ recovering RaMViD~\cite{hoppe2022ramvid}. Default Logit-normal sampling (LN) fails, LN with noise scaling achieves the best CSI, yet rescaled uniform achieves superior CRPS and RI. We therefore use rescaled Uniform sampling.
In \cref{tab:exp:paradigm}, \paradigm{} instantiations with $p_U < 1$, which preserve clean conditioning, outperform naive Diffusion Forcing.
Setting $\sigma = 0.1$, i.e., sampling correlated timesteps for predicted frames, achieves strong performance while maintaining flexibility.

\paragraph{Architecture.}
\cref{tab:exp:architecture_comparison} shows that  \paradigm{} enables effective pixel-space nowcasting with standard transformers, i.e.\ without specialized architectural modifications. Hierarchical models further improve performance, while factorized spatiotemporal attention reduces computational cost.

\begin{figure}[tb]
    \centering
    \includegraphics[width=\linewidth]{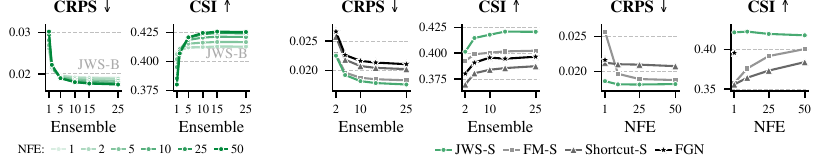}
    \caption{\textbf{Inference time scaling.} \textit{Left:} \ours~scales with ensemble size and NFEs. \textit{Right:} \ours~outperforms Flow Matching, Shortcut models, and FGN in matched comparisons using a single NFE.}
    \label{fig:exp:test_time_scaling}
\end{figure}

\begin{samepage}
\paragraph{Distributional objective.}
\cref{tab:exp:objective_nfe} compares squared error, CRPS, iDDM, and \objective~objectives.
\objective{} achieves strong performance for low and high NFE settings, while default squared-error Flow Matching requires many iterations and training only with CRPS deteriorates for high NFE.
\cref{fig:exp:test_time_scaling} highlights inference time scaling.
Further, \ours{} outperforms alternative methods even with a single function evaluation.
In summary, \objective{} achieves strong results with limited compute and scalability to larger inference budgets.
\end{samepage}

\subsection{System-level Weather Nowcasting}

\input{tabs/4_exp/sevir_nowcasting}

\input{figs/4_exp/merged_qual_comparison_latency}

\input{tabs/4_exp/meteonet_xx_nowcasting}

\paragraph{SOTA nowcasting.} 
In \cref{tab:exp:sevir_nowcasting} \ours~achieves better CRPS, SSIM, and RI on SEVIR than previous methods without classifier-free guidance (CFG). Using guidance, we additionally achieve superior HSS and perform second-best on CSI.
We achieve state-of-the-art with a simplified, efficient pipeline. \ours-T performs competitively with much larger models and enables probabilistic nowcasting with a small footprint.
\cref{fig:exp:qualitative_comp} qualitatively shows sharp realistic forecasts over lead time. Additional qualitative comparisons are found in Appendix \cref{sec:sup:qualitative_comparisoms}.
\cref{tab:exp:meteonet} evaluates \ours~on MeteoNet. \ours{} achieves the best reported CRPS and SSIM. Guidance further improves HSS and CSI, achieving the best reported scores for these metrics.
The 9M parameter \ours-T model outperforms previous methods in CRPS.
\cref{fig:exp:latency} shows that even with 50 NFE \ours~provides forecasts faster than CasCast or FREUD. One-step \ours-B produces a 10-member ensemble forecast in less than a second and performs competitively with CasCast and FREUD models. Appendix \cref{sec:sup:qualitative_nfe} qualitatively shows that \ours{} produces strong forecasts with single- or few-step sampling.

\begin{figure}[tb]
    \centering
    \includegraphics[width=\linewidth]{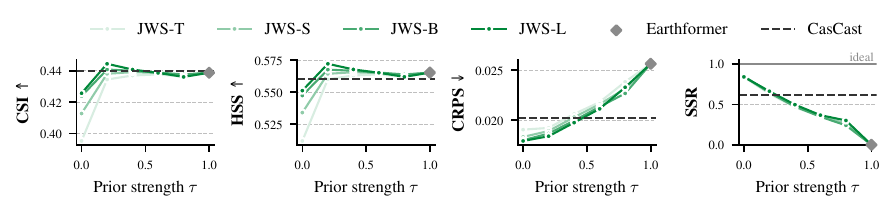}
    \vspace{-2em}
    \caption{\textbf{Deterministic conditioning.} Conditioning on a deterministic forecast as explored by CasCast~\cite{gong2024cascast} improves CSI and HSS, but consistently degrades calibration.}
    \label{fig:sup:detcond}
\end{figure}

\paragraph{Zero-shot deterministic conditioning.}
\label{p:det-cond}
Our training paradigm enables conditioning on deterministic forecasts, similar to CasCast~\cite{gong2024cascast}, but without requiring such priors during training.
Following \citet{schusterbauer2026weatherrf-freud}, we initialize denoising from noised EarthFormer~\cite{gao2022earthformer} predictions.
Hereby, the noising timestep $\tau$ controls the strength of the deterministic prior: larger $\tau$ preserves more of the forecast, while smaller $\tau$ replaces it increasingly with noise and thus gives the generative model more freedom to deviate from it.
As shown in \cref{fig:sup:detcond}, weak deterministic priors improve CSI and HSS up to $\tau=0.2$, but any deterministic conditioning consistently deteriorates calibration.

Beyond global prior strength, \paradigm{} allows assigning different diffusion timesteps to individual frames, enabling time-dependent confidence in the deterministic forecast.
Using small \ours{} models, \cref{fig:sup:confidence_forcing} shows that cleaner priors for early frames and noisier priors for later frames improve CRPS, reliability, and CSI compared to uniform prior strength.
The RaMViD baseline, which does not support such framewise confidence, underperforms particularly in calibration.

\begin{figure}[tb]
    \centering
    \includegraphics[width=\linewidth]{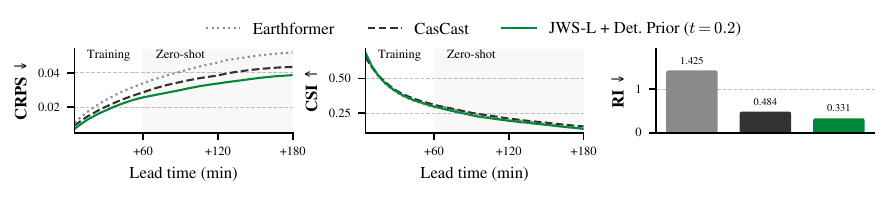}
    \caption{\textbf{Zero-shot evaluation on SEVIR-Long.} \ours{} shows superior zero-shot extrapolation performance to 180\,min into the future ($36$ frames), while also better calibrated.}
    \label{fig:exp:sevir-long}
\end{figure}

\begin{samepage}
\paragraph{Zero-shot sequence-length extrapolation.}
We evaluate zero-shot lead-time extrapolation, extending forecasts from 12 to 36 predicted frames without additional training. To enable a fair comparison with CasCast~\cite{gong2024cascast}, which relies on deterministic conditioning and to stabilize forecasts beyond the training horizon, we leverage our training paradigm’s ability to condition on deterministic priors at inference time.
Similar to CasCast, we condition on EarthFormer~\cite{gao2022earthformer} predictions.
\cref{fig:exp:sevir-long} shows that \ours{} achieves superior CRPS and calibration (RI) while remaining competitive in CSI. Moreover, \ours{} substantially improves over the deterministic EarthFormer prior, demonstrating the benefit of generative refinement for long-horizon forecasting.
\end{samepage}

\input{figs/4_exp/merged_qual_ensemble_calibration}

\begin{samepage}
\paragraph{Calibrated uncertainty estimates.}
In \cref{fig:exp:calibration} \ours~rank histograms are flatter than for FREUD and CasCast. Calibration remains superior over lead time.
In \cref{fig:exp:qualitative_ens} ensembles at $+60\,\text{min}$ maintain accurate global structure, with local variability. Heatmaps reveal uncertainty in high precipitation regions, aligned with chaotic dynamics. Appendix \cref{sec:sup:qualitative_uncertainty} shows more qualitative ensembles.
\end{samepage}

%% file: tabs/4_exp/ablation_tsample_paradigm.tex
\begin{table*}[t]
\centering
\begin{minipage}[t]{0.32\textwidth}
    \centering
    \caption{\textbf{$\tau$-sampling.} Uniform $p(\tau)$ with $T \times H \times W$ shift achieves superior performance.}
    \label{tab:exp:t-sampling}
    \setlength{\tabcolsep}{3pt}
    \resizebox{\linewidth}{!}{%
    \begin{tabular}{lccc}
        \toprule
        \textbf{T-sampling} & \textbf{CRPS} $\downarrow$ & \textbf{CSI} $\uparrow$ & \textbf{RI} $\downarrow$ \\
        \midrule
        LN
        & 0.0232 & 0.3081 & 0.3100 \\
        LN, noise $\times 1.5$
        & 0.0211 & \textbf{0.3698} & 0.8575 \\
        $\mathcal{U}$, $H{\times}W$
        & 0.0210 & 0.3285 & 0.1109 \\
        $\mathcal{U}$, $T{\times}H{\times}W$
        & \textbf{0.0203} & 0.3467 & \textbf{0.1066} \\
        \bottomrule
    \end{tabular}
    }
\end{minipage}
\hfill
\begin{minipage}[t]{0.32\textwidth}
    \centering
    \caption{\textbf{Paradigm.} We compare \paradigm{} to RaMViD and Diffusion Forcing.}
    \label{tab:exp:paradigm}
    \setlength{\tabcolsep}{3pt}
    \resizebox{\linewidth}{!}{%
    \begin{tabular}{lccccc}
        \toprule
        \textbf{Paradigm} & \textbf{CRPS} $\downarrow$ & \textbf{CSI} $\uparrow$ & \textbf{RI} $\downarrow$ \\
        \midrule
        Forcing
        & 0.0317 & 0.2280 & 0.5301 \\
        RaMViD
        & \textbf{0.0203} & 0.3467 & 0.1066 \\
        \paradigm{} uniform
        & 0.0230 & 0.3380 & 0.2870 \\
        \paradigm{} $\sigma = 0.1$
        & 0.0205 & \textbf{0.3545} & \textbf{0.0938} \\
        \bottomrule
    \end{tabular}
    }
\end{minipage}
\hfill
\begin{minipage}[t]{0.32\textwidth}
    \centering
    \footnotesize
    \setlength{\tabcolsep}{5.1pt}
    \newcommand{\cfgresult}[1]{\textcolor{black!60}{#1}}
    \caption{\textbf{Architecture.} Hierarchical models with factorized attention outperform JiT.}
    \label{tab:exp:architecture_comparison}
    
    \resizebox{\linewidth}{!}{%
    \begin{tabular}{lccccc}
        \toprule
        
        \textbf{Model {\tiny B-scale}} &
        \textbf{CRPS}$\downarrow$ &
        \textbf{CSI}$\uparrow$ &
        \textbf{ms\,/\,NFE} $\downarrow$
        \\
        
        \midrule
        
        JiT
        & 0.0193
        & 0.3782
        & 6.55
        \\
        
        Fact. JiT
        & 0.0195
        & 0.3804
        & \textbf{4.76}
        \\
        
        Hierarch.
        & 0.0186
        & 0.3895
        & 6.05
        \\
        
        Fact. Hierarch.
        & \textbf{0.0182}
        & \textbf{0.4275}
        & 5.80
        \\
        
        \bottomrule
    \end{tabular}%
    }
\end{minipage}
\end{table*}

%% file: tabs/4_exp/sevir_objective_comparison.tex
\begin{table}[t]
    \centering
    \caption{\textbf{Training objective.} Distributional training~\cite{bortoli2025distributional} enables few-step inference, but CRPS-only degrades at high NFE. \objective{} outperforms iDDM~\cite{martorella2026iddm}.}
    \label{tab:exp:objective_nfe}
    \footnotesize
    \setlength{\tabcolsep}{4pt}

    \begin{minipage}[t]{0.48\linewidth}
        \centering
        \subcaption{CRPS $\downarrow$}
        \label{tab:objective_nfe:crps}
        \resizebox{0.66\linewidth}{!}{%
            \begin{tabular}{@{}lccc@{}}
                \toprule
                \textbf{Method} & \textbf{5 NFE} & \textbf{10 NFE} & \textbf{50 NFE} \\
                \midrule
                MSE  & 0.0208 & 0.0197 & 0.0188 \\
                CRPS & \textbf{0.0182} & 0.0183 & 0.0189 \\
                iDDM & 0.0188 & 0.0185 & 0.0183 \\
                sDDM & 0.0183 & \textbf{0.0182} & \textbf{0.0178} \\
                \bottomrule
            \end{tabular}%
        }
    \end{minipage}\hfill%
    \begin{minipage}[t]{0.48\linewidth}
        \centering
        \subcaption{CSI $\uparrow$}
        \label{tab:objective_nfe:csi}
        \resizebox{0.66\linewidth}{!}{%
            \begin{tabular}{@{}lccc@{}}
                \toprule
                \textbf{Method} & \textbf{5 NFE} & \textbf{10 NFE} & \textbf{50 NFE} \\
                \midrule
                MSE  & 0.3654 & 0.3752 & 0.4004 \\
                CRPS & 0.4177 & 0.4170 & 0.4114 \\
                iDDM & 0.4088 & 0.4109 & 0.4104 \\
                sDDM & \textbf{0.4219} & \textbf{0.4219} & \textbf{0.4116} \\
                \bottomrule
            \end{tabular}%
        }
    \end{minipage}
\end{table}

%% file: tabs/4_exp/sevir_nowcasting.tex
\begin{table}[t]
    \newcommand{\cmark}{\ding{51}}%
    \newcommand{\xmark}{\ding{55}}%
    \newcommand{\vcat}[1]{\rotatebox{90}{\scriptsize#1}}%
    \newcommand{\cfgresult}[1]{\textcolor{black!60}{#1}}%
    \caption{\textbf{SOTA nowcasting on SEVIR.} \ours~achieves superior CRPS, SSIM, HSS, and RI on SEVIR. $\dag$ model trained at $128^2$ for better results. $\star$ metric computed with our pipeline.}
    \label{tab:exp:sevir_nowcasting}
    \centering
    \footnotesize
    \setlength{\tabcolsep}{5.1pt}
    \centering
    \resizebox{\linewidth}{!}{%
    \begin{tabular}{
    clcccccccc
    }
    
    \toprule
    
    \multicolumn{2}{c}{\textbf{Method}} &
    \textbf{Stages} &
    \textbf{Param}&
    \textbf{CRPS}$\downarrow$ &
    \textbf{SSIM}$\uparrow$ &
    \textbf{HSS}$\uparrow$ &
    \textbf{CSI}$\uparrow$ &
    \textbf{RI}$^\star$ $\downarrow$
    \\
    
    \midrule

    \multirow{5}{*}{\vcat{Deterministic}}
    & ConvLSTM~\citep{shi15:ConvLSTM}~\ven{NeurIPS '15}
    & 1
    & 14M
    & 0.0264        %
    & 0.7749        %
    & 0.5232        %
    & 0.4102        %
    & --
    \\
    & PredRNN~\citep{wang2017predrnn}~\ven{TPAMI '22}
    & 1
    & 47M
    & 0.0271        %
    & 0.7497        %
    & 0.5192        %
    & 0.4045        %
    & --
    \\
    & PhyDNet~\citep{guen2020disentangling}~\ven{CVPR '20}
    & 1
    & 14M
    & 0.0253        %
    & 0.7649        %
    & 0.5311        %
    & 0.4198        %
    & --
    \\
    & SimVP~\citep{gao2022simvp}~\ven{CVPR '22}
    & 1
    & 16M
    & 0.0259        %
    & 0.7772        %
    & 0.5280        %
    & 0.4153        %
    & --
    \\
    & EarthFormer~\citep{gao2022earthformer}~\ven{NeurIPS '22}
    & 1
    & 9M
    & 0.0251        %
    & 0.7756        %
    & 0.5411        %
    & 0.4310        %
    & --
    \\
    
    \arrayrulecolor{gray!50!white}
    \midrule
    \arrayrulecolor{black}

    \multirow{5}{*}{\vcat{Generative}}
    & NowcastNet~\citep{zhang2023skilful}~\ven{Nature '23}
    & 1
    & 35M
    & 0.0283        %
    & 0.5696        %
    & 0.5365        %
    & 0.4152        %
    & --
    \\
    & PreDiff$^\dag$~\citep{gao2023prediff}~\ven{NeurIPS '23}
    & 2
    & 105M
    & 0.0202        %
    & 0.7648        %
    & 0.4914        %
    & 0.3875        %
    & --
    \\
    & CasCast~\citep{gong2024cascast}~\ven{ICML '24}
    & 3
    & 402M
    & 0.0202 %
    & 0.7797 %
    & 0.5602 %
    & 0.4401 %
    & 0.3124
    \\
    & FlowCast~\cite{ribeiro2026flowcast}~\ven{ICLR '26}
    & 2
    & 160M
    & 0.0182 %
    & - %
    & 0.5863 %
    & \textbf{0.4651} %
    & 0.6835
    \\
    & FREUD~\citep{schusterbauer2026weatherrf-freud}~\ven{CVPR '26}
    & 2
    & 521M
    & 0.0190 %
    & 0.7841 %
    & 0.5011 %
    & 0.3864 %
    & 0.1355
    \\

    \arrayrulecolor{gray!50!white}
    \midrule
    \arrayrulecolor{black}

    & JWS-T/32 (\textit{ours})
    & 1
    & 9M
    & 0.0184 %
    & 0.8016 %
    & 0.5132 %
    & 0.3948 %
    & \textbf{0.0949}
    \\
    
    & JWS-S/32 (\textit{ours})
    & 1
    & 24M
    & 0.0178 %
    & 0.8057 %
    & 0.5327 %
    & 0.4116 %
    & 0.0966
    \\

    & JWS-B/32 (\textit{ours})
    & 1
    & 94M
    & 0.0176 %
    & 0.8092 %
    & 0.5408 %
    & 0.4180 %
    & 0.0956
    \\
    
    & JWS-L/32 (\textit{ours})   
    & 1
    & 340M
    & \textbf{0.0175} %
    & \textbf{0.8122} %
    & 0.5428 %
    & 0.4183 %
    & 0.1908
    \\

    & \cfgresult{\rotatebox[origin=c]{180}{$\Lsh$} with CFG = 1.5}
    & \cfgresult{1}
    & \cfgresult{340M}
    & \cfgresult{0.0177}
    & \cfgresult{0.8023}
    & \textbf{\cfgresult{0.5894}}
    & \cfgresult{0.4581}
    & \cfgresult{0.1815}
    \\
    \bottomrule
    \end{tabular}
    }
\end{table}

%% file: figs/4_exp/merged_qual_comparison_latency.tex

\begin{figure}[t]
    \centering

    \begin{minipage}[t]{0.63\textwidth}
        \vspace{0pt}
        \centering
        \setlength{\tabcolsep}{0.5pt}
        \newcommand{\imagesize}{0.15\textwidth}
        \newcommand{\imgpath}{figs/X_supp/qualitative_comparisons/idx2286}
        \newcommand{\gridimage}[3]{\includegraphics[width=\imagesize]{\imgpath/#1_#2_#3.png}}
        \newcommand{\rowlabel}[1]{%
            \raisebox{%
                \dimexpr\imagesize/2-\height/2+\depth/2\relax
            }[0pt][0pt]{%
                \rotatebox{90}{#1}%
            }%
        }
        \newcommand{\gridimgrow}[2]{\gridimage{#1}{2}{#2}
            & \gridimage{#1}{4}{#2}
            & \gridimage{#1}{6}{#2}
            & \gridimage{#1}{8}{#2}
            & \gridimage{#1}{10}{#2}
            & \gridimage{#1}{12}{#2}}
        \begin{tabular}{ccccccc}
            & +10\,min & +20\,min & +30\,min & +40\,min & +50\,min & +60\,min \\
            \rowlabel{\footnotesize CasCast} & \gridimgrow{cascast}{1} \\
            \rowlabel{\footnotesize FREUD} & \gridimgrow{freud}{4} \\
            \rowlabel{\footnotesize JWS {\tiny (Ours)}} & \gridimgrow{ours}{4} \\
            \rowlabel{\footnotesize GT} & \gridimgrow{gt}{1}
        \end{tabular}

        \caption{\textbf{Qualitative forecast comparison.}
            \ours~produces sharp forecasts that align with the ground truth and remains more realistic over leadtime than FREUD and CasCast.}
        \label{fig:exp:qualitative_comp}
    \end{minipage}%
    \hfill
    \begin{minipage}[t]{0.35\textwidth}
        \vspace{0pt}
        \centering

        \begin{minipage}[t]{\linewidth}
            \vspace{0pt}
            \centering
            \includegraphics[width=\linewidth]{%
                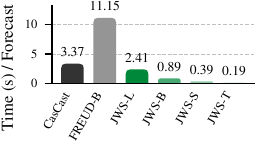}
        \end{minipage}
        \begin{minipage}[t]{\linewidth}
            \vspace{0pt}
            \centering
            \includegraphics[width=\linewidth]{%
                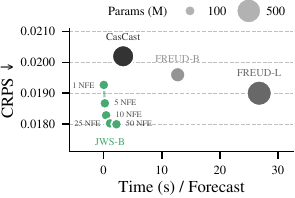}
        \end{minipage}
        \caption{\textbf{Latency.} \ours{} enables faster inference at matched (20) NFE and strong few-step results.}
        \label{fig:exp:latency}
    \end{minipage}
\end{figure}

%% file: tabs/4_exp/meteonet_xx_nowcasting.tex
\begin{table}[t]
\centering
\caption{\textbf{MeteoNet nowcasting.} \ours~achieves state-of-the-art nowcasting on MeteoNet. \ours-T outperforms $40\times$ larger models in CRPS. $\star$ metric computed with our pipeline.}
\label{tab:exp:meteonet}
\setlength{\tabcolsep}{5.1pt}
\newcommand{\vcat}[1]{\rotatebox{90}{\scriptsize#1}}%
\newcommand{\cfgresult}[1]{\textcolor{black!60}{#1}}%
\resizebox{\linewidth}{!}{%
    \begin{tabular}{clccccccc}
        \toprule
        \multicolumn{2}{c}{\textbf{Method}} &
        \textbf{Stages} &
        \textbf{Param} &
        \textbf{CRPS}$\downarrow$ &
        \textbf{SSIM}$\uparrow$ &
        \textbf{HSS}$\uparrow$ &
        \textbf{CSI}$\uparrow$ &
        \textbf{RI$^\star$}$\downarrow$
        \\
        \midrule
        
        \vcat{D.}
        & EarthFormer~\citep{gao2022earthformer}~\ven{NeurIPS '22}
        & 1
        & 9M
        & 0.0224
        & --
        & --
        & 0.2831
        & --
        \\
        
        \arrayrulecolor{gray!50!white}
        \midrule
        \arrayrulecolor{black}

        \multirow{4}{*}{\vcat{Generative}}
        & NowcastNet~\citep{zhang2023skilful}~\ven{Nature '23}
        & 1
        & 35M
        & 0.0277
        & --
        & --
        & 0.2955
        & --
        \\
        
        & PreDiff~\citep{gao2023prediff}~\ven{NeurIPS '23}
        & 2
        & 105M
        & 0.0197
        & --
        & --
        & 0.2546
        & --
        \\
        
        & CasCast~\citep{gong2024cascast}~\ven{ICML '24}
        & 3
        & 402M
        & 0.0180 %
        & -- %
        & -- %
        & 0.3156 %
        & --
        \\
        
        & FREUD~\citep{schusterbauer2026weatherrf-freud}~\ven{CVPR '26}
        & 2
        & 521M
        & 0.0193 %
        & 0.7312 %
        & 0.2082 %
        & 0.1417 %
        & --
        \\
        
        \arrayrulecolor{gray!50!white}
        \midrule
        \arrayrulecolor{black}

        & JWS-T/32 (\textit{ours})
        & 1
        & 9M
        & 0.0143 %
        & 0.7887 %
        & 0.4026 %
        & 0.2851 %
        & 0.2376
        \\
        
        & JWS-S/32 (\textit{ours})
        & 1
        & 24M
        & 0.0141 %
        & 0.7917 %
        & 0.4079 %
        & 0.2891 %
        & 0.3100
        \\
        
        & JWS-B/32 (\textit{ours})
        & 1
        & 94M
        & 0.0144 %
        & 0.7967 %
        & 0.4206 %
        & 0.2955 %
        & 0.3858
        \\
        
        & JWS-L/32 (\textit{ours})
        & 1
        & 340M
        & \textbf{0.0138} %
        & \textbf{0.7974} %
        & 0.4186 %
        & 0.2959 %
        & \textbf{0.0773}
        \\

        & \cfgresult{\rotatebox[origin=c]{180}{$\Lsh$} with CFG = 1.5}
        & \cfgresult{1}
        & \cfgresult{340M}
        & \cfgresult{0.0141} %
        & \cfgresult{0.7904} %
        & \cfgresult{\textbf{0.4816}} %
        & \cfgresult{\textbf{0.3448}} %
        & \cfgresult{0.5719}
        \\
        
        \bottomrule
    \end{tabular}%
}
\end{table}

%% file: figs/4_exp/merged_qual_ensemble_calibration.tex
\begin{figure}[tb]
    \centering
    \begin{minipage}[t]{0.63\textwidth}
        \vspace{0pt}
        \centering
        \setlength{\tabcolsep}{0.1pt}
        \newcommand{\imagesize}{0.135\textwidth}
        \newcommand{\imgpath}{figs/X_supp/qualitative_ensembles}
        \newcommand{\gridimage}[4]{\includegraphics[width=\imagesize]{\imgpath/idx#1/#2_#3_#4.png}}
        \newcommand{\gridimagerow}[1]{\gridimage{#1}{gt}{12}{1} & \gridimage{#1}{ens}{12}{1} & \gridimage{#1}{ens}{12}{2} & \gridimage{#1}{ens}{12}{3} & \gridimage{#1}{ens}{12}{4} & \gridimage{#1}{ens}{12}{5} & \gridimage{#1}{var}{12}{1} & \colorbar{#1}{12} \\}
        \newcommand{\colorbar}[2]{\includegraphics[height=0.14\textwidth]{\imgpath/idx#1/colorbar_#2.pdf}}
        \begin{tabular}{ccccccc@{\hspace{3pt}}l}
            GT & Ens 1 & Ens 2 & Ens 3 & Ens 4 & Ens 5 & Var & \\
            \gridimagerow{1679}
            \gridimagerow{12135}
            \gridimagerow{6912}
        \end{tabular}
        \caption{\textbf{Qualitative forecast ensemble.} At $+60\,\text{min}$ ensemble variance is aligned with precipitation intensity.}
        \label{fig:exp:qualitative_ens}
    \end{minipage}%
    \hfill
    \begin{minipage}[t]{0.335\textwidth}
        \vspace{0pt}
        \centering
        \begin{subfigure}[t]{\linewidth}
            \vspace{0pt}
            \centering
            \includegraphics[width=\linewidth]{%
                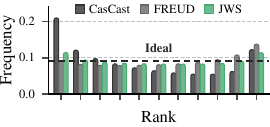}
        \end{subfigure}
        \begin{subfigure}[t]{\linewidth}
            \vspace{0pt}
            \centering
            \includegraphics[width=\linewidth]{%
                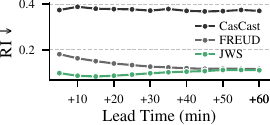}
        \end{subfigure}
        \caption{\textbf{Calibration.} \ours{} produces flatter rank histograms.}
        \label{fig:exp:calibration}
    \end{minipage}
\end{figure}

%% file: sec/5_conclusion.tex
Precipitation nowcasting requires rapid updates and calibrated uncertainty.
We present \ours{}, a simple end-to-end generative nowcasting method. \ours~does not rely on compression models. Thus, ensembling quantifies uncertainty directly in radar-space. Training with a distributional objective unlocks few-step sampling at inference time, drastically reducing latency.
Our approach achieves state-of-the-art probabilistic nowcasting performance on the SEVIR and MeteoNet benchmarks.
Our approach is scalable with respect to the model size and the available computational budget at inference time, while efficient settings perform competitively.
By greatly simplifying training and inference, we hope that \ours~can be easily adopted by meteorological practitioners.

%% file: sec/X_suppl.tex
\clearpage

\FloatBarrier

\appendix
\begin{center}
    {\LARGE\sc Supplementary Materials}
\end{center}
\vspace{1em}

\section{Implementation Details}\label{sec:sup:implementation_details}

We summarize architecture and training hyperparameters in \cref{tab:sup:hyperparams}.

\input{tabs/X_sup/hyperparams}

\begin{algorithm}[t]
\caption{JWS training step}
\label{alg:training}
\begin{lstlisting}[style=paperpython]
# x1: data of shape (bs, f, h, w, c)
# t: per-frame diffusion timesteps (*@\codecref{eq:tbar}@*), conditioning frames t=1
# K: number of particles during training
# net: model returning K predictions per batch element
# timestep dimensions broadcast implicitly

# shift framewise timesteps (*@\codecref{eq:spatiotemporal_timestep_shift}@*)
s = (256**2 / (f * h * w)) ** ((1 - t) / 2)
t = s * t / (1 - t + s * t)

# interpolate between noise and clean data (*@\codecref{eq:fm_interpolant}@*)
x0 = randn_like(x1)
xt = t * x1 + (1 - t) * x0

# predict an ensemble in pixel space
x1_pred = net(xt, t)  # (bs, K, f, h, w, c)

# convert predictions and target to velocity space (*@\codecref{eq:x_to_v}@*)
v = (x1 - xt) / (1 - t)
v_pred = (x1_pred - xt[:, None]) / (1 - t)

# confinement term: attract predictions to the target (*@\codecref{eq:ddm_loss}@*)
error = v_pred - v[:, None]
confinement = mean(abs(error), dim=1)

# interaction term: encourage ensemble diversity (*@\codecref{eq:ddm_loss}@*)
i, j = upper_triangle_indices(K, exclude_diagonal=True)
pairs = abs(v_pred[:, i] - v_pred[:, j])
interaction = sum(pairs, dim=1) / K**2

# compute L_simple (*@\codecref{eq:simple_loss}@*)
crps = confinement - interaction
mse = mean(error**2, dim=1)
loss = (1 - t) * crps + t * mse
\end{lstlisting}
\end{algorithm}

\paragraph{Architecture.}
We build \ours{} using standard Llama-style Transformers. Model variants across scales differ only in the number of layers and their hidden dimensionality, resulting in four model sizes ranging from 9M to 340M parameters.
All hierarchical models use an initial patchification of $1 \times 16 \times 16$ matching \citet{li2025back}. We follow \citet{schusterbauer2026weatherrf-freud} and use neighborhood attention~\cite{hassani2023Neighborhood} for the high-resolution transformer layers, restricting attention to a $5 \times 5$ region within each frame. All models use two high resolution layers before down- and two layers after upsampling (four in total). The remaining layers are applied after additional $1 \times 2 \times 2$ patchification ($1 \times 32 \times 32$ in total). Bottleneck layers attend to the full spatiotemporal grid but use interleaved factorized attention, interleaving only-spatial and only-temporal attention blocks.
All models use Adaptive RMSNorm~\cite{zhang2019root, peebles2023Scalable, xu2019understanding} for normalization and diffusion-time conditioning. Further, all models use SwiGLU~\cite{shazeer2020glu} activations and 3D Axial Rotary Positional Embeddings~\cite{su2023RoFormer, heo2024Rotary}.

\begin{wrapfigure}{r}{0.4\textwidth}
    \vspace{-0.4cm}
    \centering
    \includegraphics[width=0.98\linewidth]{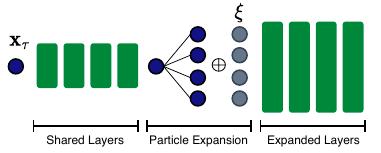}
    \caption{\textbf{$\xi$ conditioning.}}
    \label{fig:supp:xi-cond}
    \vspace{-0.3cm}
\end{wrapfigure}
All variants expand the ensemble for distributional training before the last bottleneck layer. All following layers (bottleneck and high-resolution) receive noise conditioning. The auxiliary noise $\xi$ is sampled with dimensionality 256 and concatenated to the residual stream after modulation by a diffusion time-dependent gate following~\cite{martorella2026iddm}. \cref{fig:supp:xi-cond} shows the $\xi$-conditioning schematically for a simple transformer model.

\paragraph{Training.}
During training, we sample $\bar\tau$ uniformly and draw framewise timesteps using \cref{eq:tbar} with standard deviation cap $\sigma=0.1$. We then apply dimensionality-aware shifting using the full spatiotemporal resolution $T \times H \times W$.
We sample a maximum of 13 clean conditioning frames during training and use fully unconditional training with $\mathcal{C} = \emptyset$ with probability $p_U = 0.25$, enabling classifier-free guidance~\cite{ho2022ClassifierFree} at inference time.

We train with $25 \times 384 \times 384$ sequences from the SEVIR~\cite{veillette2020SEVIR} dataset ($24 \times 400 \times 400$ for MeteoNet). Exhaustive subsequence sampling following \citet{schusterbauer2026weatherrf-freud} yields 402,500 total training sequences (259,656 for MeteoNet).
\ours-T and \ours-S are trained for 300k iterations using the AdamW~\cite{loshchilov2019Decoupled} optimizer ($\text{lr} = 10^{-4}, \text{betas} = (0.9, 0.999)$) and a global batch size of 128.
Due to overfitting on small data we train \ours-B only for 200k iteration and train \ours-L only for 100k iterations after initializing from a checkpoint pretrained on smaller crops. We provide further information on overfitting and show scalability with more data in \cref{sec:sup:extended_ablations}.
We use Exponential Moving Average (EMA) weights to stabilize training.
For MeteoNet training we resize frames to $384 \times 384$ resolution to use the matching model configurations across datasets, while restoring $400 \times 400$ for metric calculation via bilinear upsampling. The \objective{} loss is calculated following \cref{alg:training}.

\paragraph{Inference.}
During inference, we integrate the ODE using standard Euler updates. We keep the auxiliary noise $\xi$ fixed for the denoising trajectory. Further, we do not use particle expansion and instead mirror prior work~\cite{leinonen2023ldcast, gao2023prediff, ribeiro2026flowcast, schusterbauer2026weatherrf-freud} and sample different noises $\mathbf{x}_0$ to ensemble forecasts. We use 50 NFE by default.

\section{Additional Ablations}\label{sec:sup:extended_ablations}

\begin{figure}[t]
    \centering
    \includegraphics[width=.9\linewidth]{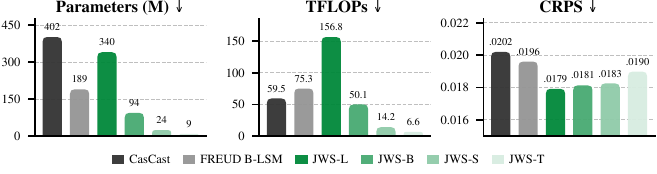}
    \caption{\textbf{FLOPs vs.\ Performance.} Our \ours-B, \ours-S, and \ours-T models are significantly more compute and parameter efficient while achieving superior nowcasting performance.}
    \label{fig:sup:params-flops-crps}
\end{figure}

\paragraph{FLOP comparison.}
In addition to wall clock results in \cref{fig:exp:latency}, \cref{fig:sup:params-flops-crps} shows model capacity and inference FLOPs in relation to resulting CRPS.
For the configurations shown, our smallest model requires 6.6\,TFLOPs compared with CasCast's 59.5\,TFLOPs, an approximately ninefold reduction.
Our models achieve superior CRPS with substantially lower parameter count and computational cost.
Note that \ours-L requires more inference FLOPs than CasCast but is faster in wall clock time, highlighting the benefit of reduced pipeline overhead.

\input{figs/4_exp/sevir_leadtime}

\paragraph{Performance over lead time.}
In addition to \cref{fig:exp:sevir-long}, we also evaluate performance over lead time for the in-domain setting.
\cref{fig:exp:leadtime} shows our model achieves superior CRPS and SSIM throughout the entire forecast horizon with advantages increasing for longer forecasts. HSS remains superior to FREUD~\cite{schusterbauer2026weatherrf-freud} and is competitive with CasCast~\cite{gong2024cascast} which relies on deterministic priors and classifier-free guidance to improve localization-centric metrics.

\begin{figure}[tb]
    \centering
    \begin{subfigure}{0.32\textwidth}
        \includegraphics[width=\textwidth]{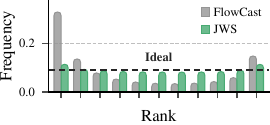}
    \end{subfigure}
    \hfill
    \begin{subfigure}{0.32\textwidth}
        \includegraphics[width=\textwidth]{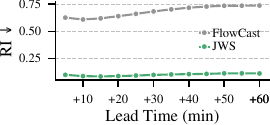}
    \end{subfigure}
    \hfill
    \begin{subfigure}{0.32\textwidth}
        \includegraphics[width=\textwidth]{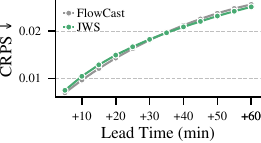}
    \end{subfigure}
    \caption{\textbf{Calibration compared to FlowCast.} FlowCast~\cite{ribeiro2026flowcast} shows substantially stronger overconfidence than \ours{} especially for long lead time.}
    \label{fig:sup:flowcast_calibration}
\end{figure}

\paragraph{FlowCast calibration comparison.} Complementing the comparison with CasCast~\cite{gong2024cascast} and FREUD~\cite{schusterbauer2026weatherrf-freud} in \cref{fig:exp:calibration}, \cref{fig:sup:flowcast_calibration} compares \ours{} and FlowCast~\cite{ribeiro2026flowcast} for calibration.
FlowCast exhibits a substantially stronger U-shaped rank histogram, indicating greater overconfidence.
Its reliability further deteriorates as lead time increases, with worse CRPS at longer horizons.
In contrast, \ours-B maintains better calibration and outperforms FlowCast on CRPS at the more challenging long-lead times.
These results demonstrate that \ours{} provides more reliable probabilistic forecasts, particularly as forecast uncertainty increases with lead time.

\begin{figure}[tb]
    \centering
    \includegraphics[width=.95\linewidth]{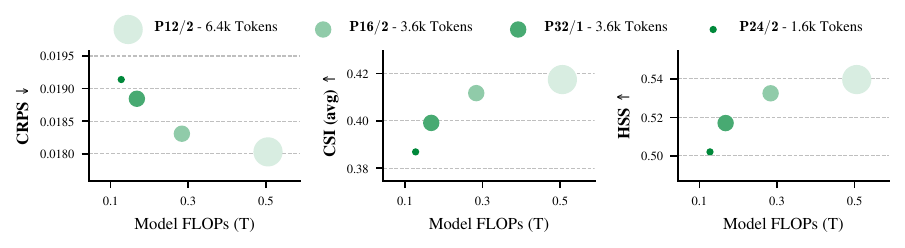}
    \caption{\textbf{Patch Size Ablation.} Varying patch sizes at both hierarchy levels changes the token count and thus increases FLOPs. More tokens/FLOPs improve CRPS, CSI, and HSS.}
    \label{fig:sup:patch-size}
\end{figure}

\paragraph{Patch size.}
Reducing the patch size at either hierarchy level increases the number of processed tokens.
\cref{fig:sup:patch-size} shows a clear scaling trend. Processing more tokens results in superior forecasts.

\begin{figure}[tb]
    \centering
    \includegraphics[width=\linewidth]{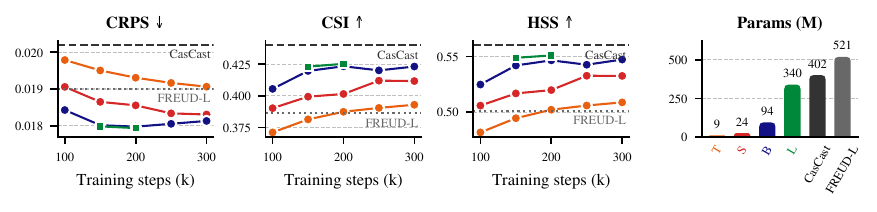}
    \caption{\textbf{Model scalability.} Our smallest \ours-T model already outperforms FREUD LSM-L \cite{schusterbauer2026weatherrf-freud}. With increased model scale, our model shows superior results.}
    \label{fig:sup:jws-scalability}
\end{figure}

\begin{figure}[tb]
    \centering
    \includegraphics[width=\linewidth]{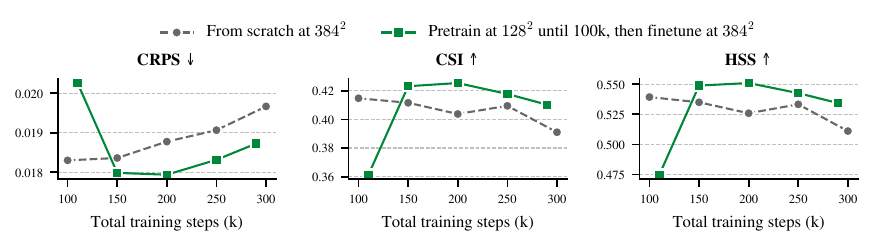}
    \caption{\textbf{Overfitting.} Training large (L) models directly on full resolution data leads to strong overfitting. Pretraining on downsampled data first mitigates this issue and preserves scalability.}
    \label{fig:sup:l-pretraining}
\end{figure}

\begin{figure}[tb]
    \centering
    \begin{subfigure}[t]{0.49\textwidth}
        \centering
        \includegraphics[width=\linewidth]{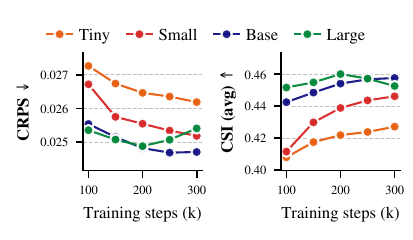}
        \caption{\textbf{50\% Data}}
    \end{subfigure}
    \hfill
    \begin{subfigure}[t]{0.49\textwidth}
        \centering
        \includegraphics[width=\linewidth]{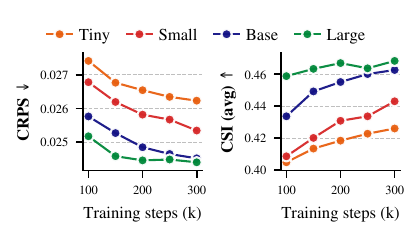}
        \caption{\textbf{100\% Data}}
    \end{subfigure}
    \caption{\textbf{Data scaling.} \ours-L overfits on reduced training data, while increased data diversity shows stable training and clearer gains from model scaling.}
    \label{fig:sup:scaling-192px}
\end{figure}

\paragraph{Scalability.}
\cref{tab:exp:sevir_nowcasting,tab:exp:meteonet} show that \ours{} scales effectively with model capacity.
However, the gains diminish for \ours-L, as shown in \cref{fig:sup:jws-scalability}.
We attribute this saturation to the limited diversity of the training data: although exhaustive subsequence sampling yields approximately 400k training sequences, these originate from only 16k distinct SEVIR weather events \cite{veillette2020SEVIR}.
We therefore hypothesize that further scaling requires increasing training data alongside model capacity, consistent with findings on compute-optimal scaling in large language models~\cite{hoffmann2022training}.
To test this hypothesis, we increase training data diversity through random cropping and compare training on 50\% and 100\% of SEVIR events.
We extract $192^2$ crops to preserve most of the spatial context and resize to $128^2$ for computational efficiency.
As shown in \cref{fig:sup:scaling-192px}, \ours-L exhibits substantially stronger overfitting than \ours-B when trained only on half of the available events.
In comparison, using the full augmented dataset eliminates overfitting. In this setting, \ours{} consistently benefits from increased model capacity. This supports our hypothesis that data availability limits model scaling.
For our main results on the full resolution data, we partially address this limitation by pretraining \ours-L at $128^2$ resolution for 100k steps, which further reduces training cost, and then finetune at full resolution.
As shown in \cref{fig:sup:l-pretraining}, training from scratch at full resolution steadily deteriorates validation performance, whereas pretraining followed by finetuning substantially improves performance and mitigates overfitting.

\begin{figure}[tb]
    \centering
    \includegraphics[width=.9\linewidth]{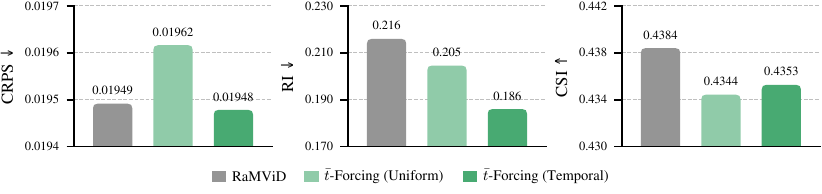}
    \caption{\textbf{Confidence-aware priors.} Using cleaner deterministic priors for early frames and noisier priors for later frames improves CRPS, calibration, and CSI over uniform prior strength, improving forecast calibration at longer lead times.}\label{fig:sup:confidence_forcing}
\end{figure}

\begin{figure}[tb]
    \centering
    \includegraphics[width=1\linewidth]{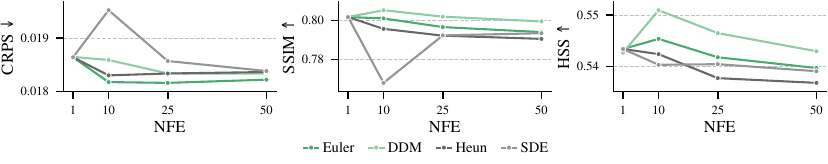}
    \caption{\textbf{Different solvers.} Standard Euler sampling outperforms Heun and SDE integration. While DDM sampling achieves superior HSS and SSIM, Euler performs best for CRPS.}
    \label{fig:sup:solver}
\end{figure}

\paragraph{Sampler.} \cref{fig:sup:solver} compares different integration methods across inference budgets. We evaluate standard Euler sampling, the distributional sampler by \citet{bortoli2025distributional}, Heun sampling, and SDE integration following the protocol of~\citet{ma2024SiT}. We find strong performance with standard Euler, outperforming Heun and SDE across all metrics and budgets. The distributional sampler achieves superior SSIM and HSS, indicating superior sharpness, but underperforms Euler in CRPS evaluation. We use Euler for simplicity.

\begin{figure}[tb]
    \centering
    \includegraphics[width=.98\linewidth]{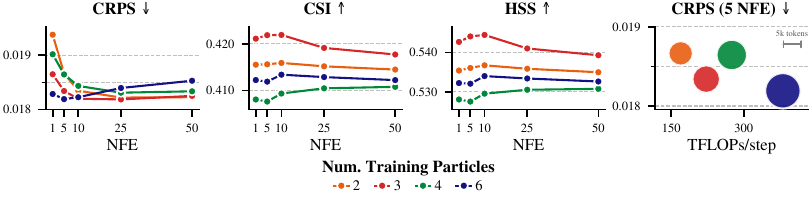}
    \vspace{-0.5cm}
    \caption{\textbf{Training particles.} Increasing training particles does not improve performance.}
    \label{fig:sup:training_particles}
\end{figure}

\paragraph{Number of Training Particles.}
Using larger training ensembles (more particles) increases computational cost. Yet, \cref{fig:sup:training_particles} does not reveal a scaling trend.
While six training particles achieve superior few-step performance, the model performs worse for high NFE. The four-particle model underperforms compared to using three particles.

\begin{figure}[tb]
    \centering
    \includegraphics[width=.98\linewidth]{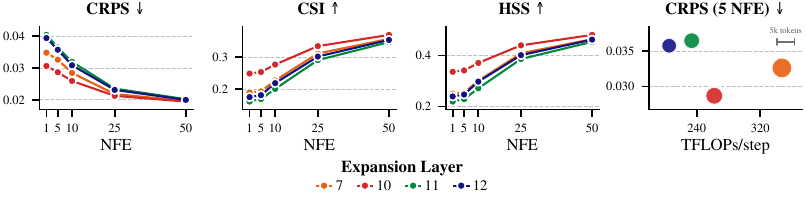}
    \vspace{-0.5cm}
    \caption{\textbf{Expansion layer.} Expanding the ensemble earlier does not yield improved performance.}
    \label{fig:sup:expansion_layer}
\end{figure}

\begin{figure}[tb]
    \centering
    \begin{subfigure}[t]{0.3\textwidth}
        \centering
        \includegraphics[width=\linewidth]{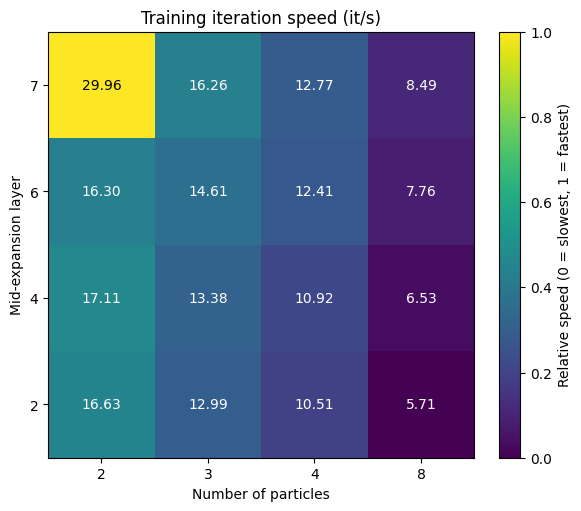}
        \caption{Throughput.}
    \end{subfigure}\hfill
    \begin{subfigure}[t]{0.3\textwidth}
        \centering
        \includegraphics[width=\linewidth]{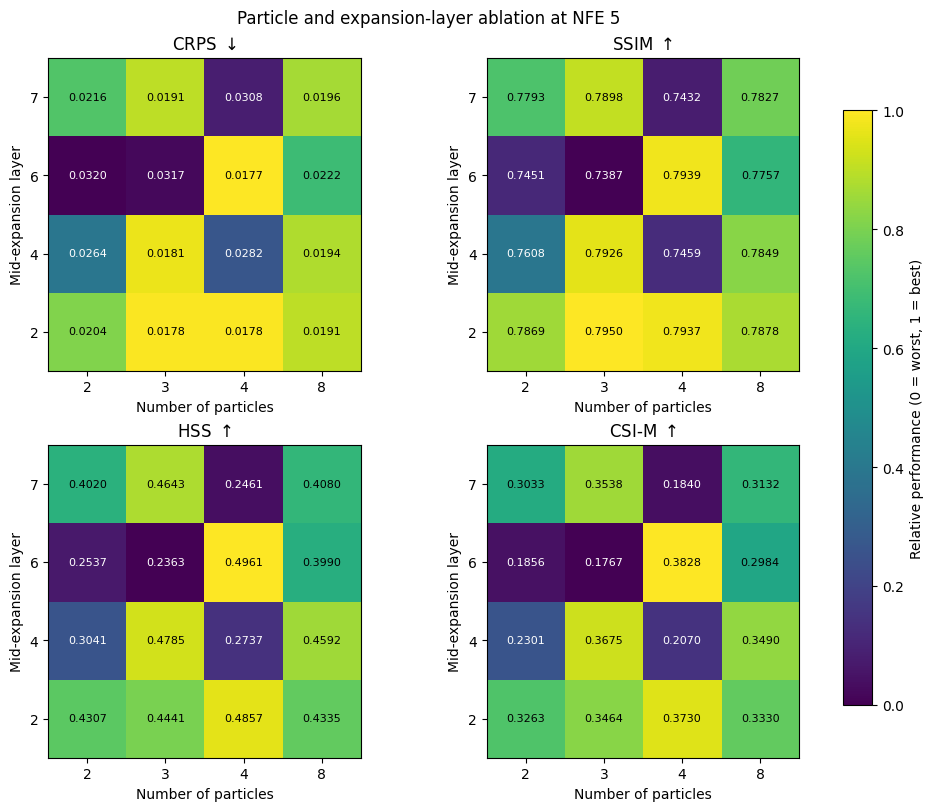}
        \caption{5 NFE.}
    \end{subfigure}\hfill
    \begin{subfigure}[t]{0.3\textwidth}
        \centering
        \includegraphics[width=\linewidth]{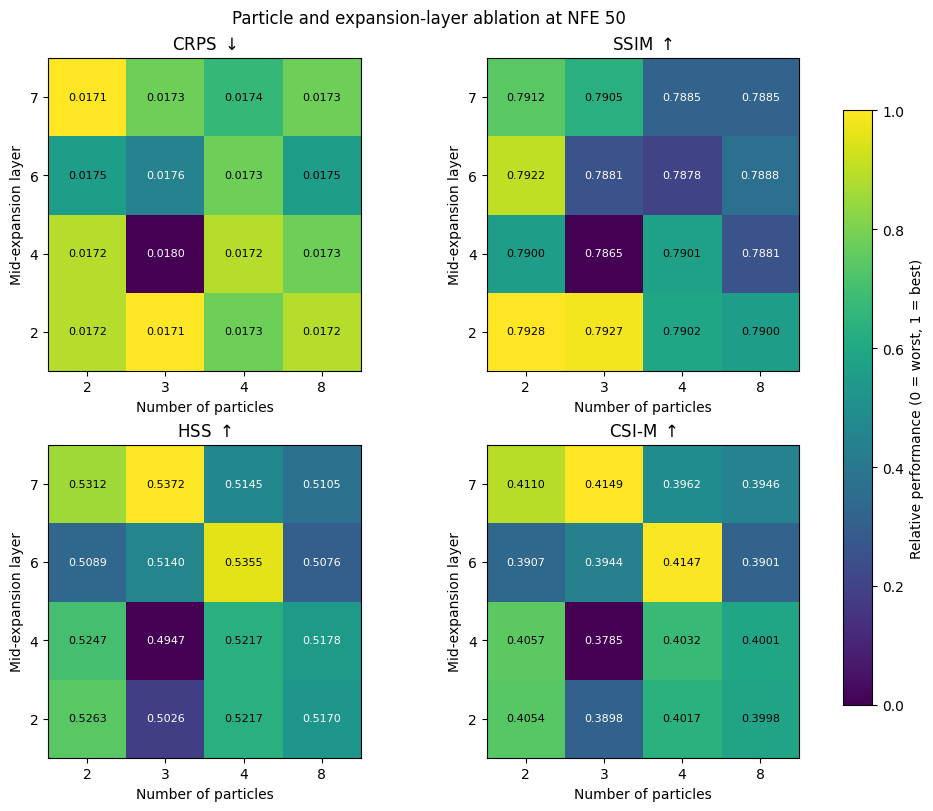}
        \caption{50 NFE.}
    \end{subfigure}
    \caption{\textbf{Small-scale verification.} We again find no gain using more particles or earlier expansion.}
    \label{fig:sup:particle-expansion-s-heatmaps}
\end{figure}

\paragraph{Expansion layer.}
Earlier expansion of the ensemble increases computational cost during training. Matching \citet{martorella2026iddm}, \cref{fig:sup:expansion_layer} does not show improved performance by expanding earlier.
Instead, ensemble expansion at the last bottleneck layer seems to provide superior performance.

Additionally, we conducted an ablation on downsampled $128\,\text{px}^2$ SEVIR data with an S-scale hierarchical model.
\cref{fig:sup:particle-expansion-s-heatmaps} shows heatmaps for the full expansion layer $\times$ training particles grid. Again, there is no consistent impact of more particles or earlier expansion on final performance.

\begin{table}[tb]
    \centering
    \resizebox{0.93\linewidth}{!}{%
    \begin{minipage}{\linewidth}
    \begin{minipage}[t]{0.48\linewidth}
        \vspace{0pt}
        \centering
        \caption{\textbf{Conditioning path.} Concatenating
        noise to the residual stream is best.}
        \label{tab:sup:xi-conditioning}
        \scriptsize
        \setlength{\tabcolsep}{5pt}
        \resizebox{\linewidth}{!}{%
            \begin{tabular}{@{}lccc@{}}
                \toprule
                $\xi$ conditioning
                & \textbf{CRPS} $\downarrow$
                & \textbf{HSS} $\uparrow$
                & \textbf{CSI-M} $\uparrow$ \\
                \midrule
                Single concatenation
                    & 0.0236 & 0.4021 & 0.3023 \\
                Residual concatenation
                    & \textbf{0.0217} & \textbf{0.4351}
                    & \textbf{0.3295} \\
                Residual concatenation + 2 clean
                    & 0.0276 & 0.3895 & 0.2916 \\
                Single addition
                    & 0.0238 & 0.4176 & 0.3125 \\
                Repeated addition
                    & 0.0243 & 0.4010 & 0.3028 \\
                Repeated addition + 2 clean
                    & 0.0245 & 0.4118 & 0.3081 \\
                \bottomrule
            \end{tabular}%
        }
    \end{minipage}\hfill%
    \begin{minipage}[t]{0.48\linewidth}
        \vspace{0pt}
        \centering
        \caption{\textbf{Tbar $\sigma$ ablation.}
        Setting $\sigma = 0.1$ achieves the best result.}
        \label{tab:sup:tbar}
        \scriptsize
        \setlength{\tabcolsep}{3pt}
        \resizebox{\linewidth}{!}{%
            \begin{tabular}{@{}lccc@{}}
                \toprule
                $\sigma$
                & \textbf{CRPS} $\downarrow$
                & \textbf{HSS} $\uparrow$
                & \textbf{CSI-M} $\uparrow$ \\
                \midrule
                $0$ {\tiny (RaMViD \cite{hoppe2022ramvid})}
                    & 0.01940 & 0.4740 & 0.3638 \\
                $0.05$
                    & 0.01921 & 0.5072 & 0.3889 \\
                $0.1$
                    & \textbf{0.01895} & \textbf{0.5154}
                    & \textbf{0.3965} \\
                $0.2$
                    & 0.01940 & 0.4886 & 0.3750 \\
                \ding{55} {\tiny (RaMViD Forcing)}
                    & 0.02771 & 0.3875 & 0.2826 \\
                \bottomrule
            \end{tabular}%
        }
    \end{minipage}

    \par\vspace{1em}

    \begin{minipage}[t]{0.48\linewidth}
        \vspace{0pt}
        \centering
        \caption{\textbf{Reweighting SEVIR.}
        Data filtering and loss weighting have minimal impact.}
        \label{tab:sup:reweighting_sevir}
        \footnotesize
        \setlength{\tabcolsep}{3pt}
        \renewcommand{\arraystretch}{1.15}
        \resizebox{\linewidth}{!}{%
            \begin{tabular}{@{}lccccc@{}}
                \toprule
                \textbf{Method}
                & \textbf{CRPS} $\downarrow$
                & \textbf{SSIM} $\uparrow$
                & \textbf{HSS} $\uparrow$
                & \textbf{CSI-M} $\uparrow$
                & \textbf{RI} $\downarrow$ \\
                \midrule
                Baseline
                    & \textbf{0.02003} & 0.7653 & 0.5034
                    & 0.3869 & 0.0993 \\

                \arrayrulecolor{gray!50!white}
                \midrule
                \arrayrulecolor{black}

                Filtering
                    & 0.02086 & 0.7416 & \textbf{0.5087}
                    & \textbf{0.3908} & \textbf{0.0712} \\
                Power law
                    & 0.02016 & \textbf{0.7680} & 0.4691
                    & 0.3587 & 0.1077 \\
                Binary
                    & 0.02015 & 0.7669 & 0.4773
                    & 0.3652 & 0.0928 \\
                \bottomrule
            \end{tabular}%
        }
    \end{minipage}\hfill%
    \begin{minipage}[t]{0.48\linewidth}
        \vspace{0pt}
        \centering
        \caption{\textbf{Reweighting MeteoNet.}
        Reweighting and resampling increase CRPS.}
        \label{tab:sup:reweighting_meteonet}
        \footnotesize
        \setlength{\tabcolsep}{3pt}
        \renewcommand{\arraystretch}{1.15}
        \resizebox{\linewidth}{!}{%
            \begin{tabular}{@{}lccccc@{}}
                \toprule
                \textbf{Method}
                & \textbf{CRPS} $\downarrow$
                & \textbf{SSIM} $\uparrow$
                & \textbf{HSS} $\uparrow$
                & \textbf{CSI-M} $\uparrow$
                & \textbf{RI} $\downarrow$ \\
                \midrule
                Baseline
                    & \textbf{0.01552} & \textbf{0.7407} & 0.3942
                    & 0.2787 & 0.3640 \\

                \arrayrulecolor{gray!50!white}
                \midrule
                \arrayrulecolor{black}

                Filtering
                    & 0.01586 & 0.7307 & \textbf{0.4107}
                    & \textbf{0.2920} & 0.4302 \\
                Power law
                    & 0.01557 & 0.7353 & 0.4062
                    & 0.2885 & 0.3918 \\
                Binary
                    & 0.01554 & 0.7386 & 0.3997
                    & 0.2838 & \textbf{0.3625} \\
                \bottomrule
            \end{tabular}%
        }
    \end{minipage}%
    \end{minipage}%
    }
\end{table}

\paragraph{Auxiliary conditioning.}
On the $128\,\text{px}^2$ SEVIR data, \cref{tab:sup:xi-conditioning} evaluates different mechanisms for $\xi$ injection.
We compare concatenation, addition, repeated addition in every layer, and residual concatenation~\cite{martorella2026iddm}.
Matching iDDM, concatenation to the residual stream is superior.
Additionally, we test keeping final layers clean, which deteriorates performance.

\paragraph{$\bar\tau$ spread.}
\cref{tab:sup:tbar} compares different spread values $\sigma$ for \paradigm{} sampling.
$\sigma = 0.1$ outperforms other spread values, including uniform timesteps and uncorrelated timesteps.

\input{tabs/X_sup/loss-reweighting}
\paragraph{Loss reweighting and data filtering.}
Previous work~\cite{gong2024cascast} filters MeteoNet and trains only on sequences with observed precipitation.
In addition to ``all data'' we test (with small JiT models): filtering, binary loss weighting, downweighting dry pixels based on their frequency in the sample, and power-law loss weighting.
However, \cref{tab:sup:reweighting_sevir} and \cref{tab:sup:reweighting_meteonet} show that neither filtering nor reweighting improves overall performance, while all such interventions deteriorate CRPS.
We therefore require no dataset filtering or loss reweighting and train directly on the full training dataset with equally weighted samples and pixels.

\section{Metrics and Evaluation Protocol}\label{sec:sup:metrics}

To enable better interpretability of our results, we summarize metrics and the evaluation protocol below.

\subsection{Metric Definitions}

We use the metrics implemented by \citet{schusterbauer2026weatherrf-freud} which adopts the evaluation by \citet{gong2024cascast}. 

\paragraph{CRPS.}
CRPS is a proper scoring rule for univariate regression that generalizes the Mean Absolute Error to probabilistic tasks~\cite{gneiting2007strictly}. For observed value $x$, cumulative forecast distribution $F(y)$, and ensemble members $\{ f_1, ..., f_N \}$ CRPS is defined as
\begin{equation}
\begin{aligned}
    &CRPS(F, x) = \int_{-\infty}^{\infty} (F(y) - \mathbf 1_{y \geq x})^2 \text{ d}y \\
    &= \mathbb{E}_{X \sim F}\left [||X - x|| \right ] - \frac{1}{2} \mathbb{E}_{X, X' \sim F} \left [ |X - X'| \right] \\
    & \approx \frac{1}{N} \sum_{i = 1}^N |f_i - x| - \frac{1}{2 N^2} \sum_{i=1}^N \sum_{j=1}^N |f_i - f_j|. \nonumber
\end{aligned}
\end{equation}
The first term pushes the expectation to the ground truth, while the second term encourages spread.

\paragraph{SSIM.} 
The Structural Similarity Index Measure is used to measure structural similarity between ground truth and mean forecasts. For single-channel data SSIM is defined as
\begin{equation}
 \text{SSIM}(x, y) =\frac{(2 \mu_x \mu_y + C_1)(2 \sigma_{xy} + C_2)}{(\mu_x^2 + \mu_y^2 + C_1)(\sigma_x^2 + \sigma_y^2 + C_2)},
 \nonumber
\end{equation}
where $x$ is the ground truth and $y$ is the prediction, $\mu_x$ and $\mu_y$ are the average intensities, $\sigma_x^2$ and $\sigma_y^2$ are the variances of intensities, $\sigma_{xy}$ is the covariance between the images and $C_1$ and $C_2$ are two small constants.
Following standard practice, SSIM is calculated by averaging sliding window results and using Gaussian weighting to calculate mean and variance values.

\paragraph{HSS.} 
Heidke Skill Score (HSS) is computed on a per-pixel basis to identify positional accuracy. Using a user-defined threshold, true positives (TP), false positives (FP), false negatives (FN), and true negatives (TN) are first calculated based on binarized ground truth and forecasted precipitation. Then HSS is computed as
\begin{equation}
    HSS = \frac{
 2 (TP  \cdot TN - FP \cdot FN)
 }{(TP + FN)(FN + TN)+(TP + FP)(FP + TN)}.
\end{equation}
Following prior work we average HSS over six thresholds (16, 74, 133, 160, 181, 219) to obtain the final value.
Intuitively, HSS indicates improvement over random chance, where 0 indicates no improvement and 1 corresponds to perfect prediction.

\paragraph{CSI.}
The Critical Success Index (CSI) or Threat Score quantifies the proportion of correctly predicted events excluding true negatives. Therefore, it remains useful when non-events dominate the data as in precipitation forecasting.
After the same binarization as for HSS, CSI is given by
\begin{equation}
    CSI = \frac{TP}{TP + FP + FN}.
\end{equation}
The reported CSI is averaged over the same thresholds as HSS.

\paragraph{RI.}
To compute the reliability index (RI) we first compute rank histograms by inserting observed pixel values into a sorted list of ensemble predictions and counting the rank. Therefore, for an $N$-member ensemble, the rank histogram shows $N+1$ possible ranks, resolving ties by random sampling.
RI then measures the deviation of the resulting relative frequencies ($p_r$) from a uniform distribution as
\begin{equation}
    \mathrm{RI} = \sum_{r=0}^{N} \left|p_r - \frac{1}{N+1}\right|.
\end{equation}
Thus, lower values indicate closer agreement with uniform rank distribution, i.e.\ superior calibration.

\paragraph{SSR.}
Spread-Skill Ratio (SSR) compares the ensemble spread with root mean squared error.
SSR is computed as
\begin{equation}
    \mathrm{SSR} = \frac{\sqrt{\sigma^2}}{\sqrt{(\bar f-x)^2}},
\end{equation}
where $\bar f = N^{-1}\sum_{i=1}^{N} f_i$ is the ensemble mean, $\sigma^2 = (N-1)^{-1}\sum_{i=1}^{N}(f_i-\bar f)^2$ is the sample variance.
A calibrated ensemble forecast should have high variance when uncertain and therefore most wrong. Thus, calibration is indicated by SSR close to 1.

\subsection{Evaluation Protocol}

\begin{table}[tb]
    \centering
    \caption{\textbf{Metric sensitivity.} Results report mean $\pm$ standard deviation across eight evaluation seeds. Mean results are close to large evaluation runs even for small sample sizes.}
    \label{tab:sup:sensitivity}
    \setlength{\tabcolsep}{4pt}
    \footnotesize
    \newcommand{\meansd}[2]{%
        $#1$\,{\scriptsize $\pm\,#2$}%
    }
    \resizebox{\linewidth}{!}{%
        \begin{tabular}{@{}cccccc@{}}
            \toprule
            \textbf{Num. Samples}
            & \textbf{CRPS} $\downarrow$
            & \textbf{SSIM} $\uparrow$
            & \textbf{HSS} $\uparrow$
            & \textbf{CSI} $\uparrow$
            & \textbf{RI} $\downarrow$ \\
            \midrule
            64
                & \meansd{0.0259}{0.0025}
                & \meansd{0.6374}{0.0143}
                & \meansd{0.3379}{0.0115}
                & \meansd{0.2529}{0.0093}
                & \meansd{0.0615}{0.0183} \\
            128
                & \meansd{0.0257}{0.0013}
                & \meansd{0.6372}{0.0123}
                & \meansd{0.3384}{0.0102}
                & \meansd{0.2527}{0.0078}
                & \meansd{0.0562}{0.0106} \\
            256
                & \meansd{0.0257}{0.0008}
                & \meansd{0.6363}{0.0065}
                & \meansd{0.3365}{0.0077}
                & \meansd{0.2515}{0.0054}
                & \meansd{0.0558}{0.0064} \\
            512
                & \meansd{0.0260}{0.0005}
                & \meansd{0.6331}{0.0050}
                & \meansd{0.3371}{0.0055}
                & \meansd{0.2521}{0.0039}
                & \meansd{0.0573}{0.0039} \\
            1024
                & \meansd{0.0260}{0.0004}
                & \meansd{0.6322}{0.0028}
                & \meansd{0.3363}{0.0033}
                & \meansd{0.2511}{0.0027}
                & \meansd{0.0581}{0.0020} \\
            \bottomrule
        \end{tabular}%
    }
\end{table}

Unless indicated otherwise we use 50 network evaluations for our method and compare all approaches using 10 ensemble members.
Experiments are mainly conducted on the SEVIR dataset~\cite{veillette2020SEVIR} using the train-test split provided by \citet{gong2024cascast}.
We use the full test set to establish state-of-the-art in \cref{tab:exp:sevir_nowcasting} and \cref{tab:exp:meteonet}.
However, following prior work~\cite{schusterbauer2026weatherrf-freud, ribeiro2026flowcast} we use a subset of 512 samples for plots and ablation experiments due to computational constraints. \cref{tab:sup:sensitivity} shows that 512 sample evaluation closely matches results with more samples and exhibits low inter-seed variance.
All evaluations are carried out on 4 Nvidia GH200 GPUs using parallelization across samples for speedup, while latency is computed with a single GPU.

For baselines we use the results reported in the literature~\cite{gong2024cascast, schusterbauer2026weatherrf-freud, ribeiro2026flowcast} after verifying reproducibility with our pipeline where possible. Metrics that were previously not reported are recomputed using available checkpoints if trained weights are provided by the authors.

\subsection{Datasets}\label{sec:sup:metrics:datasets}

\paragraph{SEVIR.}
Our main evaluations are carried out using the SEVIR~\cite{veillette2020SEVIR} dataset.
SEVIR contains 20,393 weather events of which 20\% correspond to known extreme weather events from the NOAA Storm Event Database~\cite{NOAA2020StormEvents}. The remaining 80\% were randomly chosen while oversampling high precipitation.
SEVIR is collected from 2017--2019 over the continental United States region. Each sample covers a $384\,\text{km} \times 384\,\text{km}$ area for 4\,h with 1\,km spatial and 5\,min temporal resolution.
Precipitation is indicated by the NEXRAD Vertically Integrated Liquid product (VIL).

We use the train-test split provided by \citet{gong2024cascast} and follow \citet{schusterbauer2026weatherrf-freud} in joining the validation set into training and sampling exhaustive subsequences to increase training data size. Following prior work, we use 65\,min (13 frames) conditioning to predict 60\,min (12 frames) future precipitation unless noted otherwise.
We follow the standard preprocessing~\cite{veillette2020SEVIR} and keep non-linear VIL encoding to $[0, 255]$ which we normalize further to $[-1,1]$ for training following \citet{schusterbauer2026weatherrf-freud}. 
To ensure comparability~\cite{gong2024cascast, gao2022earthformer, she2024LLMDiff, schusterbauer2026weatherrf-freud, ribeiro2026flowcast}, we use $H=W=384$, $C=13$, $L=12$, and $T = C + L = 25$ unless specified otherwise.

\paragraph{MeteoNet.}
The MeteoNet benchmark~\cite{larvor2020meteonet} includes observations over fixed regions in northwestern and southeastern France from 2016--2018. Radar measurements are provided in 5\,min intervals with $\approx 1\,\text{km}^2$ resolution.
Following \citet{gong2024cascast}, we use the old radar reflectivity product and extract the top-left $400\times400$ grid cells of the southeastern region.
Again, we combine training and validation periods and exhaustively sample subsequences for training, while testing on the same data as prior work~\cite{gong2024cascast, schusterbauer2026weatherrf-freud}.
Following~\citet{gong2024cascast} we use 24-frame subsequences and predict the next 12 frames given a 12-frame context. We normalize the reflectivity values from $[0, 70]$ to $[-1,1]$ for training. To enable using the same \ours{} architecture as for SEVIR, we bilinearly resize to $384 \times 384$ for model processing, but keep metric computation at the native resolution.
To compute HSS and CSI we use thresholds of $[19, 25, 35, 40, 47]$ dBZ~\cite{gong2024cascast}.

\section{Extended Related Work}\label{sec:sup:extended_rw}

We expand on the related works presented in \cref{sec:related} to provide additional context on existing methods and our advances upon these.

\subsection{Machine Learning for Nowcasting}

Weather nowcasting refers to short-term, high-resolution forecasting over lead times ranging from one to twelve hours. Typically nowcasting operates on $\leq 5$\,km spatial grids with $\leq 10$\,min temporal resolution. Physics-based numerical weather prediction is computationally expensive, and therefore not well-suited for applications that require rapid updates~\cite{shi15:ConvLSTM, shi17:TrajectoryGRU, sonderby20:MetNet, leinonen2023ldcast, ravuri21:GAN, chen2023Foundation}.
Early research used optical flow based forecasts \citep{woo2017Operational, cheung2012Application, germann2002ScaleDependence, sakaino2013SpatioTemporal} for rapid-update tasks, however these approaches cannot model rain onset or dissipation \citep{hur2019Iterative, shi15:ConvLSTM}.
Therefore, machine learning methods based on convolutional neural networks \cite{shi15:ConvLSTM, agrawal19:UNet}, recurrent neural networks \cite{wang2017predrnn}, or transformers~\cite{gao2022simvp, gao2022earthformer} have been developed. These approaches use patterns in observed past context to regress future weather states.
While machine learning methods also begin to dominate medium- and long-range forecasting~\cite{sonderby20:MetNet, espeholt22:MetNet2, pathak2022FourCastNet, price2024GenCast, alet2025fgn}, they are especially important in nowcasting due to their significantly reduced inference time.

However, weather dynamics are inherently chaotic and unpredictable~\cite{ravuri21:GAN} requiring probabilistic forecasting under partially observable initial conditions.
Yet, deterministic regression models lack robust uncertainty quantification and average possibilities leading to increasingly blurry forecasts over lead time.
\citet{alet2025fgn} proposed Functional Generative Networks (FGNs) for medium-range forecasting, training a deterministic model to predict an ensemble of forecasts directly by optimizing an ensemble scoring function. However, they restrict the variability of forecasts to be encoded in a low-dimensional noise vector. Local variability in nowcasting is typically higher than variance in medium-range forecasting limiting the applicability of this approach compared to diffusion models due to reduced distributional coverage. Therefore, FGNs still become increasingly blurry in our experiments (Appendix~\ref{sec:sup:qualitative_comparisoms}).

Therefore, researchers employed generative adversarial \cite{ravuri21:GAN, zhang2023skilful, liu2020MPLGAN} and generative diffusion \cite{leinonen2023ldcast, gao2023prediff, ribeiro2026flowcast} methods, trained to produce a sample from the conditional distribution of predictions rather than to regress the mean. As adversarial approaches are prone to mode collapse~\cite{kossale2022Mode} and, thus, fail to capture the full posterior distribution, diffusion-based nowcasting has become the de facto standard approach.
Early diffusion-based precipitation nowcasting approaches sampled future evolution based only on prior observations~\cite{leinonen2023ldcast, gao2023prediff}. However, later papers argued conditioning the model on a deterministic forecast~\cite{gong2024cascast} or sampling only the residual between a deterministic forecast and the ground truth~\cite{yu2024DiffCast} enables superior forecasting.
Similarly, \citet{park2026diffusionbased} build upon the deterministic exPreCast model~\cite{song2026extremea} by using a diffusion model to model only small scale variability and enable ensembling of stochastic elements.
However, FREUD~\cite{schusterbauer2026weatherrf-freud} argues conditioning on a deterministic forecast biases the generation and thus limits uncertainty quantification. Compounding evidence by \citet{ribeiro2026flowcast} supports the claim that training a sufficiently capable diffusion model suffices for accurate forecasting and superior calibration.

An additional downside is that most diffusion-based nowcasting methods~\cite{leinonen2023ldcast, gao2023prediff, gong2024cascast, ribeiro2026flowcast, schusterbauer2026weatherrf-freud} operate in a compressed latent space to enable faster sampling. However, compression is inherently lossy and thus induces additional uncertainty about reconstructions.
\citet{schusterbauer2026weatherrf-freud} propose to quantify this uncertainty using generative reconstructions. However, this means for each latent ensemble member a full ensemble of reconstructions has to be constructed to quantify all uncertainty, inducing a significant computational overhead.
Some previous works instead tackle pixel-space nowcasting using specialized architectures~\cite{nguyen2025raindiff} or by only modeling residuals from a deterministic forecast~\cite{yu2024DiffCast, zhu2026pixelflowcast} which still requires a two-stage design and biases forecasts.
In comparison, \ours{} enables pixel space nowcasting with a standard, scalable architecture and with a single end-to-end model, simplifying training and inference.

\subsection{Pixel-space Diffusion}

Initial works in diffusion-based image generation applied iterative denoising in pixel-space~\cite{dhariwal2021beatGANs, ho2020ddpm}. Yet, computational costs made these models difficult to scale and deploy.
Latent diffusion~\cite{rombach2022ldm} proposed to first compress perceptual information into a low-dimensional latent and then apply the diffusion process only in this compact latent space. This enabled high-resolution image generation on consumer hardware and training video generation models with academic resources~\cite{he2023Latenta}.

However, the complexity of training multiple stages motivated the authors of simple diffusion~\cite{hoogeboom2023simple} to develop an architecture for direct pixel-space diffusion modeling by adapting the noise schedule and downsampling early in the UNet~\cite{ronneberger2015UNeta} backbone to concentrate capacity in the bottleneck. Further improvements~\cite{hoogeboom2025simplera} identify reducing skip connections as a pathway to further enhance scalability. Through this modification, improved loss weighting, and preferring scaling the number of processed patches over increasing the model capacity they achieve strong pixel-space generation performance.

Alternatives like PixelFlow~\cite{chen2025pixelflow} propose to modify the denoising process itself to enable pixel-space modeling. \citet{chen2025pixelflow} propose to schedule the image resolution over the denoising trajectory, progressively moving from noisy low resolution images to clean high resolution images during sampling. Thus, their approach saves compute on high-frequency details while these are masked by noise.
\citet{wang2025pixnerd}, on the other hand, enable high-resolution image synthesis by predicting patch-specific neural-field weights, with the resulting fields mapping noisy pixel values and spatial coordinates to pixel-wise velocities. This concentrates expensive transformer computation on low-resolution patches while using lightweight neural fields for pixel-level prediction, avoiding the cost of processing each pixel as a separate transformer token.

Alternatively, both PixelDiT~\cite{yu2025pixeldit} and DiP~\cite{chen2026dip} propose combining a large transformer for global structure with local refinement modules to enable pixel-space high resolution image generation.
Other alternative architectures such as RIN~\cite{jabri2023rin} and FIT~\cite{chen2023fit} enable high-resolution pixel-space modeling by reducing computational complexity of expensive self-attention. RIN~\cite{jabri2023rin} concentrates computation on a compact set of latent tokens that exchange information with pixel-space data tokens through cross-attention, reusing latent representations across denoising steps. \citet{chen2023fit} propose interleaving local attention within groups of data tokens with global attention over a smaller subset of tokens.

However, \citet{li2025back} show that standard Vision Transformer architectures~\cite{dosovitskiy2020vit, peebles2023Scalable} with a large patch size are capable of accurate pixel-space generation when adjusting timestep sampling during training and predicting clean visual information. They argue that under the manifold assumption clean data has lower intrinsic dimensionality than noise or noise-containing velocities, thus making clean data easier to compress by patchification.
Similarly, HDiT~\cite{crowson2024hdit} presents a hierarchical transformer architecture that substantially reduces computational requirements in pixel-space modeling while maintaining the strong scalability of vision transformers.
However, due to the additional temporal dimension, pixel-space generation of video or weather forecasting data is especially challenging and computationally demanding~\cite{jabri2023rin}.
Closest to our work, FREUD~\cite{schusterbauer2026weatherrf-freud} uses a pixel-space HDiT to reconstruct precipitation fields from compressed latents. They show that a generative decoder modeling the full video sequence increases temporal consistency. However, the FREUD decoder receives strong framewise conditioning information and therefore operates efficiently with low model capacity and few-step inference.
Instead, \ours{} learns end-to-end weather nowcasting directly in pixel-space. We use standard transformer architectures and denoising procedures, maintaining scalability and simplicity.

\subsection{Few-Step Sampling}

Precipitation nowcasting requires rapid updates~\cite{shi15:ConvLSTM, ravuri21:GAN}, which is at odds with the iterative nature of generative diffusion and flow models that tend to improve with more function evaluations.
Reducing computational complexity and latency by reducing the number of sampling steps required to produce high-quality samples has been a focus of many researchers in recent years.
Most prominently, researchers proposed methods to reduce the number of inference steps by modifying the denoising trajectory.
DDIM~\cite{song2020ddim} defines an ODE path for trained diffusion models that can be traversed with larger steps. While Flow Matching methods~\cite{lipman2023flowmatching} explicitly define a straighter trajectory during training, which allows faster traversal~\cite{liu2022Flow}.
However, these methods typically still require about 20 network evaluations to produce high-quality samples.

Few-step distillation approaches first train a default diffusion model and then distill the model's output into a few-step generative model using progressive MSE-based distillation~\cite{salimans2022progressivea}, adversarial objectives~\cite{sauer2023adversarial}, consistency objectives~\cite{song2023consistencya}, or distribution matching objectives~\cite{yin2024onestep, yin2024improved}.
However, distillation introduces additional training complexity by requiring a two-stage process. Additionally, previous research shows distillation often reduces the diversity of samples~\cite{yin2024improved, gandikota2025distilling}, limiting applicability in domains where capturing the full distribution of potential outcomes is essential.

Consistency models~\cite{song2023consistencya} can be used for distillation as well as for training from scratch.
Their training objective puts them in the family of Flow Map models~\cite{boffi2025flow} which are trained to directly integrate the denoising trajectory between two diffusion timesteps.
Flow maps can be grouped into three categories following \citet{boffi2025how}: First, Eulerian Flow Maps enforce that a change in the starting point should not change the outcome at the destination time. Second, progressive Flow Maps train models s.t.\ one large jump agrees with the composition of two smaller jumps. Finally, Lagrangian Flow Maps enforce that a change in the destination time corresponds to following the velocity field at the original destination.

MeanFlows~\cite{geng2025mean} are the most prominent example of Eulerian Flow Maps, yet require an expensive Jacobian-vector product (JVP) operation to obtain the training target. While improved MeanFlows~\cite{geng2026improved} stabilizes the training objective and PixelMeanFlow~\cite{lu2026onestep} combines improved MeanFlow with JiT~\cite{li2025back}, they retain the expensive JVP computation.
In comparison, SplitMeanFlow~\cite{guo2025splitmeanflow} eliminates the JVP computation and resembles Shortcut model~\cite{frans2025one} training. These methods fall in the category of progressive Flow Maps. Yet, they require multiple network evaluations during training to obtain self-bootstrapped targets, instead of training for superior distributional coverage. Lagrangian Self-Distillation (LSD)~\cite{boffi2025how} is an example of the final Flow Map category. These approaches still require time differentiation through the network and additional velocity evaluation at the predicted endpoint to derive their target.
Moreover, although Flow Map methods enable few-step generation, there is compounding evidence~\cite{huang2026rmflow, nguyen2025improved} that distributional coverage is sensitive to the training formulation, potentially increasing the difficulty of applying Flow Maps in domains where uncertainty is essential.

Distribution-level objectives offer a promising alternative.
Inductive Moment Matching (IMM)~\cite{zhou2025inductive} constructs a Flow Map by learning mappings between noise levels through matching output distributions instead of individual samples with a maximum mean discrepancy objective.
Improving simplicity, Distributional Diffusion Models (DDM)~\cite{bortoli2025distributional} show that training a diffusion model to predict a distribution of outputs rather than the conditional mean by supervision with a proper scoring rule suffices to build strong few-step generative models. Distributional Diffusion Models are especially promising for weather forecasting applications as they directly train a generative model to produce an ensemble of outputs.
However, DDMs require multiple passes through the full network to produce an ensemble of outputs during training and do not perform well for the many-step setting.
Released only two weeks prior to our paper, \citet{park2026proper} propose a distributional diffusion model for medium-range weather forecasting with marginal CRPS supervision. However, they focus on one-step predictions and cannot scale to more NFEs, mirroring our experiments with only-CRPS supervision (\cref{tab:exp:objective_nfe}).
In contrast, improved Distributional Diffusion Models (iDDM)~\cite{martorella2026iddm} show that using a time-dependent interpolation between distributional and MSE objectives produces a model that performs well in few and many step inference. Additionally, combined with strong conditioning on auxiliary noise, late expansion of the ensemble for only a few final layers is sufficient to train a DDM, making training scalable.

PixelFlowCast~\cite{zhu2026pixelflowcast} built a pixel-space nowcasting method with a (Pixel-)MeanFlow objective~\cite{lu2026onestep, geng2025mean}. However, it only models the residual of a deterministic forecast. Further, due to the cost of MeanFlow training, their model was only trained for downsampled data.
We argue that Distributional Diffusion~\cite{bortoli2025distributional, martorella2026iddm} is a natural fit for weather nowcasting and makes efficient nowcasting with standard transformer architectures~\cite{li2025back, dosovitskiy2020vit, crowson2024hdit, peebles2023Scalable, ma2024SiT} feasible without biasing generation~\cite{schusterbauer2026weatherrf-freud}.

\FloatBarrier

\clearpage

\section{Qualitative Results}\label{sec:sup:qualitative}

\subsection{Qualitative Forecast Ensembles}\label{sec:sup:qualitative_forecasts}

\FloatBarrier

\input{figs/X_supp/qual_forecasts}

\FloatBarrier

\subsection{Qualitative Comparisons}\label{sec:sup:qualitative_comparisoms}

\FloatBarrier

\input{figs/X_supp/qual_comparisons}

\FloatBarrier

\subsection{Qualitative Uncertainty}\label{sec:sup:qualitative_uncertainty}

\FloatBarrier

\input{figs/X_supp/qual_ensembles}

\FloatBarrier

\clearpage

\subsection{Qualitative Inference-time Scaling}\label{sec:sup:qualitative_nfe}

\FloatBarrier

\input{figs/X_supp/qual_nfe}

\FloatBarrier

\clearpage

%% file: tabs/X_sup/hyperparams.tex
\begin{table}[h]
    \centering
    \footnotesize
    \setlength{\tabcolsep}{5pt}
    \caption{\textbf{Hyperparameters.} Settings of training runs for \ours{} across model size.}
    \label{tab:sup:hyperparams}
    \resizebox{\linewidth}{!}{%
    \begin{tabular}{@{}lcccc@{}}
        \toprule
        \multicolumn{1}{c}{\textbf{Parameter}}& \textbf{\ours-T} & \textbf{\ours-S} & \textbf{\ours-B} & \textbf{\ours-L} \\
        \midrule
        Parameters (M) & 9 & 24 & 94 & 340 \\
        High-resolution layers (down \& up) & 2 & 2 & 2 & 2 \\
        Bottleneck & 6 & 8 & 8 & 20 \\
        High resolution hidden dimension & 192 & 256 & 512 & 768 \\
        Bottleneck hidden dimension & 256 & 384 & 768 & 1024 \\
        Attention heads (HiRes/Bottleneck) & 3/4 & 4/6 & 8/12 & 12/16 \\
        Normalization & \multicolumn{4}{c}{RMSNorm~\cite{zhang2019root}} \\
        FFN expansion factor & \multicolumn{4}{c}{$3\times$} \\
        Activations & \multicolumn{4}{c}{SwiGLU~\cite{shazeer2020glu}} \\
        Positional embeddings & \multicolumn{4}{c}{3D Axial RoPE~\cite{su2023RoFormer, heo2024Rotary}} \\
        Diffusion time mapping dimension & 128 & 128 & 384 & 384 \\
        Diffusion time mapping depth & 2 & 2 & 2 & 2 \\
        Diffusion time conditioning & \multicolumn{4}{c}{AdaNorm~\cite{peebles2023Scalable, xu2019understanding}} \\
        Input patch size {\tiny ($T\times H\times W$)} & \multicolumn{4}{c}{$1\times16\times16$} \\
        Total bottleneck patch size {\tiny ($T\times H\times W$)} & \multicolumn{4}{c}{$1\times32\times32$} \\
        High resolution attention window {\tiny ($T\times H\times W$)} & \multicolumn{4}{c}{$1 \times 5\times5$} \\
        \arrayrulecolor{gray!50!white}
        \midrule
        \arrayrulecolor{black}
        Tbar $\sigma$ & \multicolumn{4}{c}{0.1} \\
        Tbar sampling & \multicolumn{4}{c}{Uniform + $T\times H \times W$ shift} \\
        Maximum conditioning frames $k_{\text{max}}$ & \multicolumn{4}{c}{13} \\
        Unconditional probability $p_U$ & \multicolumn{4}{c}{0.25} \\
        \arrayrulecolor{gray!50!white}
        \midrule
        \arrayrulecolor{black}
        Training particles & 3 & 3 & 3 & 3 \\
        Expansion before layer & 8 & 10 & 10 & 22 \\
        $\xi$ embedding dimension & \multicolumn{4}{c}{256} \\
        $\xi$ conditioning & \multicolumn{4}{c}{Timestep-gated residual concatenation~\cite{martorella2026iddm}} \\
        \arrayrulecolor{gray!50!white}
        \midrule
        \arrayrulecolor{black}
        Training frame resolution & \multicolumn{4}{c}{$384 \times 384$} \\
        Training sequence length & \multicolumn{4}{c}{25} \\
        SEVIR training sequences & \multicolumn{4}{c}{402,500 (16,100 events $\times$ 25 windows)} \\
        \arrayrulecolor{gray!50!white}
        \midrule
        \arrayrulecolor{black}
        Optimizer & \multicolumn{4}{c}{AdamW~\cite{loshchilov2019Decoupled}} \\
        AdamW $(\beta_1,\beta_2)$ & \multicolumn{4}{c}{$(0.9,0.999)$} \\
        Learning rate & \multicolumn{4}{c}{$10^{-4}$} \\
        Learning-rate schedule & \multicolumn{4}{c}{Constant with 1,000 warmup steps} \\
        EMA decay & \multicolumn{4}{c}{0.9999} \\
        \arrayrulecolor{gray!50!white}
        \midrule
        \arrayrulecolor{black}
        Precision & \multicolumn{4}{c}{BF16 mixed precision} \\
        Global Batch size & \multicolumn{4}{c}{128} \\
        Training GPUs & 8 & 8 & 8 & 16 \\
        Training steps & 300k & 300k & 200k & 200k \\
        Training schedule & N/A & N/A & N/A & \makecell{\tiny 100k on $128^2$ \tiny+finetuning} \\
        \bottomrule
    \end{tabular}
    }
\end{table}

%% file: figs/4_exp/sevir_leadtime.tex
\begin{figure}[b]
    \centering
    \includegraphics[width=\textwidth]{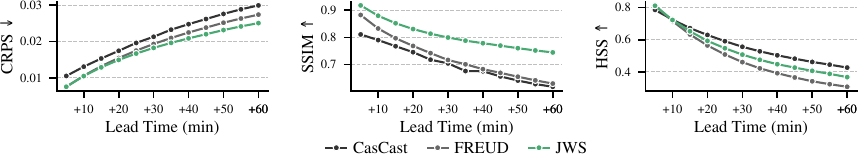}
    \caption{\textbf{Performance over lead time.} \ours-B~outperforms CasCast and FREUD across the entire lead time range for CRPS and SSIM, while achieving competitive performance for HSS.}
    \label{fig:exp:leadtime}
\end{figure}

%% file: figs/X_supp/qual_forecasts.tex
\begingroup

\newcommand{\forecastfigure}[2]{%
    \begin{figure}[htbp]
        \centering
        \setlength{\tabcolsep}{0.1pt}

        \newcommand{\imagesize}{0.11\textwidth}
        \newcommand{\gridimage}[3]{%
            \includegraphics[width=\imagesize]{#1/##1_##2_##3.png}%
        }
        \newcommand{\gridimagerow}[2]{%
            \gridimage{##1}{2}{##2} &
            \gridimage{##1}{4}{##2} &
            \gridimage{##1}{6}{##2} &
            \gridimage{##1}{8}{##2} &
            \gridimage{##1}{10}{##2} &
            \gridimage{##1}{12}{##2}%
        }
        \newcommand{\rowlabel}[1]{%
            \raisebox{%
                \dimexpr\imagesize/2-\height/2+\depth/2\relax
            }[0pt][0pt]{%
                \rotatebox{90}{\footnotesize ##1}%
            }%
        }

        \begin{tabular}{@{}ccccccc@{}}
            & +10\,min & +20\,min & +30\,min
            & +40\,min & +50\,min & +60\,min \\
            \rowlabel{E1}   & \gridimagerow{ens}{1}  \\
            \rowlabel{E2}   & \gridimagerow{ens}{2}  \\
            \rowlabel{E3}   & \gridimagerow{ens}{3}  \\
            \rowlabel{Mean} & \gridimagerow{mean}{1} \\
            \rowlabel{GT}   & \gridimagerow{gt}{1}
        \end{tabular}

        \caption{#2}
        \label{fig:forecast:#1}
    \end{figure}%
}

\forecastfigure{figs/X_supp/qualitative_forecasts/idx12149}{\textbf{Qualitative forecast.} We show three ensemble members over the full 60\,min lead time range. Zoom shows a high-precipitation region with high variability.}
\forecastfigure{figs/X_supp/qualitative_forecasts/idx3657}{\textbf{Qualitative forecast.} Three ensemble members for a sample from the SEVIR dataset.}
\forecastfigure{figs/X_supp/qualitative_forecasts/idx2286}{\textbf{Qualitative forecast.} Forecast steps in 10\,min intervals for a sample from the SEVIR dataset.}
\forecastfigure{figs/X_supp/qualitative_forecasts/idx6873}{\textbf{Qualitative forecast.} The zoom-in highlights high variability among ensemble members for a $60$\,min forecast.}

\endgroup

%% file: figs/X_supp/qual_comparisons.tex
\begingroup

\begin{figure}[h]
    \centering
    \setlength{\tabcolsep}{0.5pt}
    \newcommand{\imagesize}{0.15\textwidth}
    \newcommand{\imgpath}{figs/X_supp/qualitative_comparisons/idx520}
    \newcommand{\gridimage}[3]{\includegraphics[width=\imagesize]{\imgpath/#1_#2_#3.png}}
    \newcommand{\rowlabel}[1]{%
        \raisebox{%
            \dimexpr\imagesize/2-\height/2+\depth/2\relax
        }[0pt][0pt]{%
            \rotatebox{90}{#1}%
        }%
    }
    \newcommand{\gridimgrow}[2]{\gridimage{#1}{2}{#2}
        & \gridimage{#1}{4}{#2}
        & \gridimage{#1}{6}{#2}
        & \gridimage{#1}{8}{#2}
        & \gridimage{#1}{10}{#2}
        & \gridimage{#1}{12}{#2}}
    \begin{tabular}{ccccccc}
        & +10\,min & +20\,min & +30\,min & +40\,min & +50\,min & +60\,min \\
        \rowlabel{\footnotesize FGN$^\star$} & \gridimgrow{fgn}{1} \\
        \rowlabel{\footnotesize CasCast} & \gridimgrow{cascast}{1} \\
        \rowlabel{\footnotesize FlowCast} & \gridimgrow{flowcast}{1} \\
        \rowlabel{\footnotesize FREUD} & \gridimgrow{freud}{1} \\
        \rowlabel{\footnotesize JWS {\tiny (Ours)}} & \gridimgrow{ours}{3} \\
        \rowlabel{\footnotesize GT} & \gridimgrow{gt}{1}
    \end{tabular}
    \caption{\textbf{Qualitative comparison.} Compared to FGN~\cite{alet2025fgn}, CasCast~\cite{gong2024cascast}, FlowCast~\cite{ribeiro2026flowcast}, and FREUD~\cite{schusterbauer2026weatherrf-freud} \ours{} better captures local and global precipitation patterns. $\star$ model trained by us.}
    \label{fig:sup:qualitative_comp_3}
\end{figure}

\begin{figure}[h]
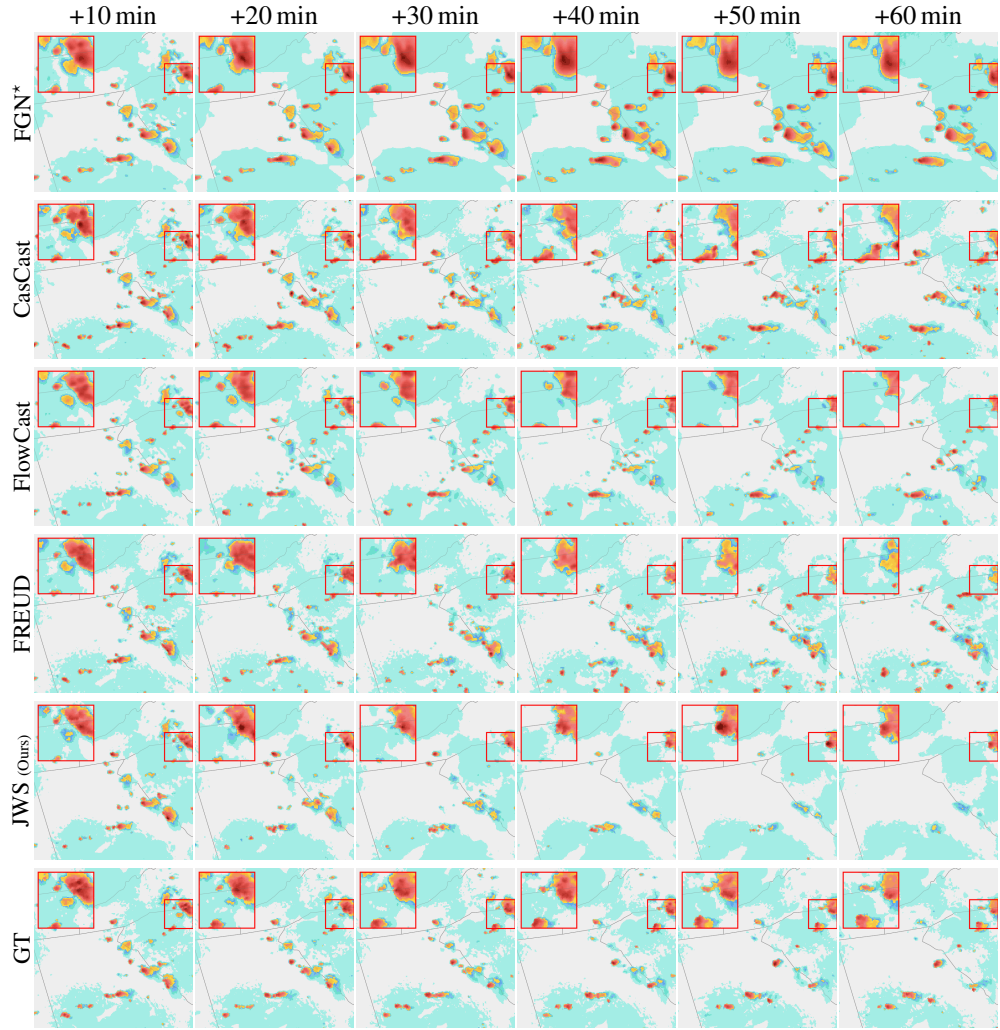

    \centering
    \setlength{\tabcolsep}{0.5pt}
    \newcommand{\imagesize}{0.15\textwidth}
    \newcommand{\imgpath}{figs/X_supp/qualitative_comparisons/idx1424}
    \newcommand{\gridimage}[3]{\includegraphics[width=\imagesize]{\imgpath/#1_#2_#3.png}}
    \newcommand{\rowlabel}[1]{%
        \raisebox{%
            \dimexpr\imagesize/2-\height/2+\depth/2\relax
        }[0pt][0pt]{%
            \rotatebox{90}{#1}%
        }%
    }
    \newcommand{\gridimgrow}[2]{\gridimage{#1}{2}{#2}
        & \gridimage{#1}{4}{#2}
        & \gridimage{#1}{6}{#2}
        & \gridimage{#1}{8}{#2}
        & \gridimage{#1}{10}{#2}
        & \gridimage{#1}{12}{#2}}
    \begin{tabular}{ccccccc}
        & +10\,min & +20\,min & +30\,min & +40\,min & +50\,min & +60\,min \\
        \rowlabel{\footnotesize FGN$^\star$} & \gridimgrow{fgn}{1} \\
        \rowlabel{\footnotesize CasCast} & \gridimgrow{cascast}{1} \\
        \rowlabel{\footnotesize FlowCast} & \gridimgrow{flowcast}{1} \\
        \rowlabel{\footnotesize FREUD} & \gridimgrow{freud}{1} \\
        \rowlabel{\footnotesize JWS {\tiny (Ours)}} & \gridimgrow{ours}{1} \\
        \rowlabel{\footnotesize GT} & \gridimgrow{gt}{1}
    \end{tabular}
    \caption{\textbf{Qualitative comparison.} FlowCast~\cite{ribeiro2026flowcast} and \ours{} capture localized high precipitation accurately. $\star$ model trained by us.}
    \label{fig:sup:qualitative_comp_4}
\end{figure}

\begin{figure}[t]
    \centering
    \vspace{0pt}
    \centering
    \setlength{\tabcolsep}{0.5pt}
    \newcommand{\imagesize}{0.15\textwidth}
    \newcommand{\imgpath}{figs/X_supp/qualitative_comparisons/idx488}
    \newcommand{\gridimage}[3]{\includegraphics[width=\imagesize]{\imgpath/#1_#2_#3.png}}
    \newcommand{\rowlabel}[1]{%
        \raisebox{%
            \dimexpr\imagesize/2-\height/2+\depth/2\relax
        }[0pt][0pt]{%
            \rotatebox{90}{#1}%
        }%
    }
    \newcommand{\gridimgrow}[2]{\gridimage{#1}{2}{#2}
        & \gridimage{#1}{4}{#2}
        & \gridimage{#1}{6}{#2}
        & \gridimage{#1}{8}{#2}
        & \gridimage{#1}{10}{#2}
        & \gridimage{#1}{12}{#2}}
    \begin{tabular}{ccccccc}
        & +10\,min & +20\,min & +30\,min & +40\,min & +50\,min & +60\,min \\
        \rowlabel{\footnotesize FGN$^\star$} & \gridimgrow{fgn}{1} \\
        \rowlabel{\footnotesize CasCast} & \gridimgrow{cascast}{1} \\
        \rowlabel{\footnotesize FlowCast} & \gridimgrow{flowcast}{1} \\
        \rowlabel{\footnotesize FREUD} & \gridimgrow{freud}{4} \\
        \rowlabel{\footnotesize JWS {\tiny (Ours)}} & \gridimgrow{ours}{4} \\
        \rowlabel{\footnotesize GT} & \gridimgrow{gt}{1}
    \end{tabular}

    \caption{\textbf{Qualitative comparison.}
        \ours~produces sharp forecasts that align with the ground truth and remains more realistic over leadtime than FREUD and CasCast.}
    \label{fig:exp:qualitative_comp_9}
\end{figure}

\begin{figure}[h]
    \centering
    \setlength{\tabcolsep}{0.5pt}
    \newcommand{\imagesize}{0.15\textwidth}
    \newcommand{\imgpath}{figs/X_supp/qualitative_comparisons/idx9195}
    \newcommand{\gridimage}[3]{\includegraphics[width=\imagesize]{\imgpath/#1_#2_#3.png}}
    \newcommand{\rowlabel}[1]{%
        \raisebox{%
            \dimexpr\imagesize/2-\height/2+\depth/2\relax
        }[0pt][0pt]{%
            \rotatebox{90}{#1}%
        }%
    }
    \newcommand{\gridimgrow}[2]{\gridimage{#1}{2}{#2}
        & \gridimage{#1}{4}{#2}
        & \gridimage{#1}{6}{#2}
        & \gridimage{#1}{8}{#2}
        & \gridimage{#1}{10}{#2}
        & \gridimage{#1}{12}{#2}}
    \begin{tabular}{ccccccc}
        & +10\,min & +20\,min & +30\,min & +40\,min & +50\,min & +60\,min \\
        \rowlabel{\footnotesize FGN$^\star$} & \gridimgrow{fgn}{1} \\
        \rowlabel{\footnotesize CasCast} & \gridimgrow{cascast}{1} \\
        \rowlabel{\footnotesize FlowCast} & \gridimgrow{flowcast}{1} \\
        \rowlabel{\footnotesize FREUD} & \gridimgrow{freud}{1} \\
        \rowlabel{\footnotesize JWS {\tiny (Ours)}} & \gridimgrow{ours}{1} \\
        \rowlabel{\footnotesize GT} & \gridimgrow{gt}{1}
    \end{tabular}
    \caption{\textbf{Qualitative comparison.} For low precipitation \ours{} is able to capture dynamics more accurately than FGN~\cite{alet2025fgn}, CasCast~\cite{gong2024cascast}, FlowCast~\cite{ribeiro2026flowcast}, or FREUD~\cite{schusterbauer2026weatherrf-freud}. $\star$ model trained by us.}
    \label{fig:sup:qualitative_comp_7}
\end{figure}

\begin{figure}[h]
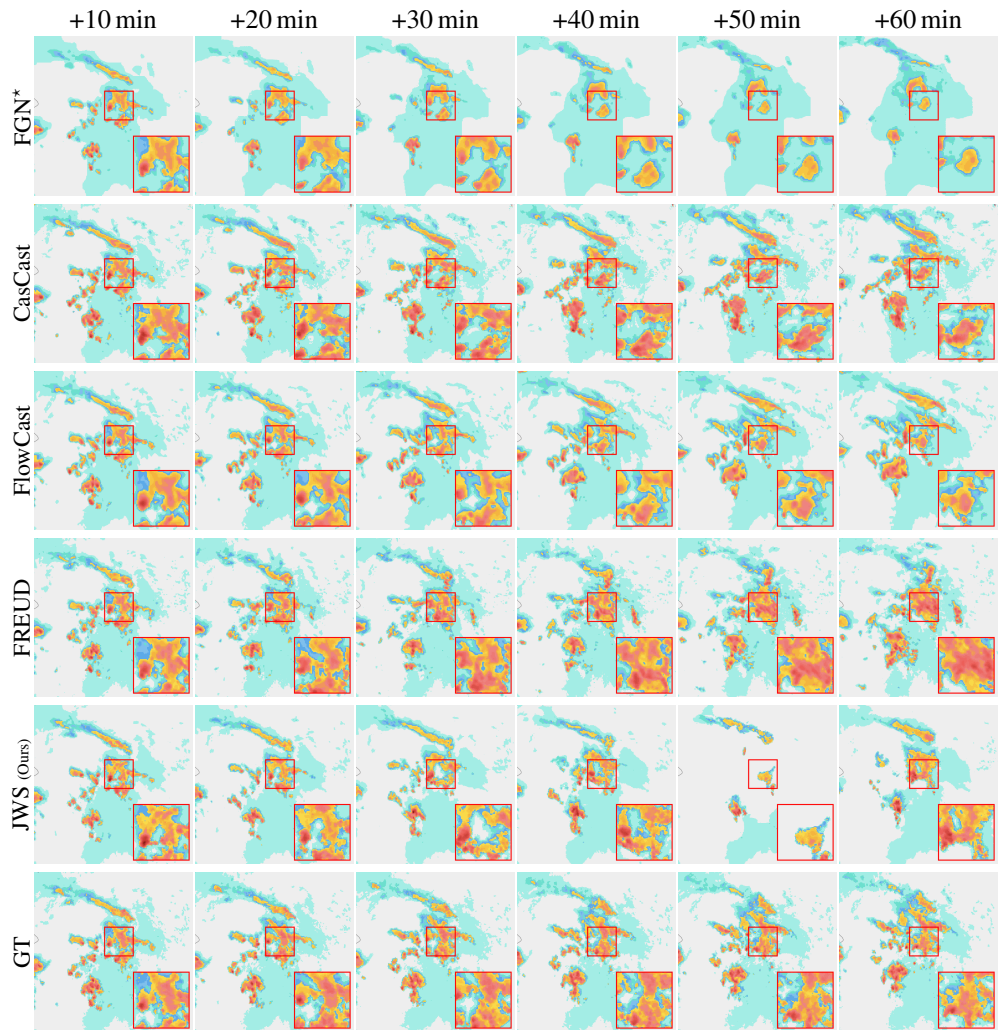

    \centering
    \setlength{\tabcolsep}{0.5pt}
    \newcommand{\imagesize}{0.15\textwidth}
    \newcommand{\imgpath}{figs/X_supp/qualitative_comparisons/idx10647}
    \newcommand{\gridimage}[3]{\includegraphics[width=\imagesize]{\imgpath/#1_#2_#3.png}}
    \newcommand{\rowlabel}[1]{%
        \raisebox{%
            \dimexpr\imagesize/2-\height/2+\depth/2\relax
        }[0pt][0pt]{%
            \rotatebox{90}{#1}%
        }%
    }
    \newcommand{\gridimgrow}[2]{\gridimage{#1}{2}{#2}
        & \gridimage{#1}{4}{#2}
        & \gridimage{#1}{6}{#2}
        & \gridimage{#1}{8}{#2}
        & \gridimage{#1}{10}{#2}
        & \gridimage{#1}{12}{#2}}
    \begin{tabular}{ccccccc}
        & +10\,min & +20\,min & +30\,min & +40\,min & +50\,min & +60\,min \\
        \rowlabel{\footnotesize FGN$^\star$} & \gridimgrow{fgn}{1} \\
        \rowlabel{\footnotesize CasCast} & \gridimgrow{cascast}{1} \\
        \rowlabel{\footnotesize FlowCast} & \gridimgrow{flowcast}{1} \\
        \rowlabel{\footnotesize FREUD} & \gridimgrow{freud}{1} \\
        \rowlabel{\footnotesize JWS {\tiny (Ours)}} & \gridimgrow{ours}{3} \\
        \rowlabel{\footnotesize GT} & \gridimgrow{gt}{1}
    \end{tabular}
    \caption{\textbf{Qualitative comparison.} \ours{} produces sharper and more intricate precipitation patterns than prior work. $\star$ model trained by us.}
    \label{fig:sup:qualitative_comp_8}
\end{figure}

\endgroup

%% file: figs/X_supp/qual_ensembles.tex
\begingroup

\begin{figure}[h]
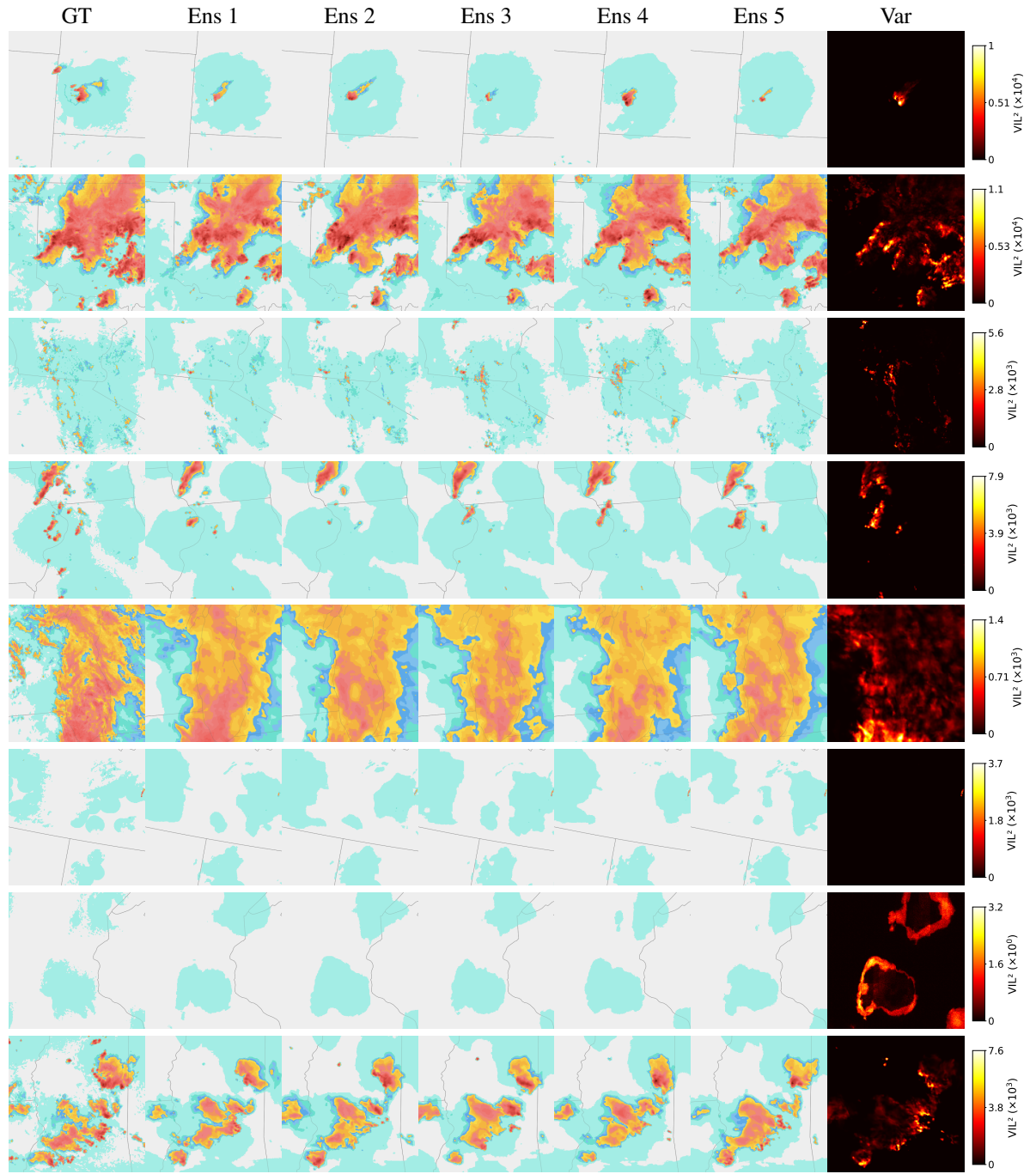

    \centering
    \setlength{\tabcolsep}{0.1pt}
    \newcommand{\imagesize}{0.15\textwidth}
    \newcommand{\imgpath}{figs/X_supp/qualitative_ensembles}
    \newcommand{\gridimage}[4]{\includegraphics[width=\imagesize]{\imgpath/idx#1/#2_#3_#4.png}}
    \newcommand{\gridimagerow}[1]{\gridimage{#1}{gt}{12}{1} & \gridimage{#1}{ens}{12}{1} & \gridimage{#1}{ens}{12}{2} & \gridimage{#1}{ens}{12}{3} & \gridimage{#1}{ens}{12}{4} & \gridimage{#1}{ens}{12}{5} & \gridimage{#1}{var}{12}{1} & \colorbar{#1}{12} \\}
    \newcommand{\colorbar}[2]{\includegraphics[height=0.14\textwidth]{\imgpath/idx#1/colorbar_#2.pdf}}
    \begin{tabular}{ccccccc@{\hspace{3pt}}l}
        GT & Ens 1 & Ens 2 & Ens 3 & Ens 4 & Ens 5 & Var & \\
        \gridimagerow{12066}
        \gridimagerow{11087}
        \gridimagerow{8935}
        \gridimagerow{222}
        \gridimagerow{5283}
        \gridimagerow{3006}
        \gridimagerow{9753}
        \gridimagerow{10712}
    \end{tabular}
    \caption{\textbf{Ensemble variance.} We show five ensemble members for samples from the SEVIR dataset at $+60\,\text{min}$ with complex precipitation patterns. Variance maps highlight localized uncertainty in complex precipitation regions.}
    \label{fig:sup:qualitative_ens}
\end{figure}

\endgroup

%% file: figs/X_supp/qual_nfe.tex
\begin{figure}[h]
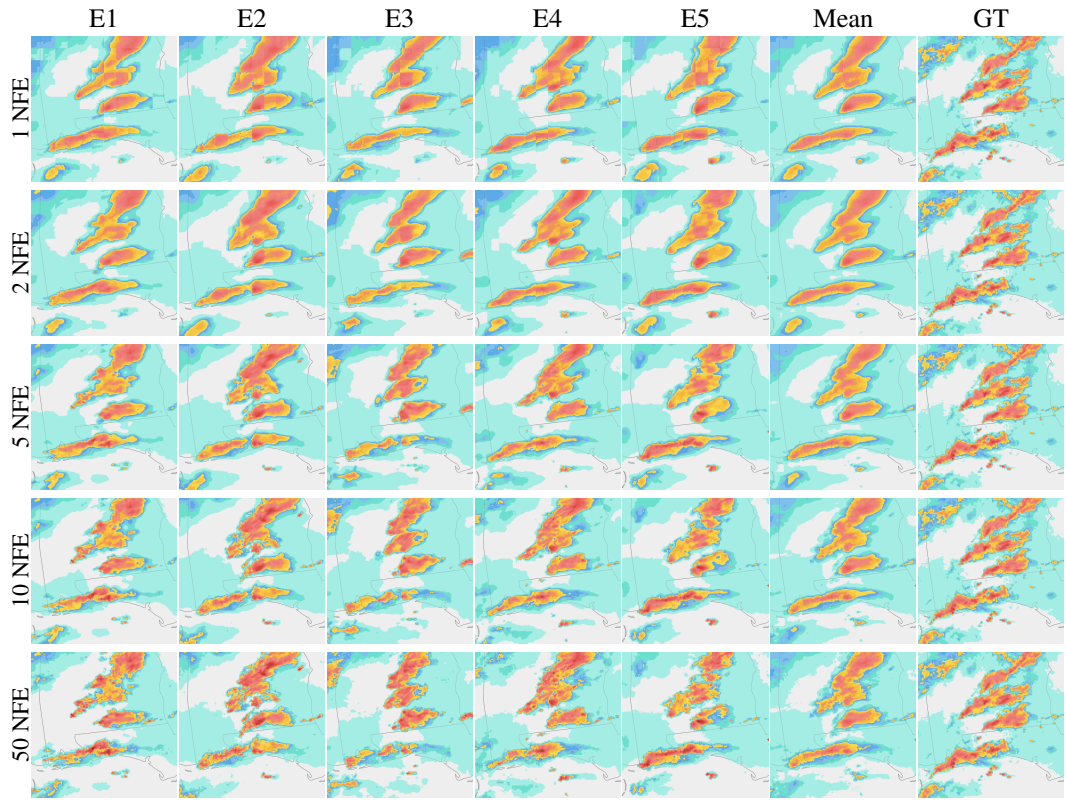

    \centering
    \vspace{0pt}
    \centering
    \setlength{\tabcolsep}{0.5pt}
    \newcommand{\imagesize}{0.1375\textwidth}
    \newcommand{\imgpath}{figs/X_supp/qualitative_nfe/idx7347}
    \newcommand{\gridimage}[4]{\includegraphics[width=\imagesize]{\imgpath/nfe#1/#2_#3_#4.png}}
    \newcommand{\gtimage}[0]{\gridimage{1}{gt}{12}{1}}
    \newcommand{\rowlabel}[1]{%
        \raisebox{%
            \dimexpr\imagesize/2-\height/2+\depth/2\relax
        }[0pt][0pt]{%
            \rotatebox{90}{#1}%
        }%
    }
    \newcommand{\gridimgrow}[1]{%
        \gridimage{#1}{ens}{12}{1} & 
        \gridimage{#1}{ens}{12}{2} & 
        \gridimage{#1}{ens}{12}{3} &
        \gridimage{#1}{ens}{12}{4} &
        \gridimage{#1}{ens}{12}{5} &
        \gridimage{#1}{mean}{12}{1} &
        \gtimage
    }
    \begin{tabular}{cccccccc}
        & E1 & E2 & E3 & E4 & E5 & Mean & GT \\
        \rowlabel{\footnotesize 1 NFE} & \gridimgrow{1} \\
        \rowlabel{\footnotesize 2 NFE} & \gridimgrow{2} \\
        \rowlabel{\footnotesize 5 NFE} & \gridimgrow{5} \\
        \rowlabel{\footnotesize 10 NFE} & \gridimgrow{10} \\
        \rowlabel{\footnotesize 50 NFE} & \gridimgrow{50}
    \end{tabular}

    \caption{\textbf{Qualitative Inference-time scaling.} At $+60\,\text{min}$ \ours{} produce accurate global structure with 1 NFE, 2 NFE resolves patch artifacts, and 10 NFE is close to the 50 NFE setting, with 50 NFE adding more local details and variability.}
    \label{fig:exp:qualitative_nfe_1}
\end{figure}

\begin{figure}[h]
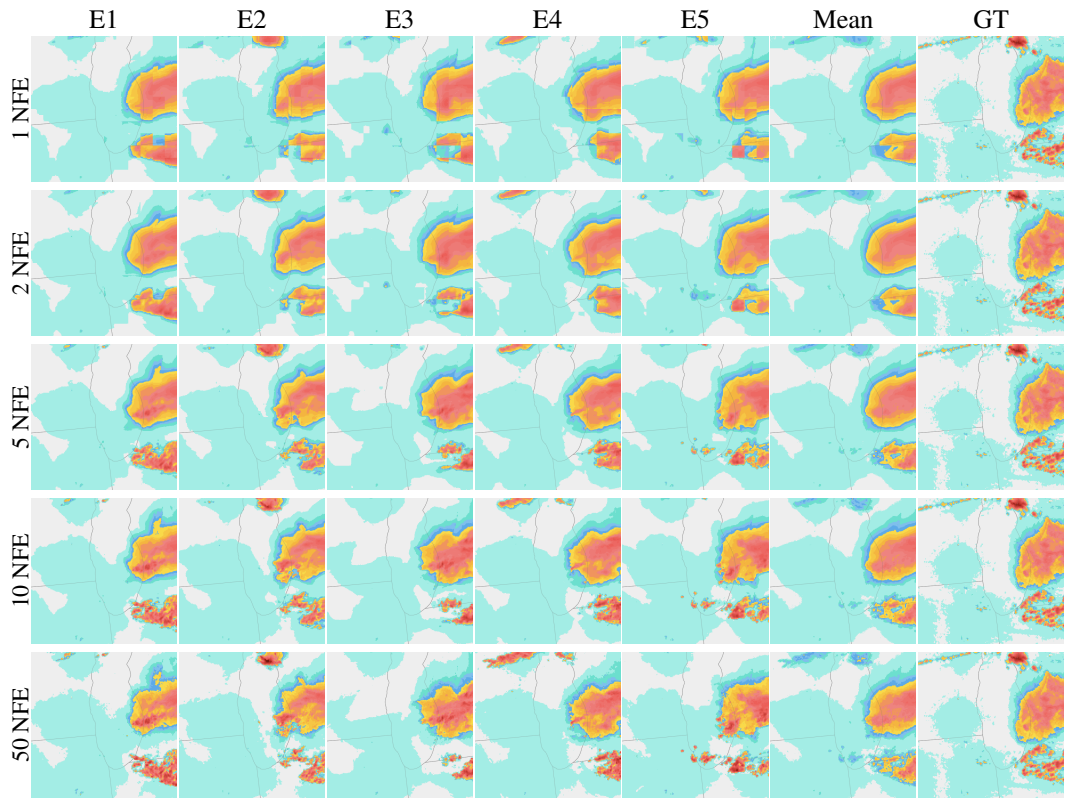

    \centering
    \vspace{0pt}
    \centering
    \setlength{\tabcolsep}{0.5pt}
    \newcommand{\imagesize}{0.1375\textwidth}
    \newcommand{\imgpath}{figs/X_supp/qualitative_nfe/idx812}
    \newcommand{\gridimage}[4]{\includegraphics[width=\imagesize]{\imgpath/nfe#1/#2_#3_#4.png}}
    \newcommand{\gtimage}[0]{\gridimage{1}{gt}{12}{1}}
    \newcommand{\rowlabel}[1]{%
        \raisebox{%
            \dimexpr\imagesize/2-\height/2+\depth/2\relax
        }[0pt][0pt]{%
            \rotatebox{90}{#1}%
        }%
    }
    \newcommand{\gridimgrow}[1]{%
        \gridimage{#1}{ens}{12}{1} & 
        \gridimage{#1}{ens}{12}{2} & 
        \gridimage{#1}{ens}{12}{3} &
        \gridimage{#1}{ens}{12}{4} &
        \gridimage{#1}{ens}{12}{5} &
        \gridimage{#1}{mean}{12}{1} &
        \gtimage
    }
    \begin{tabular}{cccccccc}
        & E1 & E2 & E3 & E4 & E5 & Mean & GT \\
        \rowlabel{\footnotesize 1 NFE} & \gridimgrow{1} \\
        \rowlabel{\footnotesize 2 NFE} & \gridimgrow{2} \\
        \rowlabel{\footnotesize 5 NFE} & \gridimgrow{5} \\
        \rowlabel{\footnotesize 10 NFE} & \gridimgrow{10} \\
        \rowlabel{\footnotesize 50 NFE} & \gridimgrow{50}
    \end{tabular}

    \caption{\textbf{Qualitative Inference-time scaling.} \ours{} captures finer structures consistently with more function evaluations at $+60\,\text{min}$ lead time.}
    \label{fig:exp:qualitative_nfe_2}
\end{figure}

\begin{figure}
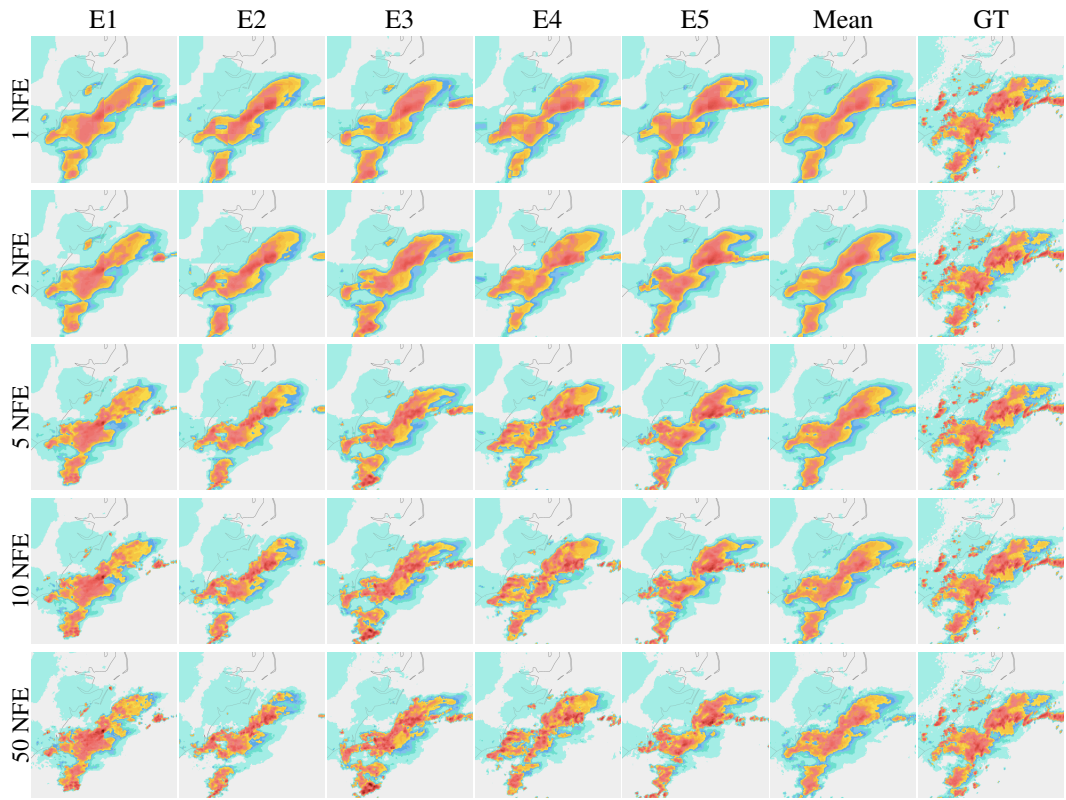

    \centering
    \vspace{0pt}
    \centering
    \setlength{\tabcolsep}{0.5pt}
    \newcommand{\imagesize}{0.1375\textwidth}
    \newcommand{\imgpath}{figs/X_supp/qualitative_nfe/idx7773}
    \newcommand{\gridimage}[4]{\includegraphics[width=\imagesize]{\imgpath/nfe#1/#2_#3_#4.png}}
    \newcommand{\gtimage}[0]{\gridimage{1}{gt}{12}{1}}
    \newcommand{\rowlabel}[1]{%
        \raisebox{%
            \dimexpr\imagesize/2-\height/2+\depth/2\relax
        }[0pt][0pt]{%
            \rotatebox{90}{#1}%
        }%
    }
    \newcommand{\gridimgrow}[1]{%
        \gridimage{#1}{ens}{12}{1} & 
        \gridimage{#1}{ens}{12}{2} & 
        \gridimage{#1}{ens}{12}{3} &
        \gridimage{#1}{ens}{12}{4} &
        \gridimage{#1}{ens}{12}{5} &
        \gridimage{#1}{mean}{12}{1} &
        \gtimage
    }
    \begin{tabular}{cccccccc}
        & E1 & E2 & E3 & E4 & E5 & Mean & GT \\
        \rowlabel{\footnotesize 1 NFE} & \gridimgrow{1} \\
        \rowlabel{\footnotesize 2 NFE} & \gridimgrow{2} \\
        \rowlabel{\footnotesize 5 NFE} & \gridimgrow{5} \\
        \rowlabel{\footnotesize 10 NFE} & \gridimgrow{10} \\
        \rowlabel{\footnotesize 50 NFE} & \gridimgrow{50}
    \end{tabular}

    \caption{\textbf{Qualitative inference-time scaling.} Results at $+60\,\text{min}$ lead time show that forecasts become blurry with few-step inference, while global structure is captured in all settings, and more sampling steps recover details.}
    \label{fig:exp:qualitative_nfe_3}
\end{figure}

%% file: main.bbl
\begin{thebibliography}{104}
\providecommand{\natexlab}[1]{#1}
\providecommand{\url}[1]{\texttt{#1}}
\expandafter\ifx\csname urlstyle\endcsname\relax
  \providecommand{\doi}[1]{doi: #1}\else
  \providecommand{\doi}{doi: \begingroup \urlstyle{rm}\Url}\fi

\bibitem[Agrawal et~al.(2019)Agrawal, Barrington, Bromberg, Burge, Gazen, and Hickey]{agrawal19:UNet}
Shreya Agrawal, Luke Barrington, Carla Bromberg, John Burge, Cenk Gazen, and Jason Hickey.
\newblock Machine learning for precipitation nowcasting from radar images.
\newblock \emph{arXiv preprint arXiv:1912.12132}, 2019.

\bibitem[Albergo et~al.(2023)Albergo, Boffi, and Vanden-Eijnden]{albergo2023stochastic}
Michael~S Albergo, Nicholas~M Boffi, and Eric Vanden-Eijnden.
\newblock Stochastic interpolants: A unifying framework for flows and diffusions.
\newblock \emph{arXiv preprint arXiv:2303.08797}, 2023.

\bibitem[Alet et~al.(2025)Alet, Price, {El-Kadi}, Masters, Markou, Andersson, Stott, Lam, Willson, {Sanchez-Gonzalez}, and Battaglia]{alet2025fgn}
Ferran Alet, Ilan Price, Andrew {El-Kadi}, Dominic Masters, Stratis Markou, Tom~R. Andersson, Jacklynn Stott, Remi Lam, Matthew Willson, Alvaro {Sanchez-Gonzalez}, and Peter Battaglia.
\newblock Skillful joint probabilistic weather forecasting from marginals, 2025.

\bibitem[Biroli et~al.(2024)Biroli, Bonnaire, de~Bortoli, and M{\'e}zard]{biroli2024dynamical}
Giulio Biroli, Tony Bonnaire, Valentin de~Bortoli, and Marc M{\'e}zard.
\newblock Dynamical regimes of diffusion models, 2024.

\bibitem[Boffi et~al.(2025{\natexlab{a}})Boffi, Albergo, and {Vanden-Eijnden}]{boffi2025flow}
Nicholas~M. Boffi, Michael~S. Albergo, and Eric {Vanden-Eijnden}.
\newblock Flow map matching with stochastic interpolants: A mathematical framework for consistency models, 2025{\natexlab{a}}.

\bibitem[Boffi et~al.(2025{\natexlab{b}})Boffi, Albergo, and {Vanden-Eijnden}]{boffi2025how}
Nicholas~M. Boffi, Michael~S. Albergo, and Eric {Vanden-Eijnden}.
\newblock How to build a consistency model: Learning flow maps via self-distillation, 2025{\natexlab{b}}.

\bibitem[Chan et~al.(2024)Chan, Molina, and Metzler]{chan2024HyperDiffusion}
Matthew~A. Chan, Maria~J. Molina, and Christopher~A. Metzler.
\newblock Hyper-{{Diffusion}}: {{Estimating Epistemic}} and {{Aleatoric Uncertainty}} with a {{Single Model}}, February 2024.

\bibitem[Chen et~al.(2024)Chen, Monso, Du, Simchowitz, Tedrake, and Sitzmann]{chen2024diffusionforcing}
Boyuan Chen, Diego~Marti Monso, Yilun Du, Max Simchowitz, Russ Tedrake, and Vincent Sitzmann.
\newblock Diffusion {{Forcing}}: Next-token prediction meets full-sequence diffusion, December 2024.

\bibitem[Chen et~al.(2023)Chen, Long, Jiang, Liu, and Zhang]{chen2023Foundation}
Shengchao Chen, Guodong Long, Jing Jiang, Dikai Liu, and Chengqi Zhang.
\newblock Foundation {{Models}} for {{Weather}} and {{Climate Data Understanding}}: {{A Comprehensive Survey}}, December 2023.

\bibitem[Chen et~al.(2025)Chen, Ge, Zhang, Sun, and Luo]{chen2025pixelflow}
Shoufa Chen, Chongjian Ge, Shilong Zhang, Peize Sun, and Ping Luo.
\newblock Pixelflow: Pixel-space generative models with flow, 2025.

\bibitem[Chen(2023)]{chen2023noiseschedule}
Ting Chen.
\newblock On the importance of noise scheduling for diffusion models.
\newblock \emph{arXiv preprint arXiv:2301.10972}, 2023.

\bibitem[Chen \& Li(2023)Chen and Li]{chen2023fit}
Ting Chen and Lala Li.
\newblock Fit: Far-reaching interleaved transformers, 2023.

\bibitem[Chen et~al.(2026)Chen, Zhu, Chen, Zhang, Hu, Zhao, Wang, Yang, and Tai]{chen2026dip}
Zhennan Chen, Junwei Zhu, Xu~Chen, Jiangning Zhang, Xiaobin Hu, Hanzhen Zhao, Chengjie Wang, Jian Yang, and Ying Tai.
\newblock Dip: Taming diffusion models in pixel space, 2026.

\bibitem[Cheung \& Yeung(2012)Cheung and Yeung]{cheung2012Application}
P.~Cheung and H.Y. Yeung.
\newblock Application of optical-flow technique to significant convection nowcast for terminal areas in {{Hong Kong}}.
\newblock In \emph{The 3rd {{WMO International Symposium}} on {{Nowcasting}} and {{Very Short-Range Forecastin}}}, Rio de Janeiro, Brazil, 2012. WMO.

\bibitem[Crowson et~al.(2024)Crowson, Baumann, Birch, Abraham, Kaplan, and Shippole]{crowson2024hdit}
Katherine Crowson, Stefan~Andreas Baumann, Alex Birch, Tanishq~Mathew Abraham, Daniel~Z. Kaplan, and Enrico Shippole.
\newblock Scalable {{High-Resolution Pixel-Space Image Synthesis}} with {{Hourglass Diffusion Transformers}}.
\newblock In \emph{Proceedings of the 41st {{International Conference}} on {{Machine Learning}}}, pp.\  9550--9575. PMLR, 2024.

\bibitem[{De Bortoli} et~al.(2025){De Bortoli}, Galashov, Guntupalli, Zhou, Murphy, Gretton, and Doucet]{bortoli2025distributional}
Valentin {De Bortoli}, Alexandre Galashov, J~Swaroop Guntupalli, Guangyao Zhou, Kevin~Patrick Murphy, Arthur Gretton, and Arnaud Doucet.
\newblock Distributional diffusion models with scoring rules.
\newblock In \emph{Forty-second International Conference on Machine Learning}, 2025.
\newblock URL \url{https://openreview.net/forum?id=N82967FcVK}.

\bibitem[Delle~Monache et~al.(2006)Delle~Monache, Hacker, Zhou, Deng, and Stull]{dellemonache2006Probabilistic}
Luca Delle~Monache, Joshua~P. Hacker, Yongmei Zhou, Xingxiu Deng, and Roland~B. Stull.
\newblock Probabilistic aspects of meteorological and ozone regional ensemble forecasts.
\newblock \emph{Journal of Geophysical Research: Atmospheres}, 111\penalty0 (D24):\penalty0 2005JD006917, December 2006.
\newblock ISSN 0148-0227.
\newblock \doi{10.1029/2005JD006917}.

\bibitem[Dhariwal \& Nichol(2021)Dhariwal and Nichol]{dhariwal2021beatGANs}
Prafulla Dhariwal and Alexander Nichol.
\newblock Diffusion models beat gans on image synthesis.
\newblock \emph{Advances in Neural Information Processing Systems}, 34:\penalty0 8780--8794, 2021.

\bibitem[Dosovitskiy et~al.(2020)Dosovitskiy, Beyer, Kolesnikov, Weissenborn, Zhai, Unterthiner, Dehghani, Minderer, Heigold, Gelly, et~al.]{dosovitskiy2020vit}
Alexey Dosovitskiy, Lucas Beyer, Alexander Kolesnikov, Dirk Weissenborn, Xiaohua Zhai, Thomas Unterthiner, Mostafa Dehghani, Matthias Minderer, G~Heigold, S~Gelly, et~al.
\newblock An image is worth 16x16 words: Transformers for image recognition at scale.
\newblock In \emph{ICLR}, 2020.

\bibitem[Espeholt et~al.(2022)Espeholt, Agrawal, S{\o}nderby, Kumar, Heek, Bromberg, Gazen, Carver, Andrychowicz, Hickey, et~al.]{espeholt22:MetNet2}
Lasse Espeholt, Shreya Agrawal, Casper S{\o}nderby, Manoj Kumar, Jonathan Heek, Carla Bromberg, Cenk Gazen, Rob Carver, Marcin Andrychowicz, Jason Hickey, et~al.
\newblock Deep learning for twelve hour precipitation forecasts.
\newblock \emph{Nature communications}, 13\penalty0 (1):\penalty0 5145, 2022.

\bibitem[Esser et~al.(2024)Esser, Kulal, Blattmann, Entezari, M{\"u}ller, Saini, Levi, Lorenz, Sauer, Boesel, et~al.]{esser2024scaling}
Patrick Esser, Sumith Kulal, Andreas Blattmann, Rahim Entezari, Jonas M{\"u}ller, Harry Saini, Yam Levi, Dominik Lorenz, Axel Sauer, Frederic Boesel, et~al.
\newblock Scaling rectified flow transformers for high-resolution image synthesis.
\newblock In \emph{ICML}, 2024.

\bibitem[Frans et~al.(2025)Frans, Hafner, Levine, and Abbeel]{frans2025one}
Kevin Frans, Danijar Hafner, Sergey Levine, and Pieter Abbeel.
\newblock One step diffusion via shortcut models, 2025.

\bibitem[Gandikota \& Bau(2025)Gandikota and Bau]{gandikota2025distilling}
Rohit Gandikota and David Bau.
\newblock Distilling diversity and control in diffusion models, 2025.

\bibitem[Gao et~al.(2022{\natexlab{a}})Gao, Tan, Wu, and Li]{gao2022simvp}
Zhangyang Gao, Cheng Tan, Lirong Wu, and Stan~Z Li.
\newblock Simvp: Simpler yet better video prediction.
\newblock In \emph{Proceedings of the IEEE/CVF Conference on Computer Vision and Pattern Recognition}, pp.\  3170--3180, 2022{\natexlab{a}}.

\bibitem[Gao et~al.(2022{\natexlab{b}})Gao, Shi, Wang, Zhu, Wang, Li, and Yeung]{gao2022earthformer}
Zhihan Gao, Xingjian Shi, Hao Wang, Yi~Zhu, Yuyang~Bernie Wang, Mu~Li, and Dit-Yan Yeung.
\newblock Earthformer: Exploring space-time transformers for earth system forecasting.
\newblock \emph{Advances in Neural Information Processing Systems}, 35:\penalty0 25390--25403, 2022{\natexlab{b}}.

\bibitem[Gao et~al.(2023)Gao, Shi, Han, Wang, Jin, Maddix, Zhu, Li, and Wang]{gao2023prediff}
Zhihan Gao, Xingjian Shi, Boran Han, Hao Wang, Xiaoyong Jin, Danielle Maddix, Yi~Zhu, Mu~Li, and Yuyang~(Bernie) Wang.
\newblock Prediff: Precipitation nowcasting with latent diffusion models.
\newblock In A.~Oh, T.~Naumann, A.~Globerson, K.~Saenko, M.~Hardt, and S.~Levine (eds.), \emph{Advances in Neural Information Processing Systems}, volume~36, pp.\  78621--78656. Curran Associates, Inc., 2023.
\newblock URL \url{https://proceedings.neurips.cc/paper_files/paper/2023/file/f82ba6a6b981fbbecf5f2ee5de7db39c-Paper-Conference.pdf}.

\bibitem[Geng et~al.(2025)Geng, Deng, Bai, Kolter, and He]{geng2025mean}
Zhengyang Geng, Mingyang Deng, Xingjian Bai, J.~Zico Kolter, and Kaiming He.
\newblock Mean flows for one-step generative modeling, 2025.

\bibitem[Geng et~al.(2026)Geng, Lu, Wu, Shechtman, Kolter, and He]{geng2026improved}
Zhengyang Geng, Yiyang Lu, Zongze Wu, Eli Shechtman, J.~Zico Kolter, and Kaiming He.
\newblock Improved mean flows: On the challenges of fastforward generative models, 2026.

\bibitem[Germann \& Zawadzki(2002)Germann and Zawadzki]{germann2002ScaleDependence}
Urs Germann and Isztar Zawadzki.
\newblock Scale-{{Dependence}} of the {{Predictability}} of {{Precipitation}} from {{Continental Radar Images}}. {{Part I}}: {{Description}} of the {{Methodology}}.
\newblock \emph{Monthly Weather Review}, 130\penalty0 (12):\penalty0 2859--2873, December 2002.
\newblock ISSN 1520-0493, 0027-0644.
\newblock \doi{10.1175/1520-0493(2002)130<2859:SDOTPO>2.0.CO;2}.

\bibitem[Gneiting \& Raftery(2007)Gneiting and Raftery]{gneiting2007strictly}
Tilmann Gneiting and Adrian~E. Raftery.
\newblock Strictly proper scoring rules, prediction, and estimation.
\newblock \emph{Journal of the American statistical Association}, 102\penalty0 (477):\penalty0 359--378, 2007.
\newblock ISBN: 0162-1459.

\bibitem[Gong et~al.(2024)Gong, Bai, Ye, Xu, Liu, Dai, Yang, and Ouyang]{gong2024cascast}
Junchao Gong, Lei Bai, Peng Ye, Wanghan Xu, Na~Liu, Jianhua Dai, Xiaokang Yang, and Wanli Ouyang.
\newblock {{CasCast}}: Skillful high-resolution precipitation nowcasting via cascaded modelling.
\newblock In \emph{Proceedings of the 41st {{International Conference}} on {{Machine Learning}}}, volume 235 of \emph{{{ICML}}'24}, pp.\  15809--15822, Vienna, Austria, 2024. JMLR.org.

\bibitem[Goodfellow et~al.(2014)Goodfellow, Pouget-Abadie, Mirza, Xu, Warde-Farley, Ozair, Courville, and Bengio]{goodfellow2014gan}
Ian Goodfellow, Jean Pouget-Abadie, Mehdi Mirza, Bing Xu, David Warde-Farley, Sherjil Ozair, Aaron Courville, and Yoshua Bengio.
\newblock Generative adversarial nets.
\newblock In Z.~Ghahramani, M.~Welling, C.~Cortes, N.~Lawrence, and K.Q. Weinberger (eds.), \emph{Advances in Neural Information Processing Systems}, volume~27. Curran Associates, Inc., 2014.
\newblock URL \url{https://proceedings.neurips.cc/paper/2014/file/5ca3e9b122f61f8f06494c97b1afccf3-Paper.pdf}.

\bibitem[Guen \& Thome(2020)Guen and Thome]{guen2020disentangling}
Vincent~Le Guen and Nicolas Thome.
\newblock Disentangling physical dynamics from unknown factors for unsupervised video prediction.
\newblock In \emph{Proceedings of the IEEE/CVF Conference on Computer Vision and Pattern Recognition}, pp.\  11474--11484, 2020.

\bibitem[Guo et~al.(2025)Guo, Wang, Yuan, Cao, Chen, Chen, Huo, Zhang, Wang, Liu, and Wang]{guo2025splitmeanflow}
Yi~Guo, Wei Wang, Zhihang Yuan, Rong Cao, Kuan Chen, Zhengyang Chen, Yuanyuan Huo, Yang Zhang, Yuping Wang, Shouda Liu, and Yuxuan Wang.
\newblock {SplitMeanFlow}: {Interval} {Splitting} {Consistency} in {Few}-{Step} {Generative} {Modeling}, July 2025.
\newblock URL \url{http://arxiv.org/abs/2507.16884}.
\newblock arXiv:2507.16884 [cs] version: 1.

\bibitem[Hassani et~al.(2023)Hassani, Walton, Li, Li, and Shi]{hassani2023Neighborhood}
Ali Hassani, Steven Walton, Jiachen Li, Shen Li, and Humphrey Shi.
\newblock Neighborhood {{Attention Transformer}}.
\newblock In \emph{Proceedings of the {{IEEE}}/{{CVF Conference}} on {{Computer Vision}} and {{Pattern Recognition}}}, pp.\  6185--6194, 2023.

\bibitem[He et~al.(2023)He, Yang, Zhang, Shan, and Chen]{he2023Latenta}
Yingqing He, Tianyu Yang, Yong Zhang, Ying Shan, and Qifeng Chen.
\newblock Latent {{Video Diffusion Models}} for {{High-Fidelity Long Video Generation}}, March 2023.

\bibitem[Heo et~al.(2024)Heo, Park, Han, and Yun]{heo2024Rotary}
Byeongho Heo, Song Park, Dongyoon Han, and Sangdoo Yun.
\newblock Rotary {{Position Embedding}} for {{Vision Transformer}}, 2024.

\bibitem[Ho \& Salimans(2022)Ho and Salimans]{ho2022ClassifierFree}
Jonathan Ho and Tim Salimans.
\newblock Classifier-{{Free Diffusion Guidance}}, 2022.

\bibitem[Ho et~al.(2020)Ho, Jain, and Abbeel]{ho2020ddpm}
Jonathan Ho, Ajay Jain, and Pieter Abbeel.
\newblock Denoising diffusion probabilistic models.
\newblock \emph{Advances in Neural Information Processing Systems}, 33:\penalty0 6840--6851, 2020.

\bibitem[Ho et~al.(2022)Ho, Salimans, Gritsenko, Chan, Norouzi, and Fleet]{ho2022video}
Jonathan Ho, Tim Salimans, Alexey Gritsenko, William Chan, Mohammad Norouzi, and David~J Fleet.
\newblock Video diffusion models.
\newblock \emph{arXiv preprint arXiv:2204.03458}, 2022.

\bibitem[Hoffmann et~al.(2022)Hoffmann, Borgeaud, Mensch, Buchatskaya, Cai, Rutherford, Casas, Hendricks, Welbl, Clark, et~al.]{hoffmann2022training}
Jordan Hoffmann, Sebastian Borgeaud, Arthur Mensch, Elena Buchatskaya, Trevor Cai, Eliza Rutherford, Diego de~Las Casas, Lisa~Anne Hendricks, Johannes Welbl, Aidan Clark, et~al.
\newblock Training compute-optimal large language models.
\newblock \emph{arXiv preprint arXiv:2203.15556}, 2022.

\bibitem[Hoogeboom et~al.(2023)Hoogeboom, Heek, and Salimans]{hoogeboom2023simple}
Emiel Hoogeboom, Jonathan Heek, and Tim Salimans.
\newblock simple diffusion: End-to-end diffusion for high resolution images.
\newblock \emph{arXiv preprint arXiv:2301.11093}, 2023.

\bibitem[Hoogeboom et~al.(2025)Hoogeboom, Mensink, Heek, Lamerigts, Gao, and Salimans]{hoogeboom2025simplera}
Emiel Hoogeboom, Thomas Mensink, Jonathan Heek, Kay Lamerigts, Ruiqi Gao, and Tim Salimans.
\newblock Simpler diffusion (sid2): 1.5 fid on imagenet512 with pixel-space diffusion, 2025.

\bibitem[H{\"o}ppe et~al.(2022)H{\"o}ppe, Mehrjou, Bauer, Nielsen, and Dittadi]{hoppe2022ramvid}
Tobias H{\"o}ppe, Arash Mehrjou, Stefan Bauer, Didrik Nielsen, and Andrea Dittadi.
\newblock Diffusion models for video prediction and infilling.
\newblock \emph{arXiv preprint arXiv:2206.07696}, 2022.

\bibitem[Huang et~al.(2026)Huang, Wang, Bertozzi, and Wang]{huang2026rmflow}
Yuhao Huang, Shih-Hsin Wang, Andrea~L. Bertozzi, and Bao Wang.
\newblock Rmflow: Refined mean flow by a noise-injection step for multimodal generation, 2026.

\bibitem[Hur \& Roth(2019)Hur and Roth]{hur2019Iterative}
Junhwa Hur and Stefan Roth.
\newblock Iterative {{Residual Refinement}} for {{Joint Optical Flow}} and {{Occlusion Estimation}}, 2019.

\bibitem[Jabri et~al.(2023)Jabri, Fleet, and Chen]{jabri2023rin}
Allan Jabri, David~J. Fleet, and Ting Chen.
\newblock Scalable adaptive computation for iterative generation.
\newblock In \emph{Proceedings of the 40th International Conference on Machine Learning}, pp.\  14569--14589. PMLR, 2023.

\bibitem[Ji et~al.(2022)Ji, Gong, Langguth, Mozaffari, and Zhi]{ji22:CLGAN}
Yan Ji, Bing Gong, Michael Langguth, Amirpasha Mozaffari, and Xiefei Zhi.
\newblock Clgan: A gan-based video prediction model for precipitation nowcasting.
\newblock \emph{EGUsphere}, pp.\  1--23, 2022.

\bibitem[Kossale et~al.(2022)Kossale, Airaj, and Darouichi]{kossale2022Mode}
Youssef Kossale, Mohammed Airaj, and Aziz Darouichi.
\newblock Mode {{Collapse}} in {{Generative Adversarial Networks}}: {{An Overview}}.
\newblock In \emph{2022 8th {{International Conference}} on {{Optimization}} and {{Applications}} ({{ICOA}})}, pp.\  1--6, October 2022.
\newblock \doi{10.1109/ICOA55659.2022.9934291}.

\bibitem[Larvor et~al.(2020)Larvor, Berthomier, Chabot, Le~Pape, Pradel, and Perez]{larvor2020meteonet}
Gwenna{\"e}lle Larvor, L{\'e}a Berthomier, Vincent Chabot, Brice Le~Pape, Bruno Pradel, and Lior Perez.
\newblock Meteonet, an open reference weather dataset by meteo-france.
\newblock \emph{URL https://meteonet.umr-cnrm.fr/}, 2020.

\bibitem[Leinonen et~al.(2023)Leinonen, Hamann, Nerini, Germann, and Franch]{leinonen2023ldcast}
Jussi Leinonen, Ulrich Hamann, Daniele Nerini, Urs Germann, and Gabriele Franch.
\newblock Latent diffusion models for generative precipitation nowcasting with accurate uncertainty quantification.
\newblock \emph{arXiv preprint arXiv:2304.12891}, 2023.

\bibitem[Li \& He(2025)Li and He]{li2025back}
Tianhong Li and Kaiming He.
\newblock Back to basics: Let denoising generative models denoise, 2025.

\bibitem[Lipman et~al.(2023)Lipman, Chen, Ben-Hamu, Nickel, and Le]{lipman2023flowmatching}
Yaron Lipman, Ricky~TQ Chen, Heli Ben-Hamu, Maximilian Nickel, and Matthew Le.
\newblock Flow matching for generative modeling.
\newblock In \emph{ICLR}, 2023.

\bibitem[Liu \& Lee(2020)Liu and Lee]{liu2020MPLGAN}
Hong-Bin Liu and Ickjai Lee.
\newblock {{MPL-GAN}}: {{Toward Realistic Meteorological Predictive Learning Using Conditional GAN}}.
\newblock \emph{IEEE Access}, 8:\penalty0 93179--93186, 2020.
\newblock ISSN 2169-3536.
\newblock \doi{10.1109/ACCESS.2020.2995187}.

\bibitem[Liu et~al.(2022)Liu, Gong, and Liu]{liu2022Flow}
Xingchao Liu, Chengyue Gong, and Qiang Liu.
\newblock Flow {{Straight}} and {{Fast}}: {{Learning}} to {{Generate}} and {{Transfer Data}} with {{Rectified Flow}}, 2022.

\bibitem[Loshchilov \& Hutter(2019)Loshchilov and Hutter]{loshchilov2019Decoupled}
Ilya Loshchilov and Frank Hutter.
\newblock Decoupled {{Weight Decay Regularization}}, 2019.

\bibitem[Lu et~al.(2026)Lu, Lu, Sun, Zhao, Jiang, Wang, Li, Geng, and He]{lu2026onestep}
Yiyang Lu, Susie Lu, Qiao Sun, Hanhong Zhao, Zhicheng Jiang, Xianbang Wang, Tianhong Li, Zhengyang Geng, and Kaiming He.
\newblock One-step latent-free image generation with pixel mean flows, 2026.

\bibitem[Ma et~al.(2024)Ma, Goldstein, Albergo, Boffi, {Vanden-Eijnden}, and Xie]{ma2024SiT}
Nanye Ma, Mark Goldstein, Michael~S. Albergo, Nicholas~M. Boffi, Eric {Vanden-Eijnden}, and Saining Xie.
\newblock {{SiT}}: {{Exploring Flow}} and {{Diffusion-based Generative Models}} with {{Scalable Interpolant Transformers}}, 2024.

\bibitem[Martorella et~al.(2026)Martorella, Galashov, Krause, Baumann, De~Bortoli, Gretton, and Ommer]{martorella2026iddm}
Tommaso Martorella, Alexandre Galashov, Felix Krause, Stefan~Andreas Baumann, Valentin De~Bortoli, Arthur Gretton, and Bj{\"o}rn Ommer.
\newblock Improved distributional diffusion models.
\newblock \emph{arXiv preprint arXiv:2609.37147}, 2026.

\bibitem[Nguyen et~al.(2025{\natexlab{a}})Nguyen, Nguyen, Vu, Dao, Tran, Tran, and Tran]{nguyen2025improved}
Anh Nguyen, Viet Nguyen, Duc Vu, Trung Dao, Chi Tran, Toan Tran, and Anh Tran.
\newblock Improved training technique for shortcut models, 2025{\natexlab{a}}.

\bibitem[Nguyen et~al.(2025{\natexlab{b}})Nguyen, Ma, Khan, Taieb, and Khan]{nguyen2025raindiff}
Thao Nguyen, Jiaqi Ma, Fahad~Shahbaz Khan, Souhaib~Ben Taieb, and Salman Khan.
\newblock Raindiff: End-to-end precipitation nowcasting via token-wise attention diffusion, 2025{\natexlab{b}}.

\bibitem[{NOAA / National Centers For Environmental Information}(2020)]{NOAA2020StormEvents}
{NOAA / National Centers For Environmental Information}.
\newblock {{NOAA Storm Events Database}}, 2020.

\bibitem[Park et~al.(2026{\natexlab{a}})Park, Song, Chang, Ham, and Hong]{park2026diffusionbased}
Dohyun Park, Changhoon Song, Tengyuan Chang, Yoo-Geun Ham, and Youngjoon Hong.
\newblock Diffusion-based refinement for kilometer-scale probabilistic precipitation nowcasting, 2026{\natexlab{a}}.

\bibitem[Park et~al.(2026{\natexlab{b}})Park, Nam, Lee, Cho, Park, and Lee]{park2026proper}
Joonhyeong Park, Giung Nam, Hyungi Lee, Kyunghyun Cho, Byoungwoo Park, and Juho Lee.
\newblock Proper scoring rule-based diffusion for probabilistic weather forecasting, 2026{\natexlab{b}}.

\bibitem[Pathak et~al.(2022)Pathak, Subramanian, Harrington, Raja, Chattopadhyay, Mardani, Kurth, Hall, Li, Azizzadenesheli, Hassanzadeh, Kashinath, and Anandkumar]{pathak2022FourCastNet}
Jaideep Pathak, Shashank Subramanian, Peter Harrington, Sanjeev Raja, Ashesh Chattopadhyay, Morteza Mardani, Thorsten Kurth, David Hall, Zongyi Li, Kamyar Azizzadenesheli, Pedram Hassanzadeh, Karthik Kashinath, and Animashree Anandkumar.
\newblock {{FourCastNet}}: {{A Global Data-driven High-resolution Weather Model}} using {{Adaptive Fourier Neural Operators}}, February 2022.

\bibitem[Peebles \& Xie(2023)Peebles and Xie]{peebles2023Scalable}
William Peebles and Saining Xie.
\newblock Scalable {{Diffusion Models}} with {{Transformers}}, 2023.

\bibitem[Price et~al.(2024)Price, {Sanchez-Gonzalez}, Alet, Andersson, {El-Kadi}, Masters, Ewalds, Stott, Mohamed, Battaglia, Lam, and Willson]{price2024GenCast}
Ilan Price, Alvaro {Sanchez-Gonzalez}, Ferran Alet, Tom~R. Andersson, Andrew {El-Kadi}, Dominic Masters, Timo Ewalds, Jacklynn Stott, Shakir Mohamed, Peter Battaglia, Remi Lam, and Matthew Willson.
\newblock {{GenCast}}: {{Diffusion-based}} ensemble forecasting for medium-range weather, May 2024.

\bibitem[Ravuri et~al.(2021)Ravuri, Lenc, Willson, Kangin, Lam, Mirowski, Fitzsimons, Athanassiadou, Kashem, Madge, et~al.]{ravuri21:GAN}
Suman Ravuri, Karel Lenc, Matthew Willson, Dmitry Kangin, Remi Lam, Piotr Mirowski, Megan Fitzsimons, Maria Athanassiadou, Sheleem Kashem, Sam Madge, et~al.
\newblock Skilful precipitation nowcasting using deep generative models of radar.
\newblock \emph{Nature}, 597\penalty0 (7878):\penalty0 672--677, 2021.

\bibitem[Ribeiro \& Pucer(2026)Ribeiro and Pucer]{ribeiro2026flowcast}
Bernardo~Perrone Ribeiro and Jana~Faganeli Pucer.
\newblock Flowcast: Advancing precipitation nowcasting with conditional flow matching, 2026.

\bibitem[Rombach et~al.(2022)Rombach, Blattmann, Lorenz, Esser, and Ommer]{rombach2022ldm}
Robin Rombach, Andreas Blattmann, Dominik Lorenz, Patrick Esser, and Bj{\"o}rn Ommer.
\newblock High-resolution image synthesis with latent diffusion models.
\newblock In \emph{Proceedings of the IEEE/CVF conference on computer vision and pattern recognition}, pp.\  10684--10695, 2022.

\bibitem[Ronneberger et~al.(2015)Ronneberger, Fischer, and Brox]{ronneberger2015UNeta}
Olaf Ronneberger, Philipp Fischer, and Thomas Brox.
\newblock U-{{Net}}: {{Convolutional Networks}} for {{Biomedical Image Segmentation}}.
\newblock In Nassir Navab, Joachim Hornegger, William~M. Wells, and Alejandro~F. Frangi (eds.), \emph{Medical {{Image Computing}} and {{Computer-Assisted Intervention}} -- {{MICCAI}} 2015}, volume 9351, pp.\  234--241. Springer International Publishing, Cham, 2015.
\newblock ISBN 978-3-319-24573-7 978-3-319-24574-4.
\newblock \doi{10.1007/978-3-319-24574-4_28}.

\bibitem[Sakaino(2013)]{sakaino2013SpatioTemporal}
Hidetomo Sakaino.
\newblock Spatio-{{Temporal Image Pattern Prediction Method Based}} on a {{Physical Model With Time-Varying Optical Flow}}.
\newblock \emph{IEEE Transactions on Geoscience and Remote Sensing}, 51\penalty0 (5):\penalty0 3023--3036, May 2013.
\newblock ISSN 1558-0644.
\newblock \doi{10.1109/TGRS.2012.2212201}.

\bibitem[Salimans \& Ho(2022)Salimans and Ho]{salimans2022progressivea}
Tim Salimans and Jonathan Ho.
\newblock Progressive distillation for fast sampling of diffusion models, 2022.

\bibitem[Salio et~al.(2025)Salio, Stoll, D'Amen, De~Coning, Msemo, Schmid, Agersten, Sivle, Simon, and Chasco]{salio2025nowcasting}
Paola Salio, Luciana Stoll, Daniela D'Amen, Estelle De~Coning, Hellen Msemo, Franziska Schmid, Solfrid Agersten, Anders Sivle, Andr{\'e} Simon, and Maria~Julia Chasco.
\newblock Nowcasting and early warning systems across wmo regions associations: A pre-ew4all assessment.
\newblock \emph{Bulletin of the American Meteorological Society}, 106\penalty0 (12):\penalty0 E2490--E2508, 2025.
\newblock ISSN 0003-0007, 1520-0477.
\newblock \doi{10.1175/BAMS-D-24-0267.1}.

\bibitem[Sauer et~al.(2023)Sauer, Lorenz, Blattmann, and Rombach]{sauer2023adversarial}
Axel Sauer, Dominik Lorenz, Andreas Blattmann, and Robin Rombach.
\newblock Adversarial diffusion distillation, 2023.

\bibitem[Schusterbauer et~al.(2026{\natexlab{a}})Schusterbauer, Gui, Li, Ma, Krause, and Ommer]{schusterbauer2026patchforcing}
Johannes Schusterbauer, Ming Gui, Yusong Li, Pingchuan Ma, Felix Krause, and Björn Ommer.
\newblock Denoising, fast and slow: Difficulty-aware adaptive sampling for image generation.
\newblock In \emph{Proceedings of the IEEE/CVF Conference on Computer Vision and Pattern Recognition}, 2026{\natexlab{a}}.

\bibitem[Schusterbauer et~al.(2026{\natexlab{b}})Schusterbauer, Wiese, Stracke, Phan, and Ommer]{schusterbauer2026weatherrf-freud}
Johannes Schusterbauer, Jannik Wiese, Nick Stracke, Timy Phan, and Bj{\"o}rn Ommer.
\newblock Probabilistic precipitation nowcasting with rectified flow transformers.
\newblock In \emph{Proceedings of the IEEE/CVF Conference on Computer Vision and Pattern Recognition}, 2026{\natexlab{b}}.

\bibitem[Shazeer(2020)]{shazeer2020glu}
Noam Shazeer.
\newblock {GLU} {Variants} {Improve} {Transformer}, February 2020.
\newblock URL \url{http://arxiv.org/abs/2002.05202}.
\newblock arXiv:2002.05202 [cs].

\bibitem[She et~al.(2024)She, Zhang, Man, and Shao]{she2024LLMDiff}
Lei She, Chenghong Zhang, Xin Man, and Jie Shao.
\newblock {{LLMDiff}}: Diffusion model using frozen llm transformers for precipitation nowcasting.
\newblock \emph{Sensors}, 24\penalty0 (18):\penalty0 6049, January 2024.
\newblock ISSN 1424-8220.
\newblock \doi{10.3390/s24186049}.

\bibitem[Shi et~al.(2015)Shi, Chen, Wang, Yeung, Wong, and Woo]{shi15:ConvLSTM}
Xingjian Shi, Zhourong Chen, Hao Wang, Dit-Yan Yeung, Wai-Kin Wong, and Wang-chun Woo.
\newblock Convolutional lstm network: A machine learning approach for precipitation nowcasting.
\newblock \emph{Advances in neural information processing systems}, 28, 2015.

\bibitem[Shi et~al.(2017)Shi, Gao, Lausen, Wang, Yeung, Wong, and Woo]{shi17:TrajectoryGRU}
Xingjian Shi, Zhihan Gao, Leonard Lausen, Hao Wang, Dit-Yan Yeung, Wai-kin Wong, and Wang-chun Woo.
\newblock Deep learning for precipitation nowcasting: A benchmark and a new model.
\newblock \emph{Advances in neural information processing systems}, 30, 2017.

\bibitem[S{\o}nderby et~al.(2020)S{\o}nderby, Espeholt, Heek, Dehghani, Oliver, Salimans, Agrawal, Hickey, and Kalchbrenner]{sonderby20:MetNet}
Casper~Kaae S{\o}nderby, Lasse Espeholt, Jonathan Heek, Mostafa Dehghani, Avital Oliver, Tim Salimans, Shreya Agrawal, Jason Hickey, and Nal Kalchbrenner.
\newblock Metnet: A neural weather model for precipitation forecasting.
\newblock \emph{arXiv preprint arXiv:2003.12140}, 2020.

\bibitem[Song et~al.(2026)Song, Chang, and Hong]{song2026extremea}
Changhoon Song, Teng~Yuan Chang, and Youngjoon Hong.
\newblock Extreme weather nowcasting via local precipitation pattern prediction, 2026.

\bibitem[Song et~al.(2020)Song, Meng, and Ermon]{song2020ddim}
Jiaming Song, Chenlin Meng, and Stefano Ermon.
\newblock Denoising diffusion implicit models.
\newblock \emph{arXiv preprint arXiv:2010.02502}, 2020.

\bibitem[Song et~al.(2023)Song, Dhariwal, Chen, and Sutskever]{song2023consistencya}
Yang Song, Prafulla Dhariwal, Mark Chen, and Ilya Sutskever.
\newblock Consistency models, 2023.

\bibitem[Sperati et~al.(2016)Sperati, Alessandrini, and Delle~Monache]{sperati_application_2016}
Simone Sperati, Stefano Alessandrini, and Luca Delle~Monache.
\newblock An application of the {ECMWF} {Ensemble} {Prediction} {System} for short-term solar power forecasting.
\newblock \emph{Solar Energy}, 133:\penalty0 437--450, August 2016.
\newblock ISSN 0038-092X.
\newblock \doi{10.1016/j.solener.2016.04.016}.
\newblock URL \url{https://www.sciencedirect.com/science/article/pii/S0038092X1630041X}.

\bibitem[Su et~al.(2023)Su, Lu, Pan, Murtadha, Wen, and Liu]{su2023RoFormer}
Jianlin Su, Yu~Lu, Shengfeng Pan, Ahmed Murtadha, Bo~Wen, and Yunfeng Liu.
\newblock {{RoFormer}}: {{Enhanced Transformer}} with {{Rotary Position Embedding}}, 2023.

\bibitem[Veillette et~al.(2020)Veillette, Samsi, and Mattioli]{veillette2020SEVIR}
Mark Veillette, Siddharth Samsi, and Chris Mattioli.
\newblock {{SEVIR}} : {{A Storm Event Imagery Dataset}} for {{Deep Learning Applications}} in {{Radar}} and {{Satellite Meteorology}}.
\newblock In \emph{Advances in {{Neural Information Processing Systems}}}, volume~33, pp.\  22009--22019. Curran Associates, Inc., 2020.

\bibitem[Wang et~al.(2025)Wang, Gao, Zhu, Huang, and Wang]{wang2025pixnerd}
Shuai Wang, Ziteng Gao, Chenhui Zhu, Weilin Huang, and Limin Wang.
\newblock Pixnerd: Pixel neural field diffusion, 2025.

\bibitem[Wang et~al.(2017)Wang, Long, Wang, Gao, and Yu]{wang2017predrnn}
Yunbo Wang, Mingsheng Long, Jianmin Wang, Zhifeng Gao, and Philip~S Yu.
\newblock Predrnn: Recurrent neural networks for predictive learning using spatiotemporal lstms.
\newblock \emph{Advances in neural information processing systems}, 30, 2017.

\bibitem[Wewer et~al.(2025)Wewer, Pogodzinski, Schiele, and Lenssen]{wewer2025spatial}
Christopher Wewer, Bart Pogodzinski, Bernt Schiele, and Jan~Eric Lenssen.
\newblock Spatial reasoning with denoising models, 2025.

\bibitem[Wilks(2019)]{wilks2019Indices}
D.~S. Wilks.
\newblock Indices of {{Rank Histogram Flatness}} and {{Their Sampling Properties}}.
\newblock \emph{Monthly Weather Review}, 147\penalty0 (2):\penalty0 763--769, February 2019.
\newblock ISSN 0027-0644, 1520-0493.
\newblock \doi{10.1175/MWR-D-18-0369.1}.

\bibitem[Woo \& Wong(2017)Woo and Wong]{woo2017Operational}
Wang-chun Woo and Wai-kin Wong.
\newblock Operational {{Application}} of {{Optical Flow Techniques}} to {{Radar-Based Rainfall Nowcasting}}.
\newblock \emph{Atmosphere}, 8\penalty0 (3):\penalty0 48, February 2017.
\newblock ISSN 2073-4433.
\newblock \doi{10.3390/atmos8030048}.

\bibitem[Wu et~al.(2018)Wu, Su, Wu, Hong, and Hassan]{wu2018Probabilistic}
Yuan-Kang Wu, Po-En Su, Ting-Yi Wu, Jing-Shan Hong, and Mohammad~Yusri Hassan.
\newblock Probabilistic {{Wind-Power Forecasting Using Weather Ensemble Models}}.
\newblock \emph{IEEE Transactions on Industry Applications}, 54\penalty0 (6):\penalty0 5609--5620, November 2018.
\newblock ISSN 0093-9994, 1939-9367.
\newblock \doi{10.1109/TIA.2018.2858183}.

\bibitem[Xu et~al.(2019)Xu, Sun, Zhang, Zhao, and Lin]{xu2019understanding}
Jingjing Xu, Xu~Sun, Zhiyuan Zhang, Guangxiang Zhao, and Junyang Lin.
\newblock Understanding and improving layer normalization, 2019.

\bibitem[Yin et~al.(2024{\natexlab{a}})Yin, Gharbi, Park, Zhang, Shechtman, Durand, and Freeman]{yin2024improved}
Tianwei Yin, Micha{\"e}l Gharbi, Taesung Park, Richard Zhang, Eli Shechtman, Fredo Durand, and William~T. Freeman.
\newblock Improved distribution matching distillation for fast image synthesis, 2024{\natexlab{a}}.

\bibitem[Yin et~al.(2024{\natexlab{b}})Yin, Gharbi, Zhang, Shechtman, Durand, Freeman, and Park]{yin2024onestep}
Tianwei Yin, Micha{\"e}l Gharbi, Richard Zhang, Eli Shechtman, Fredo Durand, William~T. Freeman, and Taesung Park.
\newblock One-step diffusion with distribution matching distillation, 2024{\natexlab{b}}.

\bibitem[Yu et~al.(2024)Yu, Li, Ye, Zhang, Luo, Dai, Wang, and Chen]{yu2024DiffCast}
Demin Yu, Xutao Li, Yunming Ye, Baoquan Zhang, Chuyao Luo, Kuai Dai, Rui Wang, and Xunlai Chen.
\newblock {{DiffCast}}: {{A Unified Framework}} via {{Residual Diffusion}} for {{Precipitation Nowcasting}}.
\newblock In \emph{Proceedings of the {{IEEE}}/{{CVF Conference}} on {{Computer Vision}} and {{Pattern Recognition}}}, pp.\  27758--27767, 2024.

\bibitem[Yu et~al.(2025)Yu, Xiong, Nie, Sheng, Liu, and Luo]{yu2025pixeldit}
Yongsheng Yu, Wei Xiong, Weili Nie, Yichen Sheng, Shiqiu Liu, and Jiebo Luo.
\newblock Pixeldit: Pixel diffusion transformers for image generation, 2025.

\bibitem[Zhang \& Sennrich(2019)Zhang and Sennrich]{zhang2019root}
Biao Zhang and Rico Sennrich.
\newblock Root {Mean} {Square} {Layer} {Normalization}.
\newblock In \emph{Advances in {Neural} {Information} {Processing} {Systems}}, volume~32. Curran Associates, Inc., 2019.
\newblock URL \url{https://papers.nips.cc/paper_files/paper/2019/hash/1e8a19426224ca89e83cef47f1e7f53b-Abstract.html}.

\bibitem[Zhang et~al.(2018)Zhang, Isola, Efros, Shechtman, and Wang]{zhang2018lpips}
Richard Zhang, Phillip Isola, Alexei~A Efros, Eli Shechtman, and Oliver Wang.
\newblock The unreasonable effectiveness of deep features as a perceptual metric.
\newblock In \emph{CVPR}, 2018.

\bibitem[Zhang et~al.(2023)Zhang, Long, Chen, Xing, Jin, Jordan, and Wang]{zhang2023skilful}
Yuchen Zhang, Mingsheng Long, Kaiyuan Chen, Lanxiang Xing, Ronghua Jin, Michael~I Jordan, and Jianmin Wang.
\newblock Skilful nowcasting of extreme precipitation with nowcastnet.
\newblock \emph{Nature}, 619\penalty0 (7970):\penalty0 526--532, 2023.

\bibitem[Zhou et~al.(2025)Zhou, Ermon, and Song]{zhou2025inductive}
Linqi Zhou, Stefano Ermon, and Jiaming Song.
\newblock Inductive moment matching, 2025.

\bibitem[Zhu et~al.(2026)Zhu, Shi, Feng, and Niu]{zhu2026pixelflowcast}
Yufeng Zhu, Chunlei Shi, Yongchao Feng, and Dan Niu.
\newblock Pixelflowcast: Latent-free precipitation nowcasting via pixel mean flows, 2026.

\end{thebibliography}
